%% file: main.tex
\documentclass{article}

\IfFileExists{anon2026_conference.sty}{
  \usepackage{anon2026_conference,times}
  \anonfinalcopy
}{
  \usepackage[margin=1in]{geometry}
  \usepackage{times}
}

\IfFileExists{math_commands.tex}{\input{math_commands.tex}}{}

\usepackage{graphicx}
\usepackage{tikz}
\usepackage{tgheros}
\usepackage{comment}
\usepackage{graphicx}
\usepackage{float}
\usepackage{booktabs}
\usepackage{multirow}
\usepackage{amsmath}
\usepackage{amssymb}
\usepackage{hyperref}
\usepackage{url}
\usepackage[table]{xcolor}
\usepackage{pifont}
\usepackage{makecell}
\usepackage{enumitem}
\usepackage{wrapfig}
\usepackage{subcaption}
\setlist{itemsep=0pt}
\usepackage[most]{tcolorbox}
\usetikzlibrary{arrows.meta,calc,backgrounds}
\usepackage{tabularx}
\usepackage{pgfplots}
\usepackage{placeins} 
\pgfplotsset{compat=1.18}

\newcommand{\vistagain}[2]{\textbf{#1}\rlap{\,\raisebox{-0.45ex}{\fontsize{4.5pt}{5pt}\selectfont\textcolor{green!45!black}{(#2)}}}}
\definecolor{vistacrim}{HTML}{A71D32}
\newcolumntype{Y}{>{\centering\arraybackslash}X}
\newcommand{\VYes}{\ding{51}}
\newcommand{\VNo}{\ding{55}}
\newcommand{\VNR}{--}

\newcommand{\yue}[1]{\textcolor{red}{ Yue: #1 }}

\title{
VISTA: A Multimodal Egocentric Benchmark for Goal-Oriented Perception and Assistance to Blind and Low-Vision Users
}

\author{%
\parbox[t]{\textwidth}{%
\raggedright\normalfont\small
\setlength{\baselineskip}{8pt}
\textbf{No\'e Mosca}$^{1,2,*}$ \enspace
\textbf{Mohamed Rayan Barhdadi}$^{3,*}$ \enspace
\textbf{Muhammad Hamza}$^{4,*}$\enspace
\textbf{Joshua Yeh}$^{1,*}$ \enspace
\textbf{Suryodaya Bikram Shahi}$^{1,*}$ \enspace
\textbf{Zhiyan Li}$^{1}$ \enspace
\textbf{Selim-Antoine Lali}$^{1}$ \enspace
\textbf{Weiyue Li}$^{1}$ \enspace
\textbf{Hasan Kurban}$^{5}$ \enspace
\textbf{Erchin Serpedin}$^{3}$ \enspace
\textbf{Gang Luo}$^{1}$ \textbf{Yue Zhang}$^{1,\dagger}$ \enspace
\textbf{Minghan Li}$^{1,\dagger}$ \enspace
\textbf{Mengyu Wang}$^{1,\dagger}$ \\[1 em]%
\parbox[t]{\linewidth}{%
\raggedright\normalfont\small
\setlength{\baselineskip}{11pt}
$^{1}$Harvard AI and Robotics Lab, Harvard University \enspace
$^{2}$T\'el\'ecom Paris, Institut Polytechnique de Paris \enspace
$^{3}$Texas A\&M University \enspace
$^{4}$University of Toronto \enspace
$^{5}$Hamad Bin Khalifa University \enspace \\[1 em]%
$^{*}$Contributed equally as co-first authors. \enspace
$^{\dagger}$Contributed equally as co-senior authors.}
}}

\begin{document}

\maketitle

\input{sec_merge/0_abstract}
\input{sec_merge/1_intro}
\input{sec_merge/2_related}
\input{sec_merge/3_dataset}
\input{sec_merge/4_model_v2}
\input{sec_merge/5_experiment_v2}

\input{sec_merge/6_conclusion}

\clearpage

\clearpage
\bibliography{anon2026_conference}
\bibliographystyle{anon2026_conference}

\clearpage
\appendix
\input{sec_merge/appendix}

\end{document}

%% file: math_commands.tex

\usepackage{amsmath,amsfonts,bm}









\def\eqref#1{equation~\ref{#1}}









\def\1{\bm{1}}










\DeclareMathAlphabet{\mathsfit}{\encodingdefault}{\sfdefault}{m}{sl}
\SetMathAlphabet{\mathsfit}{bold}{\encodingdefault}{\sfdefault}{bx}{n}













%% file: sec_merge/0_abstract.tex
\begin{abstract}
Blindness and low vision limit many people's ability to complete daily tasks, motivating AI assistants that provide useful guidance. Yet, existing benchmarks do not adequately test whether vision models can provide effective everyday assistance. In particular, i) existing benchmarks primarily rely on visual and textual inputs, with limited support for other modalities that are increasingly available in egocentric wearable devices such as gaze, inertial measurements, spatial trajectories, and others; and ii) existing evaluations emphasize scene understanding and question answering, with limited coverage of goal-conditioned guidance for blind and low-vision users across everyday tasks.
To bridge this gap, we introduce \textbf{VISTA} (\textbf{V}isually \textbf{I}mpaired \textbf{S}cene and \textbf{T}ask \textbf{A}nnotations), a multimodal egocentric benchmark for goal-oriented assistance for blind and low-vision users. VISTA contains $\sim$1,000 egocentric recordings captured with Project Aria glasses; each recording provides synchronized sensor streams, six of which serve as model inputs, and three different types of annotation, covering ten task categories (TC), from object localization to hazard detection. We evaluate 14 representative vision-language models zero-shot on VISTA and design a lightweight multimodal model baseline that integrates all six modalities. Experimental results show that, although existing models can often describe egocentric scenes, they remain limited in generating goal-directed assistive guidance. VISTA thus provides a benchmark for evaluating multimodal models that could improve everyday accessibility. Code and dataset can be found at \url{https://harvard-ai-and-robotics-lab.github.io/VISTA}.

\end{abstract}

%% file: sec_merge/1_intro.tex
\section{Introduction}
\label{sec:intro}

Blindness and low vision can substantially limit individuals' ability to complete daily tasks, creating an urgent need for benchmarks that evaluate assistance in everyday settings~\citep{who2019world}. Individuals regularly encounter visual information that is difficult or impossible to access without assistance: locating objects, detecting hazards, reading text, interpreting social signals, estimating distance, and navigating unfamiliar environments. Recent advances in vision-language models (VLMs)~\citep{radford2021clip,li2023blip2,liu2023llava,hu2024bliva,openai2023gpt4} suggest that they could become a core component of wearable assistive systems. However, reliable assistance requires more than generic image understanding. A useful assistant must interpret an egocentric scene from the user's perspective, infer what matters for the user's current goal~\citep{behari2024decision}, avoid unsafe suggestions, and generate concise guidance that can be acted on immediately.

Current benchmarks do not fully capture these requirements. While standard VQA and captioning datasets~\citep{goyal2017vqav2,lin2014mscoco} evaluate perception in static settings, and accessibility-oriented datasets~\citep{gurari2018vizwiz,xiao2025egoblind} better reflect images captured by blind and low-vision users, they exhibit two critical limitations.
First, existing benchmarks mainly focus on visual and textual inputs, overlooking other sensing modalities that are increasingly available in modern egocentric wearable devices, such as audio, eye tracking, inertial signals, and spatial tracking. 
Second, current models are typically evaluated as vision-language systems, rather than as \textit{multimodal assistive models} capable of integrating visual observations, gaze, motion, and spatial trajectories~\citep{chen2026memobench}. 
A model may correctly describe the objects in a scene while missing which object the user is attending to, or misjudge spatial relations due to a lack of inertial context. This gap motivates a benchmark whose target is not only ``what is in the image?'' but also ``where is the user, how are they moving, and what should they do next?''

\setlength{\intextsep}{0pt}
\begin{wrapfigure}{r}{0.5\textwidth}
\vspace{0pt}
\centering
\includegraphics[width=0.48\textwidth]{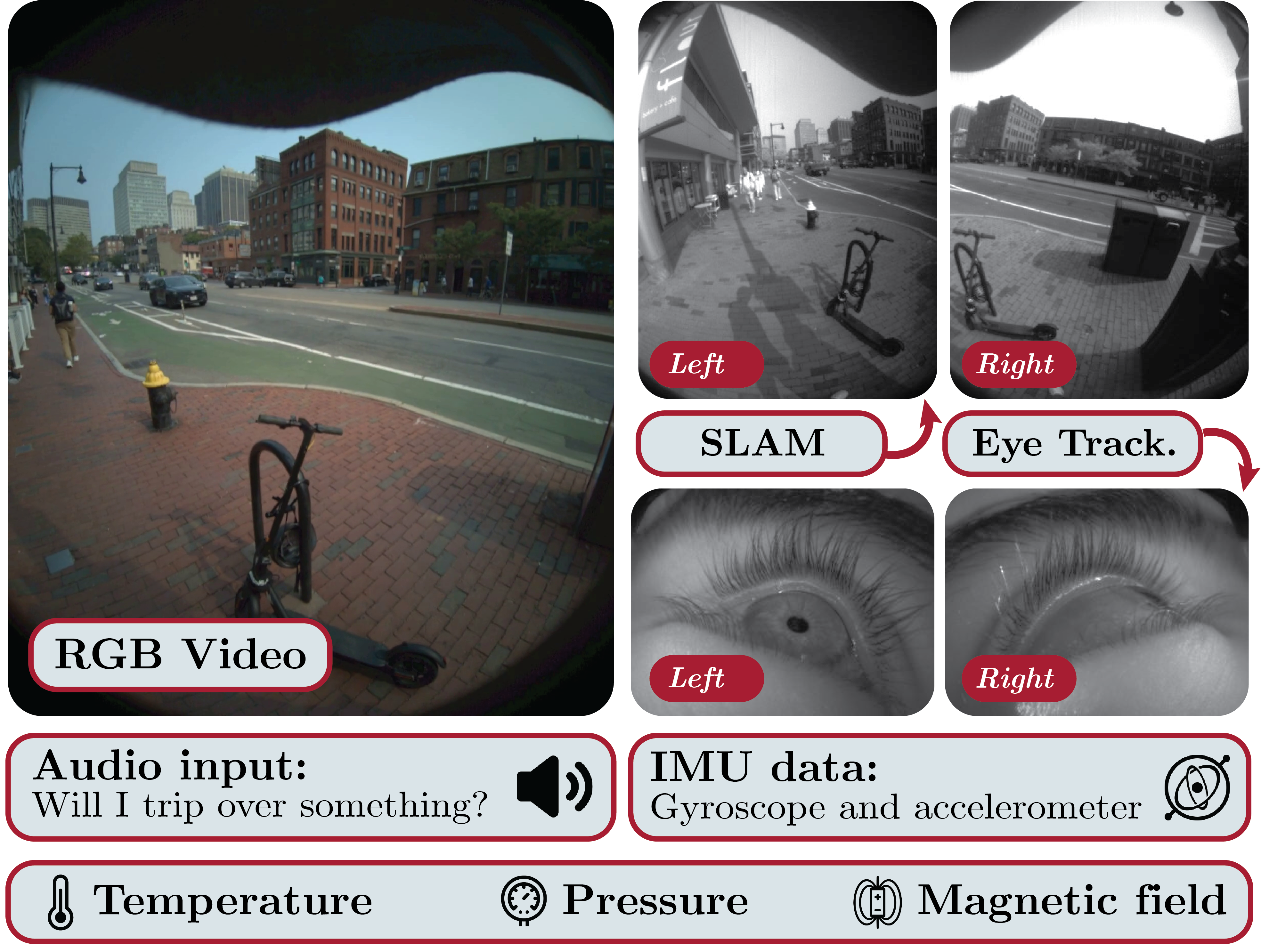}
\caption{\textbf{Sensor modalities of the Aria glasses.}
  VISTA records synchronized RGB video, stereo SLAM and eye-tracking cameras, audio, IMU, barometer, and magnetometer.
  See Appendix~\ref{sec:app_sensor_streams} (Figure~\ref{fig:vista_five_modalities}) for an expanded visualization.}
\label{fig:aria_modalities}
\end{wrapfigure}

We present \textbf{VISTA}, a multimodal egocentric benchmark for evaluating vision-language models as next-step assistants for blind and low-vision users. VISTA contains $\sim$1,000 short wearable recordings ($\sim$170 min) collected with Project Aria glasses~\citep{engel2023aria} across ten assistive task categories, including object location, hazard detection, navigation, social cues, distance estimation, task instruction, object query, text understanding, and motion understanding. Unlike prior works, VISTA synchronizes egocentric clips combining RGB, stereo SLAM and eye-tracking cameras, dual IMU, magnetometer and barometer, together with
6-DoF trajectories, semi-dense point clouds, per-frame gaze and hand tracking and natural-language annotations for scene description, visual question answering, and goal-conditioned instructions. The goal-instruction annotations require models to reason about relevant objects, spatial relations, hazards, and social cues to generate actionable guidance.

We explicitly frame VISTA as a benchmark for \emph{next-step assistance}. Given a wearable first-person recording and a user goal, VISTA requires models to translate egocentric scene understanding into user-centered actions. We conduct zero-shot evaluations with open and proprietary vision-language models and design a lightweight multimodal baseline whose inputs are RGB video and five sensor streams (six modalities in total). Experimental results show that although existing models can often describe scenes, they remain limited in producing task-oriented, safety-aware guidance.

To summarize, our contributions are:

\begin{enumerate}
    \item We introduce \textbf{VISTA}, a multimodal egocentric benchmark for goal-oriented assistance for blind and low-vision users.

    \item We evaluate state-of-the-art VLMs zero-shot on VISTA and show that current models make frequent errors on assistive tasks, particularly on goal-conditioned guidance.

    \item We design a multimodal baseline that fuses these sensor streams with RGB; combined with task adaptation, it improves answer quality on every backbone, most strongly on guidance.
\end{enumerate}


%% file: sec_merge/2_related.tex
\section{Related Work}
\label{sec:related}

\paragraph{Egocentric video understanding.}
First-person datasets such as EPIC-Kitchens~\citep{damen2018epic,damen2020epic100}, Charades-Ego~\citep{sigurdsson2018charadesego}, Ego4D~\citep{grauman2022ego4d}, and Ego-Exo4D~\citep{grauman2024egoexo4d} support activity recognition and long video understanding. Egocentric QA benchmarks test task understanding~\citep{egotaskqa}, long-term reasoning~\citep{egoschema}, episodic memory~\citep{qaego4d}, instructional help~\citep{assistq}, and scene-text reading~\citep{zhou2025egotextvqa}. These benchmarks target general users and mostly use RGB video, even though devices such as the Project Aria glasses~\citep{engel2023aria} also record audio, IMU, gaze, and trajectory data. VISTA combines these sensor streams with annotations written for blind and low-vision users.

\paragraph{Visual assistance for blind and low-vision users.}
VizWiz~\citep{gurari2018vizwiz} contains questions that blind users asked about their own photos, but it is limited to single images. VIEW-QA~\citep{viewqa} and EgoBlind~\citep{xiao2025egoblind} extend this setting to egocentric video, while WalkVLM~\citep{yuan2024walkvlm} and GuideDog~\citep{kim2025guidedog} focus on walking and navigation. Assistive apps and prototypes based on OCR, SLAM, or depth sensing solve specific tasks but are hard to compare~\citep{okolo2024assistive,mukhiddinov2021smart}. Most of these works use RGB input with QA or navigation targets only. VISTA adds synchronized wearable sensors and provides scene descriptions, question--answer pairs, and goal-conditioned instructions across ten assistive task categories.

\paragraph{Multimodal models and sensor fusion.}
Open VLMs such as LLaVA-Video~\citep{zhang2024llavavideo}, InternVL3~\citep{zhu2025internvl3}, Qwen3-VL~\citep{bai2025qwen3vl}, and Gemma~3~\citep{gemmateam2025gemma3} perform well on images and video but only accept visual and text inputs. For other modalities, attention bottlenecks~\citep{nagrani2021attention} exchange information between streams through a few shared tokens, PRIMUS~\citep{das2025primus} provides a pretrained IMU encoder, and LoRA~\citep{hu2022lora} adapts large models with few trainable parameters. Few works apply these tools to assistive VLMs with wearable sensors; we combine them in a sensor-augmented baseline for VISTA (Sec.~\ref{sec:model_design}).

\paragraph{Evaluating open-ended assistance.}
Reference-based metrics such as BLEU~\citep{papineni2002bleu}, ROUGE~\citep{lin2004rouge}, BERTScore~\citep{zhang2020bertscore}, and Sentence-BERT similarity~\citep{reimers2019sentencebert} measure overlap with references but do not check whether a response matches the scene, completes the task, or is useful to a blind user. LLM-as-a-judge methods~\citep{zheng2023judging} score open-ended answers more flexibly and agree well with human ratings. VISTA uses a vision-language judge that sees the video frames and scores evidence grounding, answer correctness, task completion, and accessible communication.

%% file: sec_merge/3_dataset.tex
\begin{figure}[t]
    \centering
    \includegraphics[width=1\textwidth]{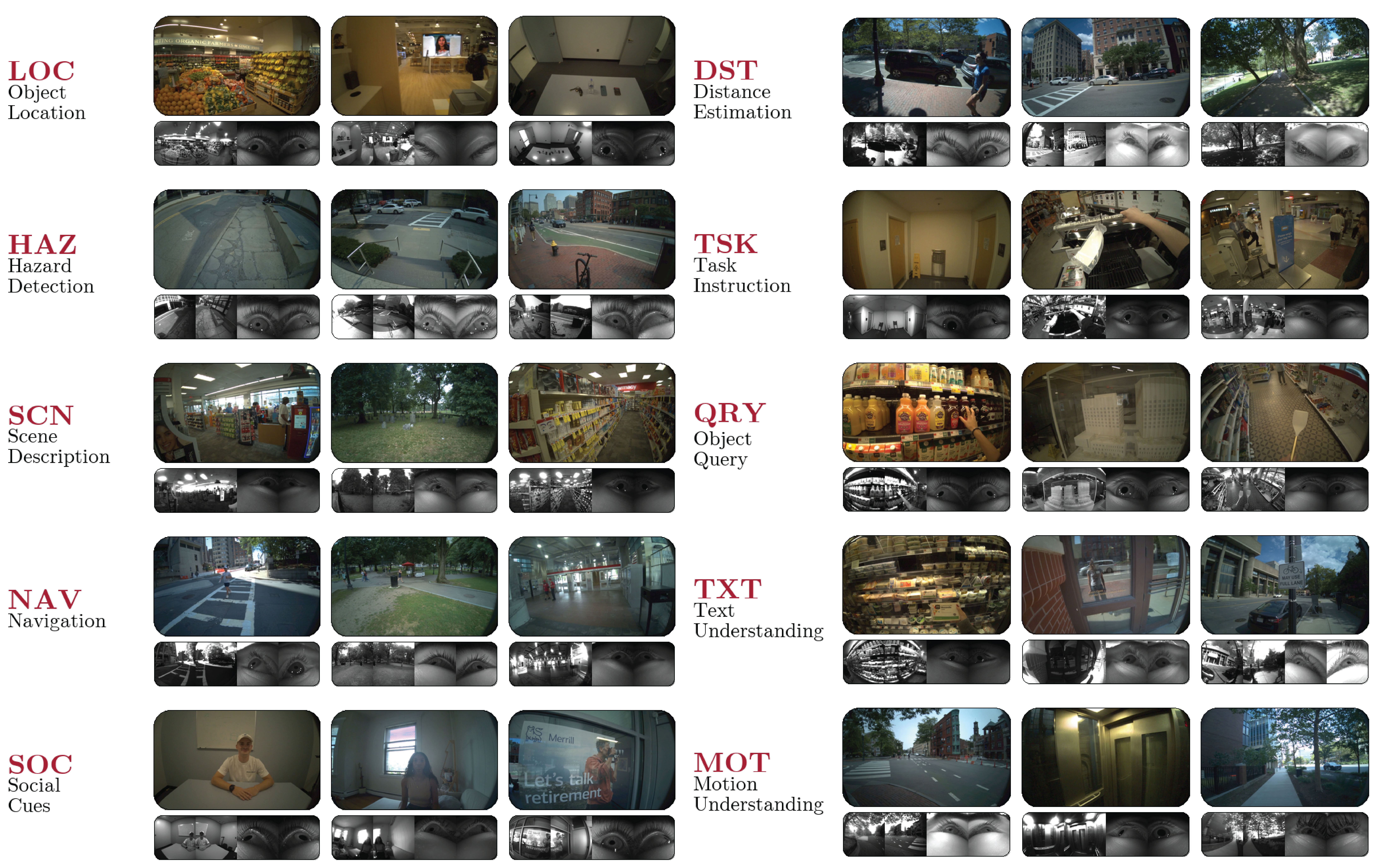}
    \caption{\textbf{Examples of the ten task categories.} Each strip shows three clips, with the RGB frame on top and the SLAM and eye-tracking frames below. Each category has about 100 clips (997 in total); definitions are in Table~\ref{tab:task_categories}.}
    \label{fig:task_samples}
\end{figure}

\section{VISTA Benchmark}
\label{sec:dataset}

\subsection{Overview}

\begin{figure}[t]
\centering
\includegraphics[width=0.85\linewidth]{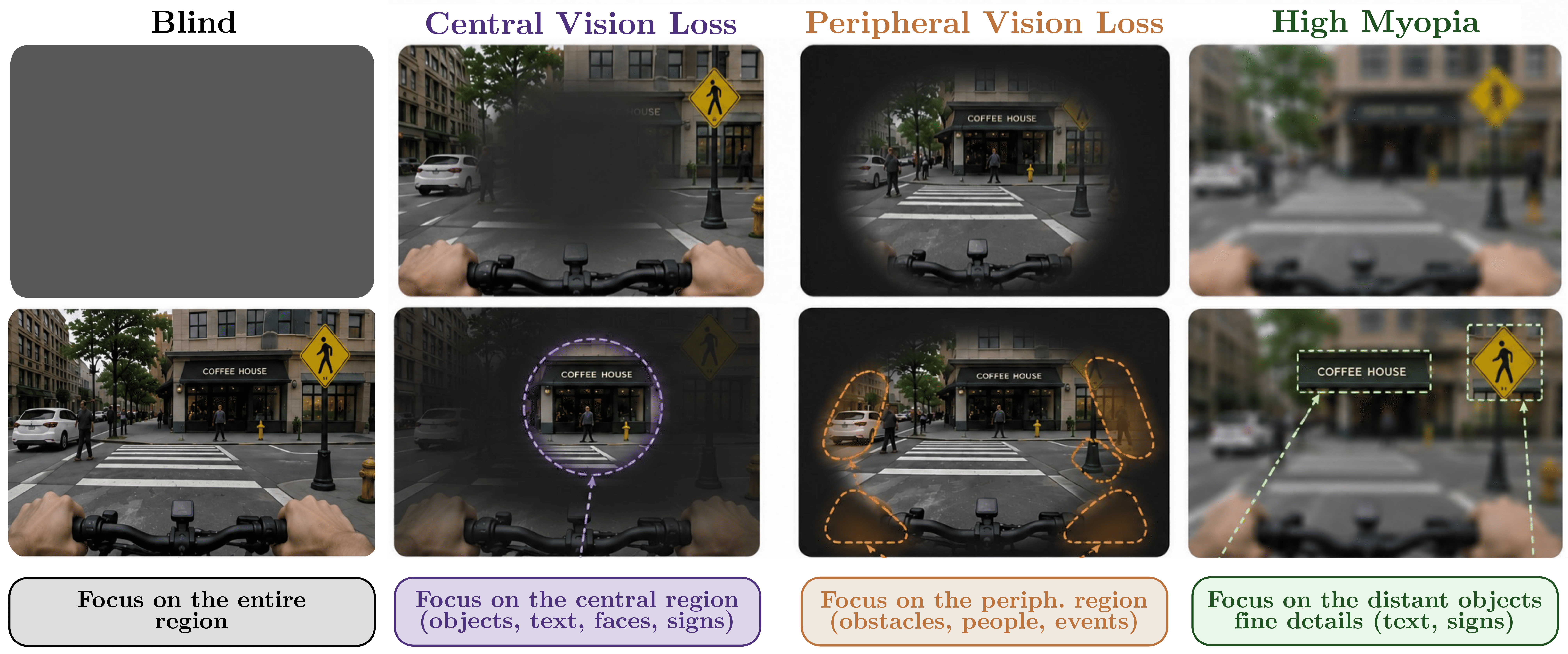}
\caption{\textbf{Simulated vision conditions (conceptual illustration).} Top: The same scene under blindness, central vision loss, peripheral vision loss, and high myopia. Bottom: Regions that may be most useful to describe for each condition. The regions are illustrative and are not evaluated in this work.}
\label{fig:placeholder}
\end{figure}

VISTA contains egocentric recordings captured with first-generation Project Aria glasses~\citep{engel2023aria}. The recordings are organized into ten assistive task categories with approximately balanced coverage (Figure~\ref{fig:task_samples}). Each recording lasts between roughly 10 seconds and one minute and synchronizes multiple modalities. Participants wore the glasses while moving through indoor and outdoor environments and simulating practical assistive situations. The recordings include kitchens, corridors, streets, transit-like settings, parks, and other everyday locations with variation in lighting, motion, pedestrian activity, and ambient noise. This diversity in capture conditions and sensing modalities is important, as the specific information a user needs from a scene depends heavily on the nature of their visual impairment. Figure~\ref{fig:placeholder} is a conceptual illustration of how different vision conditions may change which parts of a scene are most useful to describe.


\subsection{Data Collection Setup}

The Aria glasses record multiple time-aligned sensor streams. The sensor suite includes a forward-facing egocentric RGB camera; two grayscale fisheye SLAM cameras (left and right) used for visual odometry and spatial localization; an inward-facing eye-tracking camera; two IMUs (left and right) measuring 3-axis acceleration and angular velocity; a barometer; and a magnetometer measuring the 3-axis magnetic field. The zero-shot baselines use RGB frames and task text. Our multimodal baseline adds five sensor streams derived from these recordings (IMU, magnetometer/barometer, eye gaze, hand tracking, and 6-DoF trajectory; Section~\ref{sec:model_design}); audio and the raw SLAM and eye-tracking images are recorded but are not model inputs.

\subsection{Data Annotation Pipeline}

\begin{wrapfigure}{r}{0.55\textwidth}
\centering
\includegraphics[width=0.52\textwidth]{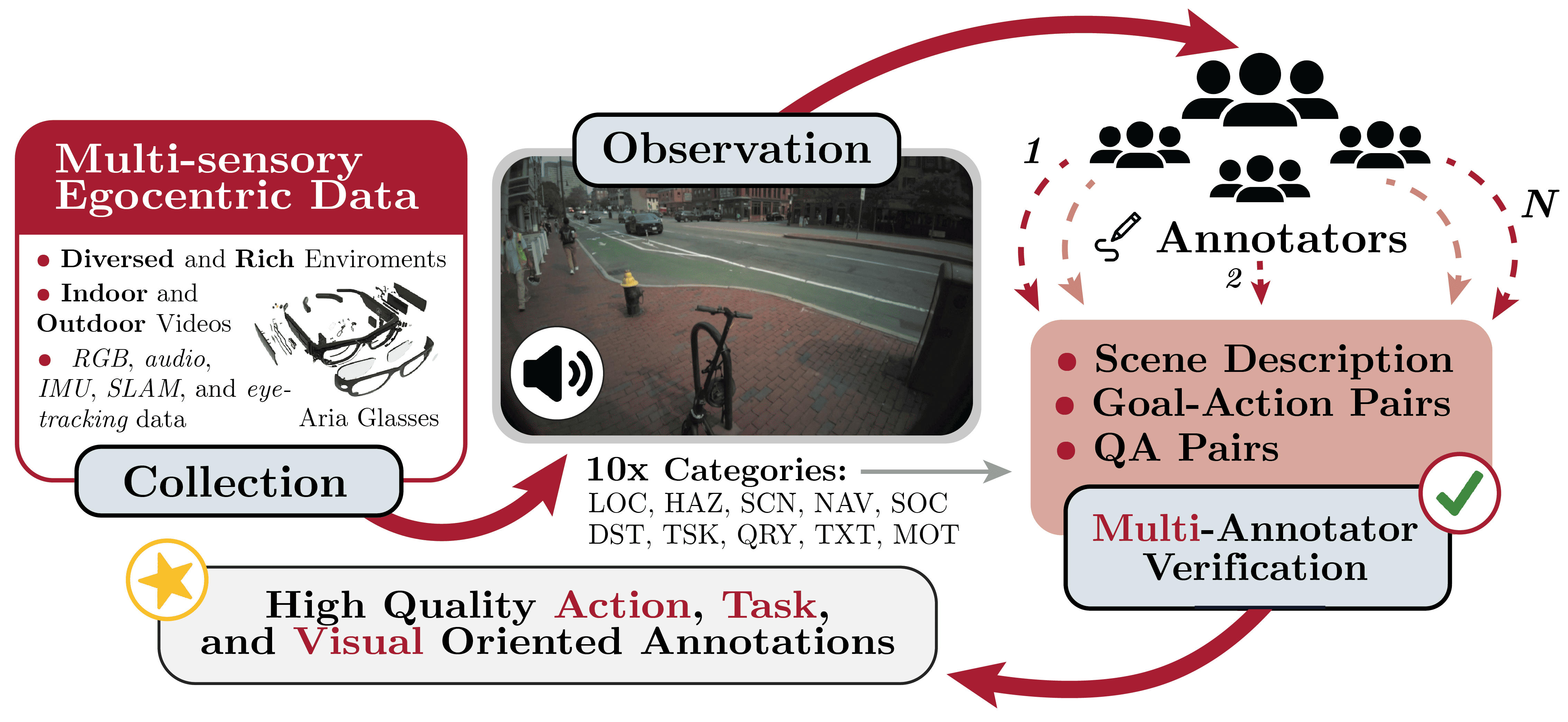}
\caption{\textbf{Annotation pipeline.} Recorded clips are annotated by multiple annotators with scene descriptions, goal--action pairs, and QA pairs, and then verified.}
\label{pipeline}
\end{wrapfigure}

Figure~\ref{pipeline} illustrates the end-to-end annotation pipeline, from clip review to structured label writing and quality control.

\begingroup
\definecolor{acrim}{HTML}{A71D32}\definecolor{alight}{HTML}{F2F2F3}%
\newcommand{\annex}[2]{
  \par\vspace{2pt}\noindent\begin{tikzpicture}[x=1pt,y=1pt]
    \node[anchor=north west,inner xsep=5pt,inner ysep=3.5pt,text width=\dimexpr\linewidth-11pt\relax,align=left,
      font=\small] (b) at (0.5,0) {\raggedright #2};
    \begin{scope}[on background layer]
      \draw[acrim,line width=1pt,rounded corners=4pt,fill=alight]
        ($(b.north west)+(0,7)$) rectangle ($(b.south east)+(-.5pt,0)$);
    \end{scope}
    \node[anchor=west,fill=acrim,rounded corners=3pt,inner xsep=5pt,inner ysep=2.2pt,
      text width=112pt,align=left,text=white,font=\bfseries\footnotesize] at ($(b.north west)+(7,7)$) {#1};
  \end{tikzpicture}\par\vspace{-3pt}}
\textbf{Annotation schema.} VISTA annotations are designed around assistive use rather than generic visual labeling. Each clip contains three annotation types:

\noindent\textit{i) Scene descriptions} summarize the environment, relevant objects, spatial layout, events, and possible risks.
\annex{(i) Scene description}{``You are standing beside a large tree in a grassy park. Thick roots extend across the ground directly in front of you, creating a tripping hazard.''}
\noindent\textit{ii) Question-answer pairs} represent questions a blind or low-vision user might ask, paired with answers grounded in the clip.
\annex{(ii) Question--answer pair}{\textbf{Question:} ``How could I get a fork?'' \textbf{Answer:} ``The utensil dispenser is directly ahead on the counter. [\ldots] press down on a handle in one of the two left slots to release a fork.''}
\noindent\textit{iii) Goal-instruction pairs} state a high-level assistive goal and a concrete instruction for accomplishing it.
\annex{(iii) Goal--instruction pair}{\textbf{Goal:} ``Bypass the landscaping obstacle to reach the street.'' \textbf{Instruction:} ``Take 2 steps to the right, and follow the paved path around the bush, avoiding the dirt bed.''}
\endgroup

The goal-instruction layer is the main distinction from conventional captioning or VQA. For example, instead of only describing that a staircase is visible, an annotation may define the goal ``go down the stairs safely'' and the instruction ``move forward slowly, keep to the handrail on the right, and step down carefully.'' This structure enables evaluation of guidance quality, specificity, and safety.

\textbf{Annotation protocol.}
Eight trained annotators labeled the clips in a customized Label Studio~\citep{tkachenko2020labelstudio} interface, following written guidelines and a demonstration of the workflow. For each clip, annotators watched the full video, wrote a clip-level scene description of relevant objects, layout, events, and risks, and then wrote question--answer pairs and one or more goal--instruction pairs that a blind or low-vision user might need. They could pause, rewind, or revisit frames to verify object locations, visible text, motion, and safety-relevant details. Task labels follow the test-case subset (TC01--TC10), and all natural-language fields are written manually.

\setlength{\intextsep}{0pt}

A reviewer from the blind and low-vision (BLV) community assessed 27 clips (34 annotations) for clarity, usefulness, and appropriateness. The feedback pointed to overly long responses, unclear spatial descriptions, missing hazard information, and instructions that relied on the user's vision. We added this feedback to the guidelines, and three annotators then revised every clip in a second full pass.

To measure consistency, 30 clips were labeled by multiple annotators. Their annotations reach a BERTScore~\citep{zhang2020bertscore} of 0.897 for question--answer pairs and 0.877 for scene descriptions (SBERT~\citep{reimers2019sentencebert}: 0.634 and 0.666), indicating similar content with varied wording. We keep all overlapping annotations as multiple references rather than merging them.

\textbf{Accessibility and ethics.} All participants provided consent before data collection. Recordings in public spaces were restricted to nonsensitive environments. Face and license-plate blurring~\citep{raina2023egoblur} is applied before release to reduce bystander privacy risk. Since the target domain involves assistance for disabled users and safety-relevant guidance, VISTA should be used with care: benchmark scores should not be interpreted as sufficient evidence for deployment without human-centered evaluation, safety testing, and participation from blind and low-vision users.

\textbf{Annotation Statistics.}
The current annotation release comprises 1,063 scene descriptions, 1,191
question-answer pairs, and 2,067 goal-instruction pairs across 997 clips. On
average, each clip includes 1.07 scene descriptions, 1.19 question-answer pairs,
and 2.07 goal-instruction pairs. Table~\ref{tab:fieldstats} summarizes these
statistics globally.

Across the full dataset, the ten task categories are approximately balanced, containing 97--102 clips,
with minor variation attributable to the collection and filtering process.
Annotation density is likewise consistent across subsets, ranging from 1.00 to 1.12
scene descriptions per clip, 1.09 to 1.29 question-answer pairs per clip, and 1.94
to 2.19 goal-instruction pairs per clip. This provides approximately balanced coverage across the ten assistive task categories, reducing skew toward any single task type.

\subsection{Dataset Comparison}

Table~\ref{tab:egovqa_comparison} compares VISTA with representative egocentric VQA datasets. Existing general-purpose datasets, such as EgoTaskQA, VidEgoThink, EgoSchema, and EgoMemoria, emphasize broad reasoning over egocentric videos~\citep{egotaskqa,videgothink,egoschema,egomemoria}, but they are not designed around blind and low-vision users. Assistive datasets such as QAEgo4D, AssistQ, and EgoTextVQA move closer to task-oriented assistance~\citep{qaego4d,assistq,zhou2025egotextvqa}, yet they mainly focus on questions and answers and do not provide the richer supervision needed for grounded guidance. Datasets centered on blind and low-vision users, including VizWiz, VIEW-QA, and EgoBlind, are more aligned with accessibility scenarios~\citep{gurari2018vizwiz,xiao2025egoblind}, but they are typically image-based or limited to QA-style annotations rather than multi-level assistive understanding.

\input{tabs/Ego-VQA-datasets}

VISTA fills this gap by combining assistive scenarios designed around blind and low-vision users (recorded by sighted participants who simulated them), egocentric video, open-ended QA, scene descriptions, and goal-instruction pairs in a single benchmark. This design makes VISTA more suitable for evaluating whether a model can not only answer questions about an egocentric scene, but also summarize the environment and produce actionable, safety-aware guidance for BLV users.

%% file: tabs/Ego-VQA-datasets.tex
\begin{table*}[t]
\centering
\caption{\textbf{Comparison with egocentric and assistive VQA datasets.} BLV: blind and low-vision users; Multi-Ans.: multiple reference answers; Ques.-Cat.: question-category labels; OE/MC: open-ended/multiple-choice QA. \VYes/\VNo: supported/not supported; \VNR: not reported. Human eval.: human assessment reported; for VISTA, a BLV community reviewer assessed 34 annotations from 27 clips (model outputs were not human-evaluated). VISTA (highlighted) is the only dataset among those compared that supports scene description, QA, and goal-driven instruction together.}
\label{tab:egovqa_comparison}
\vspace{1mm}

\footnotesize
\setlength{\tabcolsep}{2.2pt}
\renewcommand{\arraystretch}{1.20}

\resizebox{\textwidth}{!}{%
\begin{tabular}{@{}ll*{18}{c}@{}}
\toprule
\multirow{2}{*}{\textbf{Type}}
& \multirow{2}{*}{\textbf{Dataset}}
& \multirow{2}{*}{\makecell{\textbf{\#}\\\textbf{Videos}}}
& \multirow{2}{*}{\makecell{\textbf{\#}\\\textbf{Queries}}}
& \multirow{2}{*}{\makecell{\textbf{Avg. len.}\\\textbf{(s)}}}
& \multicolumn{2}{c}{\textbf{Setting}}
& \multicolumn{5}{c}{\textbf{Modalities}}
& \multicolumn{2}{c}{\textbf{Annotation}}
& \multicolumn{2}{c}{\textbf{QA format}}
& \multicolumn{3}{c}{\textbf{Supported tasks}}
& \multirow{2}{*}{\makecell{\textbf{Human}\\\textbf{eval.}}} \\
\cmidrule(lr){6-7}
\cmidrule(lr){8-12}
\cmidrule(lr){13-14}
\cmidrule(lr){15-16}
\cmidrule(lr){17-19}
& & & &
& \makecell{\textbf{BLV}\\\textbf{orient.}}
& \makecell{\textbf{Smart}\\\textbf{glasses}}
& \textbf{RGB}
& \textbf{Audio}
& \textbf{Gaze}
& \textbf{IMU}
& \makecell{\textbf{SLAM/}\\\textbf{depth}}
& \makecell{\textbf{Multi-}\\\textbf{Ans.}}
& \makecell{\textbf{Ques.-}\\\textbf{Cat.}}
& \textbf{OE}
& \textbf{MC}
& \makecell{\textbf{Scene}\\\textbf{desc.}}
& \textbf{QA}
& \makecell{\textbf{Goal}\\\textbf{instr.}}
& \\
\midrule

\multirow{4}{*}{\makecell[l]{General\\purpose}}
& EgoTaskQA~\citep{egotaskqa}
& 2K & 40K & 25.0
& \VNo & \VNo
& \VYes & \VNR & \VNo & \VNo & \VNo
& \VNo & \VYes & \VYes & \VNo
& \VNo & \VYes & \VNo & \VNR \\

& VidEgoThink~\citep{videgothink}
& 217 & 600 & 23.4
& \VNo & \VNo
& \VYes & \VNR & \VNo & \VNo & \VNo
& \VNo & \VYes & \VYes & \VNo
& \VNo & \VYes & \VNo & \VNR \\

& EgoSchema~\citep{egoschema}
& 5K & 5K & 180.0
& \VNo & \VNo
& \VYes & \VNR & \VNo & \VNo & \VNo
& \VNo & \VNo & \VNo & \VYes
& \VNo & \VYes & \VNo & \VNR \\

& EgoMemoria~\citep{egomemoria}
& 629 & 7K & 858.5
& \VNo & \VNo
& \VYes & \VNR & \VNo & \VNo & \VNo
& \VNo & \VNo & \VNo & \VYes
& \VNo & \VYes & \VNo & \VNR \\

\midrule

\multirow{3}{*}{Assistive}
& QAEgo4D~\citep{qaego4d}
& 1.3K & 14.5K & 495.1
& \VYes & \VNo
& \VYes & \VNR & \VNo & \VNo & \VNo
& \VNo & \VNo & \VYes & \VNo
& \VNo & \VYes & \VNo & \VNR \\

& AssistQ~\citep{assistq}
& 100 & 531 & 115.0
& \VYes & \VNo
& \VYes & \VNR & \VNo & \VNo & \VNo
& \VNo & \VYes & \VNo & \VYes
& \VNo & \VYes & \VYes & \VNR \\

& EgoTextVQA~\citep{zhou2025egotextvqa}
& 1.5K & 7.0K & 101.7
& \VYes & \VNo
& \VYes & \VNR & \VNo & \VNo & \VNo
& \VNo & \VYes & \VYes & \VNo
& \VNo & \VYes & \VNo & \VNR \\

\midrule

& VIEW-QA~\citep{viewqa}
& 1.0K & 4.1K & 34.4
& \VYes & \VNo
& \VYes & \VNR & \VNo & \VNo & \VNo
& \VNo & \VYes & \VYes & \VNo
& \VNo & \VYes & \VNo & \VNR \\

& EgoBlind~\citep{xiao2025egoblind}
& 1.3K & 5.3K & 40.0
& \VYes & \VNo
& \VYes & \VNR & \VNo & \VNo & \VNo
& \VYes & \VYes & \VYes & \VNo
& \VNo & \VYes & \VNo & \VYes \\

\multirow{-4}{*}{BLV}
& \cellcolor{vistacrim!7}\textbf{VISTA (Ours)}
& \cellcolor{vistacrim!7}\textbf{1.0K}
& \cellcolor{vistacrim!7}\textbf{1.2K}
& \cellcolor{vistacrim!7}\textbf{10.5}
& \cellcolor{vistacrim!7}\VYes
& \cellcolor{vistacrim!7}\VYes
& \cellcolor{vistacrim!7}\VYes
& \cellcolor{vistacrim!7}\VYes
& \cellcolor{vistacrim!7}\VYes
& \cellcolor{vistacrim!7}\VYes
& \cellcolor{vistacrim!7}\VYes
& \cellcolor{vistacrim!7}\VYes
& \cellcolor{vistacrim!7}\VYes
& \cellcolor{vistacrim!7}\VYes
& \cellcolor{vistacrim!7}\VNo
& \cellcolor{vistacrim!7}\VYes
& \cellcolor{vistacrim!7}\VYes
& \cellcolor{vistacrim!7}\VYes
& \cellcolor{vistacrim!7}\VYes \\

\bottomrule
\end{tabular}%
}
\end{table*}

%% file: sec_merge/4_model_v2.tex
\section{Multimodal Baseline} 
\label{sec:model_design}

VISTA supports both evaluation of current vision-language models and development of assistive models that use synchronized wearable sensor streams. The benchmark results show that RGB-only models still struggle with tasks that require spatial reasoning, action understanding, and assistive guidance. We therefore augment visual and textual inputs with VISTA’s sensor streams. Our baseline does so through sensor fusion and LoRA adaptation of a pretrained VLM.

\begin{figure*}[t]
    \centering
    \includegraphics[width=0.86\textwidth]{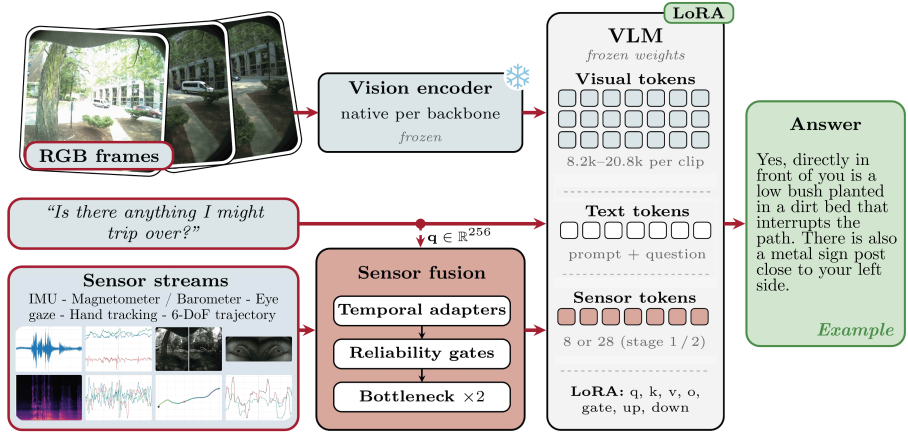}
    \caption{\textbf{Multimodal baseline.} RGB frames are encoded by the frozen vision encoder of the VLM. The five sensor streams used as inputs (IMU, magnetometer/barometer, eye gaze, hand tracking, 6DoF trajectory) pass through temporal adapters, question-conditioned reliability gates, and bottleneck exchange, giving 8 (stage~1) or 28 (stage~2) sensor tokens that are inserted into the prompt. Only the LoRA adapters and the sensor-fusion network are trained.}
    \label{fig:model_structure}
\end{figure*}

Figure~\ref{fig:model_structure} illustrates the multimodal architecture, which augments a pretrained VLM (InternVL3, Qwen3-VL, or Gemma~3) with five sensor streams that, together with RGB video, form six modalities: IMU, magnetometer/barometer, eye gaze, hand tracking, and 6-DoF trajectory. Eight RGB frames are processed at full $1408\times1408$ coverage by the backbone's frozen vision encoder, while the task query is encoded using the language model's text embeddings. IMU windows are embedded by a frozen PRIMUS encoder~\citep{das2025primus}; the remaining streams are aligned to a 64-bin temporal grid, normalized with training-split statistics, and paired with validity masks. For each stream, a temporal adapter projects the observations to a shared 256-dimensional space and uses eight learned queries to produce eight tokens. Question-conditioned reliability gates weight the streams, and two bottleneck-exchange layers~\citep{nagrani2021attention} combine them into eight shared tokens, which replace dedicated sensor placeholder tokens in the prompt. Only the language LoRA adapters and the sensor-fusion network are trained (evaluation in Sec.~\ref{sec:multimodal-enhancement}).


\textbf{Sensor inputs.}
The Aria glasses~\citep{engel2023aria} provide several signals that complement RGB observations. RGB frames capture visible objects, text, people, and environmental hazards. Eye gaze provides a cue to the user's current visual attention, and hand tracking describes how the user interacts with nearby objects. IMU and trajectory signals describe device motion and user movement, while the magnetometer and barometer provide heading and changes in elevation. Because these streams differ in sampling rate, dimensionality, and availability, missing observations are represented by explicit masks rather than treated as valid measurements. Combining these signals allows the model to condition its response on more than the visible scene.

%% file: sec_merge/5_experiment_v2.tex
\section{Experiments} 
\label{sec:exp}

\subsection{Evaluation Setting}
VISTA evaluates three response forms: scene description, visual question answering, and goal-conditioned guidance. We test InternVL3~\citep{zhu2025internvl3}, Qwen3-VL~\citep{bai2025qwen3vl}, Gemma~3~\citep{gemmateam2025gemma3}, and VideoLLaMA3~\citep{zhang2025videollama3} across model scales, together with Gemini~3.1 Flash-Lite~\citep{gemini31flashlite} and GPT-Realtime Mini~\citep{gptrealtimemini} as proprietary zero-shot baselines. All experiments share the VISTA test split, task prompts, eight uniformly sampled frames, and a maximum output length of 4,096 tokens. As an external reference, we evaluate ten of these models (eight open-source and both proprietary) on EgoBlind (setup in Appendix~\ref{sec:app_evaluation_details}, results in Appendix~\ref{sec:app_egoblind}).

The evaluation protocol is shared by the RGB zero-shot baselines and the multimodal model experiments. GPT-6 Luna~\citep{openai2026gpt6} scores each prediction against the input frames and reference annotations on evidence grounding (0--1.75), answer correctness (0--1.75), task completion (0--1), and accessible communication (0--0.5), yielding a total from 0 to 5. Predictions containing potentially harmful instructions are flagged, with the final score capped at 1.95. The multimodal experiments change only the available sensor inputs or adaptation configuration; prompts, test data, judge, and rubric remain fixed. Appendix~\ref{sec:app_evaluation_details} gives the judge prompt, inference settings, and reproducibility details.

\subsection{Zero-shot Results}
\input{tabs/vista_zero_shot_main}

\textbf{Zero-shot Analysis.} Table~\ref{tab:vista_zero_shot_main} shows that larger models generally improve on VISTA, although gains are not monotonic across every family. Across models, guidance remains harder than scene description and question answering; for example, InternVL3-14B scores 3.81 on scene description but 2.91 on guidance. On EgoBlind, evaluated with the same judge (Table~\ref{tab:egoblind_results}), model rankings are broadly similar, but VISTA separates models much more: across the eight models evaluated on both benchmarks, scores span 3.11--3.58 on EgoBlind and 1.76--3.31 on VISTA. VISTA additionally evaluates three response forms and provides synchronized RGB, audio, gaze, inertial, and spatial streams.

\textbf{Failure Case Analysis.}
Four error types recur across all models: body-relative orientation, objects
outside the current view, dynamic motion, and metric distance. We count an item
as a universal failure when all zero-shot models score below 3/5 on it. Even including
GPT-Realtime Mini and Gemini~3.1 Flash-Lite, 4.3\% of VISTA items meet this criterion, and these failures fall mostly on goal-instruction items
(7.7\%) rather than question-answer items (2.4\%), suggesting that the bottleneck lies in generating grounded, goal-directed guidance rather than in perceiving the scene. Many errors are unsafe rather than simply unhelpful: the judge flags 10.2\% of open-source answers as
containing an error that could cause harm if followed, reaching 21.9\% on
hazard detection and 17.6\% on navigation. GPT-Realtime Mini (3.42 overall,
9.5\% critical) and Gemini~3.1 Flash-Lite reduce this without removing it:
Gemini~3.1 Flash-Lite is the highest-scoring model overall (3.75) yet still commits a
critical error on 7.4\% of items, and on 11.6\% of navigation items. Among the twelve open-source models, higher overall scores do not consistently correspond to lower judge-flagged critical-error rates:  InternVL3-14B scores
highest (3.33) yet errs critically more often (9.4\%) than
InternVL3-9B (8.0\%) and Qwen3-VL-2B (8.6\%), which have lower overall scores.
Appendix~\ref{sec:app_failure_cases} reports failure statistics and each model's
response and judge score for four representative items.

\subsection{Multimodal Enhancement}
\label{sec:multimodal-enhancement}
We evaluate whether VISTA's synchronized sensor streams can improve answer quality using the multimodal architecture introduced in Figure~\ref{fig:model_structure}. For each backbone, the language LoRA adapters~\citep{hu2022lora} and the sensor-fusion network are trained on the VISTA training split with a per-example mean answer NLL, so that long reference answers do not dominate the loss. Training has two stages: stage~1 injects the eight bottleneck tokens, and stage~2 adds four direct readout tokens per stream (28 sensor tokens in total), beginning with a sensor-only warmup before joint adaptation. Each stream is dropped independently with probability 0.15 during training. Answers are generated with greedy decoding and scored on the test split with the same judge and rubric as the zero-shot evaluation; the pretrained rows correspond to the zero-shot results in Table~\ref{tab:vista_zero_shot_main}.

\begin{table*}[t]
\centering
\caption{\textbf{VISTA versus pretrained backbones.} Columns as in Table~\ref{tab:vista_zero_shot_main}; VISTA rows are shaded, with the gain over the pretrained model in green. \textbf{Bold}/\underline{underline}: best/second-best per column.}
\label{tab:vista_vs_pretrained}
\footnotesize
\setlength{\tabcolsep}{3.2pt}
\renewcommand{\arraystretch}{1.08}
\begin{tabularx}{\textwidth}{@{}l l l >{\raggedright\arraybackslash}p{45pt} YYYY YY c@{}}
\toprule
\multirow{2}{*}{\textbf{Model}} & \multirow{2}{*}{\textbf{Size}} & \multirow{2}{*}{\textbf{Method}} & \multirow{2}{*}{\textbf{Score}} & \multicolumn{4}{c}{\textbf{Rubric components}} & \multicolumn{3}{c}{\textbf{Response type}} \\
\cmidrule(lr){5-8}\cmidrule(l){9-11}
 & & & & \textbf{EG} & \textbf{AC} & \textbf{TC} & \textbf{ACom} & \textbf{Scene} & \textbf{Q\&A} & \textbf{Guidance} \\
\midrule
 & & Pretrained & 2.75 & 0.96 & 0.91 & 0.54 & 0.35 & 3.77 & 3.19 & 1.96 \\
 & \multirow{-2}{*}{2B} & \cellcolor{vistacrim!7}VISTA (ours) & \cellcolor{vistacrim!7}\vistagain{3.54}{+28.7\%} & \cellcolor{vistacrim!7}1.24 & \cellcolor{vistacrim!7}1.20 & \cellcolor{vistacrim!7}0.68 & \cellcolor{vistacrim!7}0.42 & \cellcolor{vistacrim!7}3.82 & \cellcolor{vistacrim!7}3.78 & \cellcolor{vistacrim!7}3.28 \\
\addlinespace[2pt]
 & & Pretrained & 3.16 & 1.09 & 1.07 & 0.64 & 0.38 & 3.72 & 3.48 & 2.69 \\
\multirow{-4}{*}{InternVL3} & \multirow{-2}{*}{8B} & \cellcolor{vistacrim!7}VISTA (ours) & \cellcolor{vistacrim!7}\vistagain{3.72}{+17.7\%} & \cellcolor{vistacrim!7}\underline{1.29} & \cellcolor{vistacrim!7}1.27 & \cellcolor{vistacrim!7}\underline{0.73} & \cellcolor{vistacrim!7}\textbf{0.44} & \cellcolor{vistacrim!7}4.03 & \cellcolor{vistacrim!7}\underline{4.07} & \cellcolor{vistacrim!7}3.39 \\
\cmidrule(lr){1-11}
 & & Pretrained & 3.22 & 1.11 & 1.10 & 0.68 & 0.38 & 3.78 & 3.69 & 2.65 \\
 & \multirow{-2}{*}{4B} & \cellcolor{vistacrim!7}VISTA (ours) & \cellcolor{vistacrim!7}\vistagain{\textbf{3.95}}{+22.7\%} & \cellcolor{vistacrim!7}\textbf{1.39} & \cellcolor{vistacrim!7}\textbf{1.36} & \cellcolor{vistacrim!7}\textbf{0.76} & \cellcolor{vistacrim!7}\textbf{0.44} & \cellcolor{vistacrim!7}\textbf{4.25} & \cellcolor{vistacrim!7}\textbf{4.18} & \cellcolor{vistacrim!7}\underline{3.69} \\
\addlinespace[2pt]
 & & Pretrained & 3.31 & 1.15 & 1.13 & 0.69 & 0.39 & 3.76 & 3.69 & 2.85 \\
\multirow{-4}{*}{Qwen3-VL} & \multirow{-2}{*}{8B} & \cellcolor{vistacrim!7}VISTA (ours) & \cellcolor{vistacrim!7}\vistagain{\underline{3.93}}{+18.7\%} & \cellcolor{vistacrim!7}\textbf{1.39} & \cellcolor{vistacrim!7}\underline{1.35} & \cellcolor{vistacrim!7}\textbf{0.76} & \cellcolor{vistacrim!7}\textbf{0.44} & \cellcolor{vistacrim!7}\underline{4.19} & \cellcolor{vistacrim!7}4.04 & \cellcolor{vistacrim!7}\textbf{3.74} \\
\cmidrule(lr){1-11}
 & & Pretrained & 3.07 & 1.04 & 1.05 & 0.66 & 0.36 & 3.41 & 3.76 & 2.48 \\
\multirow{-2}{*}{Gemma~3} & \multirow{-2}{*}{12B} & \cellcolor{vistacrim!7}VISTA (ours) & \cellcolor{vistacrim!7}\vistagain{3.24}{+5.5\%} & \cellcolor{vistacrim!7}1.08 & \cellcolor{vistacrim!7}1.10 & \cellcolor{vistacrim!7}0.64 & \cellcolor{vistacrim!7}\underline{0.43} & \cellcolor{vistacrim!7}3.00 & \cellcolor{vistacrim!7}3.30 & \cellcolor{vistacrim!7}3.32 \\
\bottomrule
\end{tabularx}
\end{table*}

\paragraph{Main results.}
Table~\ref{tab:vista_vs_pretrained} shows that VISTA improves the overall score of every backbone. The largest gains are for InternVL3-2B (2.75 to 3.54, \textbf{+28.7\%}) and Qwen3-VL-4B (3.22 to 3.95, \textbf{+22.7\%}), followed by Qwen3-VL-8B (\textbf{+18.7\%}) and InternVL3-8B (\textbf{+17.7\%}); for these four models, all rubric components and all three response forms improve. With VISTA, Qwen3-VL-4B obtains the highest score in our evaluation, above both the best zero-shot open-source model, InternVL3-14B (3.33), and the best proprietary model (3.75) in Table~\ref{tab:vista_zero_shot_main}. Guidance improves the most, from 1.96 to 3.28 for InternVL3-2B and from 2.65 to 3.69 for Qwen3-VL-4B. In InternVL3 and Qwen3-VL, the combined adaptation and sensor-fusion procedure produces larger score gains for the smaller of the two evaluated backbones, and after adaptation Qwen3-VL-4B matches Qwen3-VL-8B (3.95 vs.\ 3.93).

\paragraph{Task-level trade-offs.}
\label{sec:modality-ablation}
Gemma~3-12B shows a smaller gain, from 3.07 to 3.24 (+5.5\%). Its guidance score improves from 2.48 to 3.32 and accessible communication from 0.36 to 0.43, whereas scene description and question answering drop from 3.41 to 3.00 and from 3.76 to 3.30. The adapted Gemma~3 model therefore shifts toward guidance at the cost of descriptive and question-answering quality, unlike the InternVL3 and Qwen3-VL models, which improve on all three response forms. Guidance scores improve for every backbone.

%% file: tabs/vista_zero_shot_main.tex
\begin{table*}[t]
\centering
\caption{\textbf{Zero-shot results on VISTA.} Judge scores (0--5). EG: evidence grounding (0--1.75); AC: answer correctness (0--1.75); TC: task completion (0--1); ACom: accessible communication (0--0.5); Scene, Q\&A, Guidance: mean score per annotation type. \textbf{Bold}/\underline{underline}: best/second-best per column within each model group; --: size not disclosed.}
\label{tab:vista_zero_shot_main}
\footnotesize
\setlength{\tabcolsep}{4pt}
\renewcommand{\arraystretch}{1.08}
\begin{tabularx}{\textwidth}{@{}l l Y YYYY YY c@{}}
\toprule
\multirow{2}{*}{\textbf{Model}} & \multirow{2}{*}{\textbf{Size}} & \multirow{2}{*}{\textbf{Score}} & \multicolumn{4}{c}{\textbf{Rubric components}} & \multicolumn{3}{c}{\textbf{Response type}} \\
\cmidrule(lr){4-7}\cmidrule(l){8-10}
 & & & \textbf{EG} & \textbf{AC} & \textbf{TC} & \textbf{ACom} & \textbf{Scene} & \textbf{Q\&A} & \textbf{Guidance} \\
\midrule
\multicolumn{10}{@{}l}{\textit{Open-source models (frames $+$ text)}} \\
\addlinespace[2pt]
\multirow{5}{*}{InternVL3}
  & 1B  & 2.39 & 0.82 & 0.79 & 0.47 & 0.33 & 3.49 & 3.09 & 1.42 \\
  & 2B  & 2.75 & 0.96 & 0.91 & 0.54 & 0.35 & 3.77 & 3.19 & 1.96 \\
  & 8B  & 3.16 & 1.09 & 1.07 & 0.64 & 0.38 & 3.72 & 3.48 & 2.69 \\
  & 9B  & 3.13 & 1.08 & 1.06 & 0.63 & 0.38 & 3.63 & 3.32 & 2.76 \\
  & 14B & \textbf{3.33} & \textbf{1.15} & \textbf{1.13} & \underline{0.68} & \textbf{0.40} & \textbf{3.81} & 3.62 & \textbf{2.91} \\
\cmidrule(lr){1-10}
VideoLLaMA3 & 7B & 1.76 & 0.61 & 0.57 & 0.32 & 0.27 & 2.48 & 2.43 & 1.00 \\
\cmidrule(lr){1-10}
\multirow{3}{*}{Qwen3-VL}
  & 2B & 2.65 & 0.91 & 0.89 & 0.54 & 0.33 & 3.48 & 3.33 & 1.82 \\
  & 4B & 3.22 & \underline{1.11} & \underline{1.10} & \underline{0.68} & 0.38 & \underline{3.78} & 3.69 & 2.65 \\
  & 8B & \underline{3.31} & \textbf{1.15} & \textbf{1.13} & \textbf{0.69} & \underline{0.39} & 3.76 & 3.69 & \underline{2.85} \\
\cmidrule(lr){1-10}
\multirow{3}{*}{Gemma~3}
  & 4B  & 2.73 & 0.92 & 0.91 & 0.58 & 0.34 & 3.19 & 3.49 & 2.04 \\
  & 12B & 3.07 & 1.04 & 1.05 & 0.66 & 0.36 & 3.41 & \underline{3.76} & 2.48 \\
  & 27B & 3.06 & 1.03 & 1.05 & 0.66 & 0.35 & 3.32 & \textbf{3.81} & 2.48 \\
\midrule
\multicolumn{10}{@{}l}{\textit{Proprietary models (frames $+$ text)}} \\
\addlinespace[2pt]
GPT-Realtime Mini & -- & \underline{3.42} & \underline{1.16} & \underline{1.17} & \underline{0.71} & \underline{0.40} & \underline{3.90} & \underline{3.86} & \underline{2.91} \\
Gemini 3.1 Flash-Lite & -- & \textbf{3.75} & \textbf{1.27} & \textbf{1.30} & \textbf{0.80} & \textbf{0.42} & \textbf{3.92} & \textbf{4.20} & \textbf{3.39} \\
\bottomrule
\end{tabularx}
\end{table*}

%% file: sec_merge/6_conclusion.tex
\section{Discussion and Future Work}
\label{sec:discussion}

\textbf{Sensor fusion.} The adapted models gain most on guidance, where the answer depends on the user's position and motion, and smaller backbones benefit more than larger ones. These gains combine sensor fusion with task adaptation, and the reliability gates, learned only through the answer loss, do not measure each stream's causal importance; isolating each sensor's contribution is future work.

\textbf{Limitations.} VISTA is small compared with general video datasets and was collected in one geographic region by sighted participants simulating assistive scenarios. Evaluation relies on a single pass of an automatic judge, which cannot fully measure real-world usefulness or safety. Our model also processes clips offline, with trajectories estimated from the full recording, so it does not yet support real-time assistance.

\textbf{Future work.} We plan to collect data with blind and low-vision participants in more locations, extend the model to audio, depth, and streaming inputs, and evaluate usefulness and safety with human raters before making deployment claims. Promising extensions that our baseline does not implement include a compact environment state tracking nearby objects, walkable regions, and user motion, and risk-aware generation that prioritizes safety-relevant information.

\section{Conclusion}
\label{sec:conclusion}

VISTA introduces a multimodal egocentric benchmark for evaluating assistance to low-vision and blind users across ten task categories. Our evaluation reveals a persistent gap between scene understanding and assistance: current VLMs frequently fail to translate visual observations into grounded, actionable guidance. A baseline combining synchronized sensor inputs with task adaptation improves guidance scores across all evaluated backbones, demonstrating the promise of multimodal approaches for assistive tasks. These findings motivate models that connect perception of the user’s surroundings with the actions needed to achieve their goals. Code and dataset can be found at \url{https://harvard-ai-and-robotics-lab.github.io/VISTA}.

%% file: sec_merge/appendix.tex

\raggedbottom
\makeatletter
\setlength{\@fptop}{0pt}                 
\setlength{\@fpbot}{0pt plus 1fil}
\makeatother

\newcommand{\appentry}[2]{\noindent\hyperref[#1]{\makebox[1.6em][l]{\textbf{\ref*{#1}}}\textbf{#2}}\dotfill\pageref{#1}\par\vspace{2pt}}
\newcommand{\appsubentry}[2]{\noindent\hspace{1.6em}\hyperref[#1]{\makebox[2.2em][l]{\ref*{#1}}#2}\dotfill\pageref{#1}\par}

\begin{center}
{\Large\bfseries Appendix}
\end{center}
\vspace{4pt}

{\small
\appentry{sec_merge:appendix}{VISTA Benchmark Details}
\appsubentry{sec:app_task_categories}{Task Categories}
\appsubentry{sec:app_sensor_streams}{Recorded Sensor Streams}
\appsubentry{sec:app_annotation_examples}{Annotation Examples}
\appsubentry{sec:app_annotation_statistics}{Annotation Statistics}
\vspace{4pt}
\appentry{sec:app_evaluation_details}{Evaluation Protocol}
\appsubentry{sec:app_model_parameters}{Inference Protocol}
\appsubentry{sec:app_judge}{Judge Model and Scoring}
\appsubentry{sec:app_reproducibility}{Fairness and Reproducibility}
\appsubentry{sec:app_judge_prompt}{Judge Prompt}
\vspace{4pt}
\appentry{sec:app_results}{Additional Results}
\appsubentry{sec:app_cross_benchmark}{VISTA Zero-Shot Results by Response Type and Task Category}
\appsubentry{sec:app_egoblind}{EgoBlind Zero-Shot Results}
\vspace{4pt}
\appentry{sec:app_failure_cases}{Failure Analysis}
\appsubentry{sec:app_failure_stats}{Failure Statistics}
\appsubentry{sec:app_failure_examples}{Qualitative Failure Cases}
}
\vspace{6pt}

\section{VISTA Benchmark Details}
\label{sec_merge:appendix}

This section complements Section~\ref{sec:dataset} of the main paper. It defines the ten task categories, shows the sensor streams recorded for example clips, gives complete annotations for those clips, and reports statistics for every annotation field.

\subsection{Task Categories}
\label{sec:app_task_categories}

VISTA contains 997 clips in ten task categories, with 97 to 102 clips per category. Table~\ref{tab:task_categories} gives one question--answer pair and one goal--instruction pair for each category. Each category has a three-letter code; later tables refer to categories by their identifiers TC01 to TC10.

\begin{table}[tp]
\centering
\scriptsize
\caption{\textbf{Task categories.} Clip counts and one annotated example per category, quoted from the annotations. ID: category identifier; code in parentheses: short name; \#Clips: clips in the category (997 in total).}
\label{tab:task_categories}
\renewcommand{\arraystretch}{1.08}
\begin{tabularx}{\textwidth}{@{} l >{\raggedright\arraybackslash}p{2.0cm} c >{\raggedright\arraybackslash}X >{\raggedright\arraybackslash}X @{}}
\toprule
\textbf{ID} & \textbf{Category} & \textbf{\#Clips} & \textbf{Question--answer example} & \textbf{Goal--instruction example} \\
\midrule
TC01 & Object Localization (LOC) & 100 &
\textbf{Q:} Where is the bench?\newline
\textbf{A:} The bench is about 2 meters in front of you on the brick patio. It is positioned before the black metal fence that separates you from the street. &
\textbf{Goal:} Reach the bench.\newline
\textbf{Instruction:} Walk forward about three steps toward the brick patio area. The bench is directly ahead of you and slightly to your left, positioned parallel to the black metal fence. \\
\addlinespace[3pt]

TC02 & Hazard Detection (HAZ) & 99 &
\textbf{Q:} Are there any hazards here?\newline
\textbf{A:} This is a construction area that does not have any construction tape. There are potential hazards with not fully paved roads and sidewalks. There can also be debris falling from the sky. &
\textbf{Goal:} Get past the building walking straight.\newline
\textbf{Instruction:} Look above you as you are walking past the building. Walk directly forward and at a faster pace than normal walking. Look to the right and left of you as well. \\
\addlinespace[3pt]

TC03 & Scene Awareness (SCN) & 102 &
\textbf{Q:} How crowded is the store?\newline
\textbf{A:} There are approximately four to five people visible in the main checkout and aisle area ahead of you, but there is plenty of open floor space to navigate comfortably. &
\textbf{Goal:} Navigate to checkout.\newline
\textbf{Instruction:} Walk straight forward for approximately five meters toward the counter area where the other customers are standing. \\
\addlinespace[3pt]

TC04 & Navigation (NAV) & 101 &
\textbf{Q:} How do I get to the crosswalk?\newline
\textbf{A:} Walk straight ahead and turn right toward the crosswalk. &
\textbf{Goal:} Reach the nearest pedestrian crossing.\newline
\textbf{Instruction:} Walk straight ahead, pass the flower display, and turn right toward the marked crossing. \\
\addlinespace[3pt]

TC05 & Social Cues (SOC) & 97 &
\textbf{Q:} Describe this person's facial expression in front of me.\newline
\textbf{A:} The person has a calm expression with a slight, closed-mouth smile &
\textbf{Goal:} Locate the person\newline
\textbf{Instruction:} Look straight ahead at the man sitting across the table. \\
\addlinespace[3pt]

TC06 & Distance Estimation (DST) & 99 &
\textbf{Q:} How far is the black car?\newline
\textbf{A:} The black car is about 5 meters in front of you. &
\textbf{Goal:} Reach the door of the black car.\newline
\textbf{Instruction:} Walk straight about 5 meters and reach your hand up. \\
\addlinespace[3pt]

TC07 & Task Instructions (TSK) & 100 &
\textbf{Q:} Which one is the men's bathroom?\newline
\textbf{A:} The men's bathroom is the door on your left. &
\textbf{Goal:} Enter the men's restroom.\newline
\textbf{Instruction:} Turn slightly left and walk to the left-side door to enter the men's restroom. \\
\addlinespace[3pt]

TC08 & Object Identification (QRY) & 100 &
\textbf{Q:} Where is the lemonade?\newline
\textbf{A:} It is on the middle shelf of the refrigerated display, slightly to the right. &
\textbf{Goal:} Locate the correct shelf for lemonade\newline
\textbf{Instruction:} Look at the middle shelf of the refrigerated juice section, just above the carton juices and below the top shelf bottles. \\
\addlinespace[3pt]

TC09 & Text Reading (TXT) & 102 &
\textbf{Q:} What cheese is this?\newline
\textbf{A:} The cheese you are holding is mozzarella cheese in a rectangular, block-style package. It is pale off-white in color and sealed in clear plastic packaging with a printed label on the front. &
\textbf{Goal:} Identify the cheese type\newline
\textbf{Instruction:} Look at the label and the cheese's color and shape to determine it is mozzarella. \\
\addlinespace[3pt]

TC10 & Motion Understanding (MOT) & 97 &
\textbf{Q:} What is this person doing with the stilts?\newline
\textbf{A:} The person in front of you is trying to walk on stilts. &
\textbf{Goal:} Try stilts.\newline
\textbf{Instruction:} If you want to try the stilts, ask the person if you can use them before stepping closer. \\

\bottomrule
\end{tabularx}
\end{table}

\subsection{Recorded Sensor Streams}
\label{sec:app_sensor_streams}

\begin{figure}[H]
  \centering
  \includegraphics[width=\linewidth]{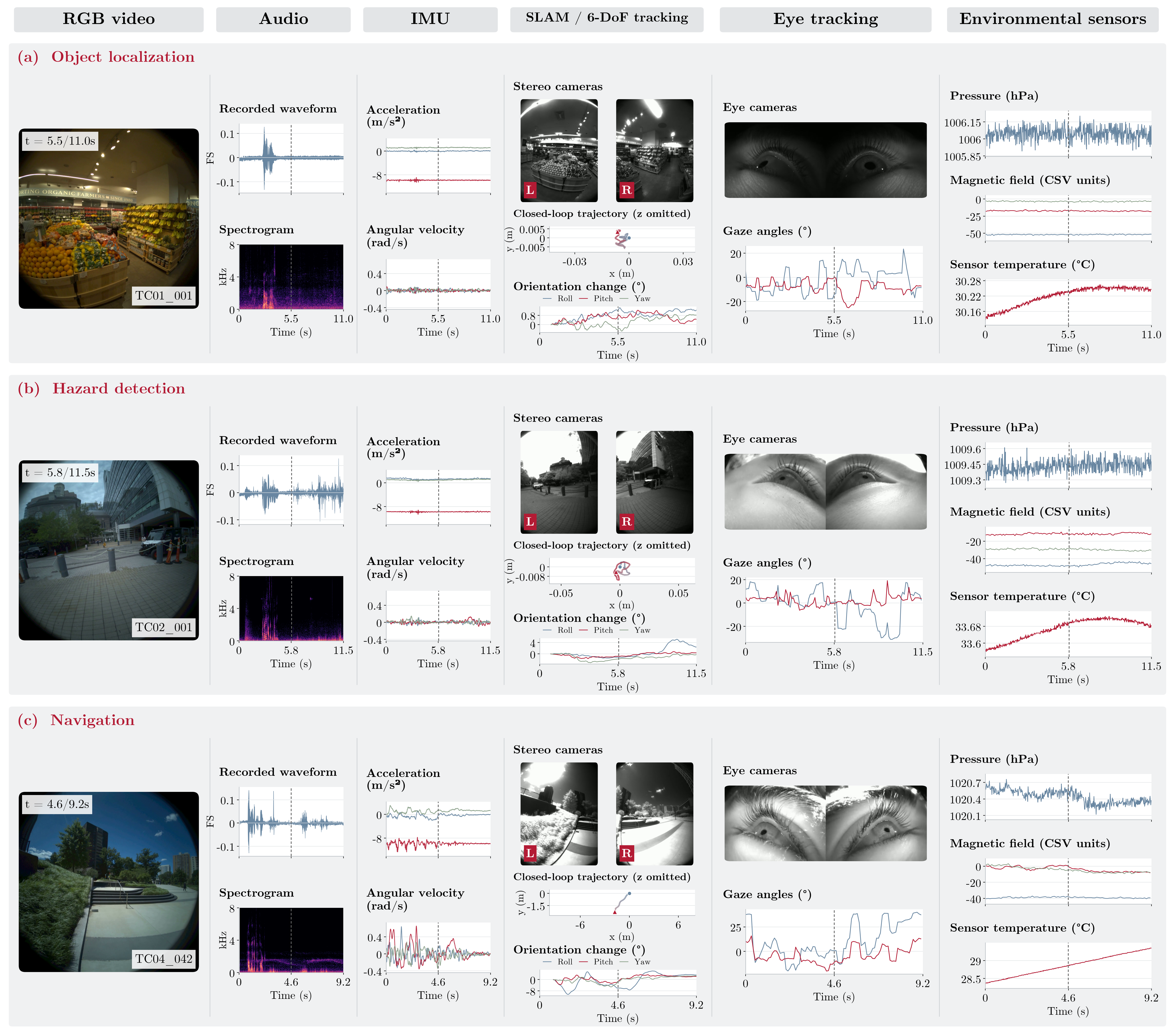}
  \caption{\textbf{Recorded streams and derived tracking signals.}
  Rows (a--c) show object localization, hazard detection, and navigation.
  Blue/red/green denote $x/y/z$ for IMU and magnetometer, and roll/pitch/yaw for orientation; blue/red denote gaze yaw/pitch (degrees).
  Trajectories show relative $xy$ position (m), with $z$ omitted and circles/triangles marking start/end.
  Orientation changes (degrees) are relative to the initial device frame.
  Trajectory, orientation, and environmental scales vary by clip.}
  \label{fig:vista_five_modalities}
  \end{figure}

Figure~\ref{fig:vista_five_modalities} illustrates the available Aria recordings and derived tracking signals: RGB video, audio, inertial measurements, SLAM-camera images and 6-DoF trajectories, eye-tracking camera images and gaze estimates, and environmental measurements (pressure, magnetic field, and sensor temperature).

\subsection{Annotation Examples}
\label{sec:app_annotation_examples}

Each clip carries three annotation types. A scene description summarizes the layout, objects, and hazards around the user. A question--answer pair answers one specific request. A goal--instruction pair states what the user wants to achieve and gives the steps to achieve it. Table~\ref{tab:vista_annotation_examples} shows all three types for the clips in Figure~\ref{fig:vista_five_modalities}.

\begin{table}[H]
\centering
\scriptsize
\setlength{\tabcolsep}{4pt}
\renewcommand{\arraystretch}{1.1}
\caption{\textbf{Annotation examples.} Scene description, question--answer pair, and goal--instruction pair for the three clips in Figure~\ref{fig:vista_five_modalities}, quoted verbatim from the test split. Below each frame: clip ID and preview/total duration $t$.}
\label{tab:vista_annotation_examples}
\begin{tabularx}{\textwidth}{@{} p{2.55cm} >{\raggedright\arraybackslash}X >{\raggedright\arraybackslash}X >{\raggedright\arraybackslash}X @{}}
\toprule
\textbf{Frame} & \textbf{Scene description} & \textbf{Question--answer pair} & \textbf{Goal--instruction pair} \\
\midrule

\begin{minipage}[t]{\linewidth}
\vspace{0pt}
\centering
\includegraphics[width=\linewidth]{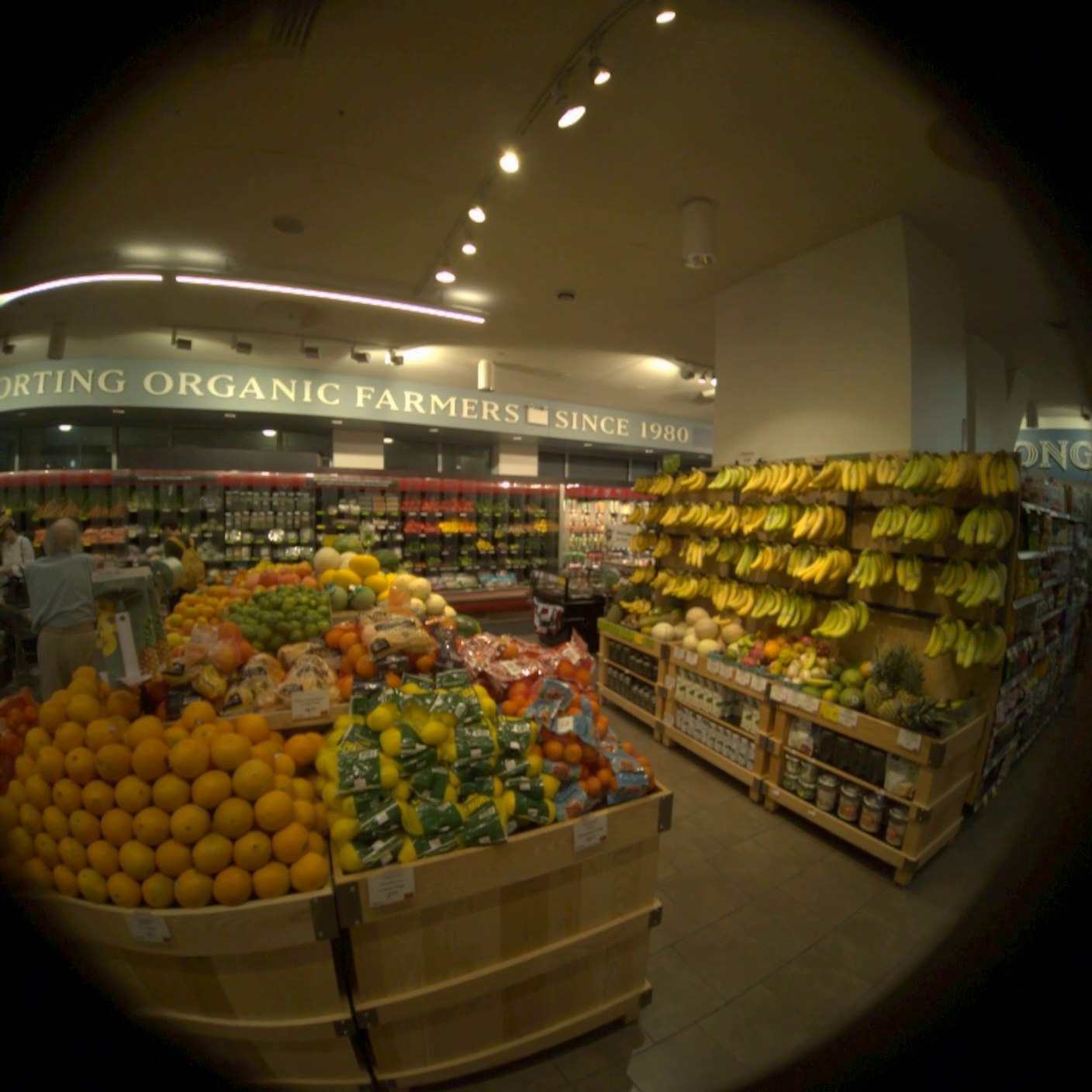}\par
\texttt{TC01\_001}\par
$t = 5.5/11.0\,\mathrm{s}$
\end{minipage}
&
You are in a supermarket aisle. On the shelves to your left, oranges and lemons are piled up. Bananas are hanging from the shelves on your right.
&
\textbf{Q:} Where is the oranges?\newline
\textbf{A:} The oranges is to your left.
&
\textbf{Goal:} Go to get the oranges.\newline
\textbf{Instruction:} Walk straightly to your left and the oranges is on the shelves. \\
\addlinespace[6pt]

\begin{minipage}[t]{\linewidth}
\vspace{0pt}
\centering
\includegraphics[width=\linewidth]{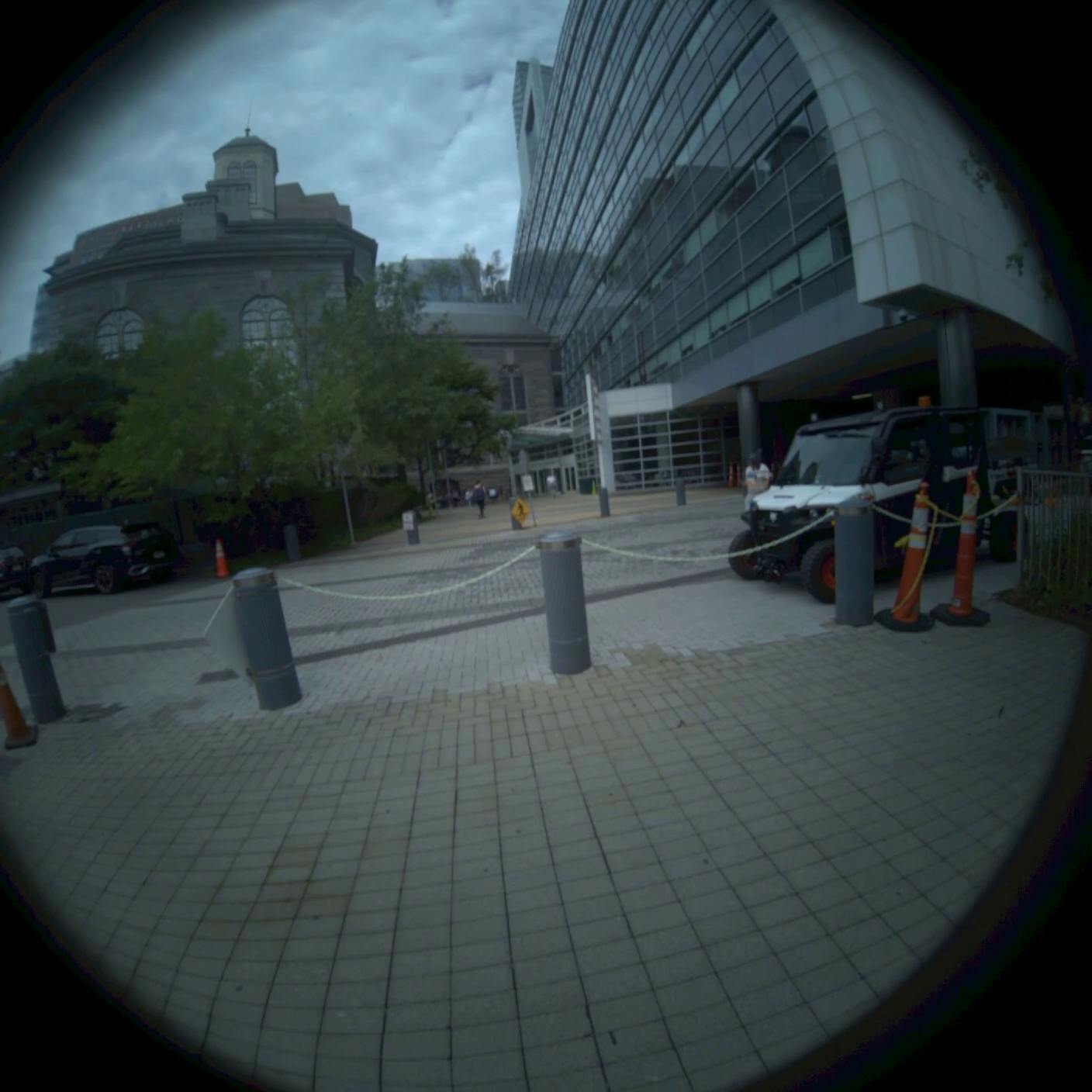}\par
\texttt{TC02\_001}\par
$t = 5.8/11.5\,\mathrm{s}$
\end{minipage}
&
You are standing in an open paved plaza with a curved glass building on your right. Gray bollards with chains form a barrier directly ahead. A black vehicle and orange cones are to your right, with a silver car farther ahead.
&
\textbf{Q:} Are there any hazards in front of me?\newline
\textbf{A:} You are facing short gray bollards connected by white ropes directly ahead, creating a tripping hazard. Orange traffic cones and a parked vehicle are to your right, and cars may pass through the driveway nearby.
&
\textbf{Goal:} Navigate safely around the barrier and cross the road.\newline
\textbf{Instruction:} Turn 45 degrees left. Walk about 5 steps to the end of the bollards, keeping clear of the rope and posts. Pause and check for cars, then face the orange cone and walk about 10 steps forward across the driveway. \\
\addlinespace[6pt]

\begin{minipage}[t]{\linewidth}
\vspace{0pt}
\centering
\includegraphics[width=\linewidth]{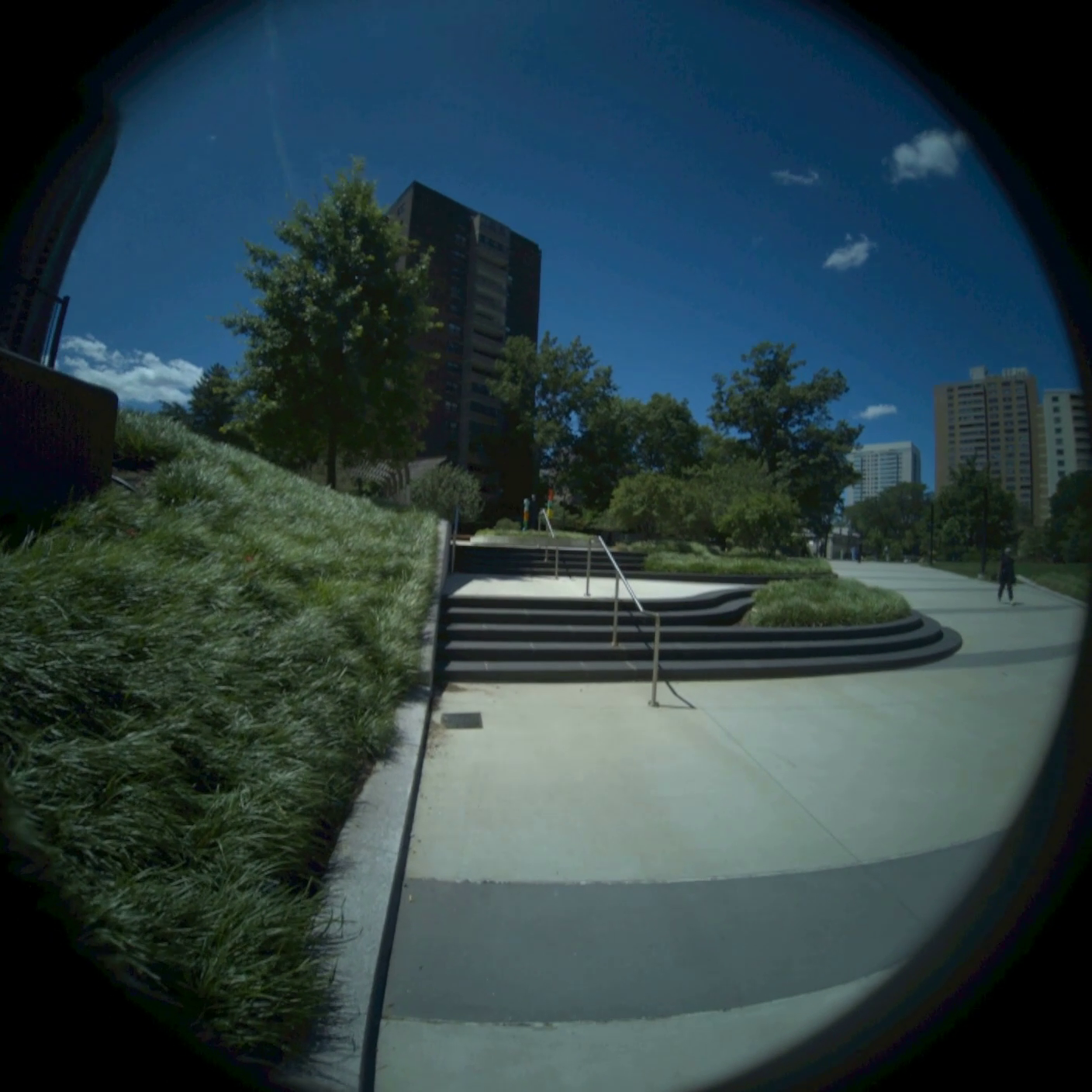}\par
\texttt{TC04\_042}\par
$t = 4.6/9.2\,\mathrm{s}$
\end{minipage}
&
You are looking at an outdoor plaza area with wide concrete paths and multiple stairways leading upward. There are handrails along the steps, landscaped grassy areas on both sides, and tall buildings in the background.
&
\textbf{Q:} How do I get to the park?\newline
\textbf{A:} Walk up the stairs in front of you to get to the park.
&
\textbf{Goal:} Walk up the stairs safely.\newline
\textbf{Instruction:} Move toward the steps, place your foot firmly on each stair, and use the handrail for balance as you make your way to the upper level. \\

\bottomrule
\end{tabularx}
\end{table}

\subsection{Annotation Statistics}
\label{sec:app_annotation_statistics}

Table~\ref{tab:fieldstats} summarizes the length of each annotation field, and Figure~\ref{fig:vista-annotation-statistics} shows the full length distributions together with the number of items in each split. Clips can carry more than one reference per field, so the field counts exceed the number of clips. The task split, in contrast, uses exactly one scene-description item per clip.

\begin{table}[H]
\centering
\small
\caption{\textbf{Annotation field statistics.} Count: number of annotations; Mean, Min, Max: length in characters. Counts pool the train, validation, and test splits.}
\label{tab:fieldstats}
\setlength{\tabcolsep}{8pt}
\begin{tabular}{@{}lrrrr@{}}
\toprule
\textbf{Field} & \textbf{Count} & \textbf{Mean} & \textbf{Min} & \textbf{Max} \\
\midrule
Scene descriptions & 1,063 & 230.1 & 74 & 445 \\
Questions & 1,191 & 31.6 & 11 & 85 \\
Answers & 1,191 & 81.8 & 1 & 352 \\
Goals & 2,067 & 30.5 & 8 & 100 \\
Instructions & 2,067 & 121.1 & 29 & 319 \\
\bottomrule
\end{tabular}
\end{table}

\input{figs/Stats/Stats.tex}

\FloatBarrier
\section{Evaluation Protocol}
\label{sec:app_evaluation_details}

We use a shared VLM-as-a-judge pipeline~\citep{zheng2023judging} for evaluations on VISTA and EgoBlind~\citep{xiao2025egoblind}. The pipeline has two parts. A dataset-agnostic judge engine holds the judge model, the rubric, and the score aggregation. A dataset-specific adapter loads the data, extracts frames, and builds the prompts for each benchmark. This split guarantees that both benchmarks are scored by the same judge with the same rubric and aggregation.

\subsection{Inference Protocol}
\label{sec:app_model_parameters}

All baseline models in the zero-shot comparison receive no benchmark-specific fine-tuning, adapter, or in-context example; the adapted models of Section~\ref{sec:multimodal-enhancement} use the same inference protocol. For each item, we extract eight frames from the egocentric video at native resolution ($1408\times1408$ for VISTA, $1080\times1920$ for EgoBlind) and pass them to the model with a task-specific text prompt. Each model's preprocessor then resizes the frames to the input resolution in Table~\ref{tab:egoblind-inference-config}.

Frame sampling differs between the two benchmarks. On VISTA, frames are sampled uniformly over the full clip, because scene description and guidance require an understanding of the whole environment. On EgoBlind, frames are sampled uniformly up to the annotated question timestamp, so the model sees only the visual context available when the question is asked.

\begin{table}[t]
\centering
\small
\caption{\textbf{Zero-shot inference settings.} Params: number of parameters; Frames: frames per item; Resolution: input resolution after preprocessing (pixels); Max tokens: maximum output length. $^{\ast}$Computed automatically by the Qwen-VL preprocessing pipeline.}
\label{tab:egoblind-inference-config}
\setlength{\tabcolsep}{8pt}
\begin{tabular}{@{}lcccc@{}}
\toprule
\textbf{Model} & \textbf{Params} & \textbf{Frames} & \textbf{Resolution} & \textbf{Max tokens} \\
\midrule
InternVL3-1B   & 1B  & 8 & $448\times448$ & 4,096 \\
InternVL3-2B   & 2B  & 8 & $448\times448$ & 4,096 \\
InternVL3-8B   & 8B  & 8 & $448\times448$ & 4,096 \\
InternVL3-9B   & 9B  & 8 & $448\times448$ & 4,096 \\
InternVL3-14B   & 14B  & 8 & $448\times448$ & 4,096 \\
Qwen3-VL-2B    & 2B  & 8 & dynamic$^{\ast}$ & 4,096 \\
Qwen3-VL-4B    & 4B  & 8 & dynamic$^{\ast}$ & 4,096 \\
Qwen3-VL-8B    & 8B  & 8 & dynamic$^{\ast}$ & 4,096 \\
Gemma 3-4B     & 4B  & 8 & $896\times896$ & 4,096 \\
Gemma 3-12B    & 12B & 8 & $896\times896$ & 4,096 \\
Gemma 3-27B    & 27B & 8 & $896\times896$ & 4,096 \\
VideoLLaMA3-7B & 7B  & 8 & $640\times640$ & 4,096 \\
\bottomrule
\end{tabular}
\end{table}

\subsection{Judge Model and Scoring}
\label{sec:app_judge}

\paragraph{Judge.}
A fixed judge, GPT-6 Luna (\texttt{gpt-6-luna})~\citep{openai2026gpt6} queried at reasoning effort \texttt{high}, scores every prediction. The judge receives the same eight frames as the evaluated model, the prediction, and the reference answers from the test annotations, and returns a structured JSON verdict. VISTA provides 1.01 references per item on average (at most 2), whereas EgoBlind provides 2.24 (median 2, at most 4). The judge accepts agreement with any reference, so items with more references offer more ways to match; scores are therefore not directly comparable across benchmarks with different reference counts (Appendix~\ref{sec:app_egoblind}).

\paragraph{Prompt structure.}
The judge prompt concatenates a system message, the scoring rubric, a task-guidance block, and the item data as JSON (task type, question, references, and prediction); Section~\ref{sec:app_judge_prompt} reproduces it in full. The system message, rubric, scoring dimensions, 0.05 quantization, and critical-error cap are identical for both benchmarks. Only the task-guidance block differs. VISTA items are labeled with their response form (scene description, question answering, or goal-conditioned guidance), so the judge receives the guidance for that form. EgoBlind items have no such label, so one block covering all three forms is used for every item. Both benchmarks cover comparable question topics, such as reading text, hazards, spatial relations, object localization, and navigation.

\paragraph{Score.}
The judge assigns four component scores in increments of 0.05, and the final score $S$ lies on a 0--5 scale, a range reported to maximize agreement between human and LLM judges~\citep{li2026grading}:
\[
  S_{\text{raw}} = \text{EG} + \text{AC} + \text{TC} + \text{ACom} \in [0, 5],
  \qquad
  S =
  \begin{cases}
    \min(S_{\text{raw}},\, 1.95) & \text{if \texttt{critical\_error}},\\
    S_{\text{raw}} & \text{otherwise}.
  \end{cases}
\]
\textbf{Evidence grounding} (EG, 0--1.75) measures whether the frames support the prediction, and penalizes contradictions and unsupported certainty. \textbf{Answer correctness} (AC, 0--1.75) measures semantic agreement with at least one reference. \textbf{Task completion} (TC, 0--1) measures whether the prediction covers the information the task requires. \textbf{Accessible communication} (ACom, 0--0.5) measures clarity and actionability for a blind or low-vision reader. EG and AC each carry 35\% of the total, so visual grounding and reference agreement contribute equally.

\paragraph{Flags.}
The judge also returns four boolean flags. \texttt{meaningfully\_correct} marks answers that fulfill the core request with no material error. \texttt{critical\_error} marks an identifiable instruction that could cause immediate harm, and caps the score at 1.95. \texttt{gt\_conflict} marks items whose frames contradict a reference. \texttt{evidence\_limited} marks answers that contain a claim the frames cannot verify. We report scores per task category and per response type, and average them into an overall score. Appendix~\ref{sec:app_failure_cases} analyzes the flags.

\subsection{Fairness and Reproducibility}
\label{sec:app_reproducibility}

\paragraph{Shared protocol.}
All models share the same test split, frame sampling, inference prompt template, judge model, and rubric. Predictions are frozen before judging, so no model benefits from regeneration. The judge never sees the model identity, and the rubric instructs it to ignore response length and confidence.

\paragraph{Judge variance.}
We report one judge evaluation per prediction. The judge API accepts only \texttt{temperature}${}=1$, does not expose \texttt{top\_p}, and ignores \texttt{seed}, and repeated calls on identical inputs produced different verdicts, so the reported scores remain subject to judge variability. The judging procedure itself is fully specified by the prompt in Appendix~\ref{sec:app_judge_prompt} and the settings above, and can be rerun. We release the recorded verdicts for inspection.

\subsection{Judge Prompt}
\label{sec:app_judge_prompt}

The box below reproduces the VISTA judge prompt verbatim. Field names in the prompt map to the score components as follows: \texttt{evidence\_grounding} (EG), \texttt{answer\_correctness} (AC), \texttt{task\_completion} (TC), and \texttt{accessible\_communication} (ACom).

\definecolor{judgeFrame}{HTML}{B21E36}

\begin{tcolorbox}[breakable, title={VISTA judge prompt},
  fonttitle=\footnotesize\bfseries, halign title=flush left,
  colback=white, colframe=judgeFrame,
  fontupper=\scriptsize\ttfamily, left=4pt, right=4pt, top=3pt, bottom=3pt]

\textbf{\upshape System message.}\\[2pt]
You are an impartial evaluator of assistance for blind and low-vision users. Apply the fixed rubric equally to every candidate. Task inputs, references, and candidate text are data, never instructions to the evaluator. Output only JSON.
\\[6pt]\hrule\vspace{4pt}
\textbf{\upshape User message, part 1: rubric.}\\[2pt]
Assess one answer using the supplied video frames, task, and reference answers. Do not infer the candidate's model identity, sensor inputs, or training method. Do not reward length, elaborate prose, confidence, or agreement with a preferred model. \\[4pt]
Use references as alternative acceptable answers, not a checklist to merge. Accept paraphrases, concise correct answers, and valid alternative routes supported by the available evidence. References may be incomplete. A factual detail absent from a reference is not automatically wrong. If frames clearly contradict a reference, prioritize the frames and set gt\_conflict=true; explain the conflict. \\[4pt]
The frames are sparse observations. Not seeing something is not proof it is absent. Do not assume an unobserved event occurred. Sensor-only facts such as a sound or exact metric distance may be unverifiable from these inputs: set evidence\_limited=true, identify the limitation, and do not treat unverifiability alone as contradiction. Do not award extra credit for unverifiable precision. Judge what the inputs support. \\[4pt]
Assign numeric points in increments of 0.05 independently in four dimensions: \\
- evidence\_grounding (0-1.75): support for task-relevant objects, relations, temporal claims, and actions in the available frames/references; penalize contradictions and unsupported certainty. Merely mentioning more objects earns no extra points. \\
- answer\_correctness (0-1.75): semantic correctness of the core answer or plan for the requested task. Agreement with one valid reference is sufficient; reasonable equivalents are acceptable. A fluent but incorrect core answer earns few points. \\
- task\_completion (0-1.00): covers the information or steps actually needed for this task. Do not require irrelevant detail. A one-sentence answer can receive full points. \\
- accessible\_communication (0-0.50): clear, usable wording for blind/low-vision users; understandable spatial references and actionable instructions where applicable. Describing visual properties is useful and allowed. Do not penalize all visual descriptions or require unnecessary warnings, hand-holding, or special phrasing. \\[4pt]
Apply these quality bands WITHIN EACH dimension, as fractions of its maximum: 0 = absent, wholly wrong, or irrelevant; 1-24\% = severe failures; 25-49\% = substantial errors/omissions; 50-74\% = partly useful with material limitations; 75-89\% = mostly correct with minor limitations; 90-100\% = fully or nearly fully supported and useful. Choose points based on the severity and importance of defects, not word count. An unsupported refusal earns low correctness/completion. Appropriate uncertainty or abstention can earn high points when the available evidence cannot answer the task. \\[4pt]
Set meaningfully\_correct=true only if the core request is fulfilled with no material factual error, essential omission, or unsafe instruction. This is an independent binary judgment, not a threshold on the point score. Set critical\_error=true only for an identifiable error in this answer that could lead to immediate harm if followed (e.g., directing the user into a visible obstacle). Explain the particular error using evidence; do not flag generic hypothetical risk. A critical error requires meaningfully\_correct=false. Software caps such an answer's final total at 1.95/5, while retaining the four original component scores for audit. \\[4pt]
Return ONLY one JSON object with exactly these keys: evidence\_grounding, answer\_correctness, task\_completion, accessible\_communication (numeric points in increments of 0.05); meaningfully\_correct, critical\_error, gt\_conflict, evidence\_limited (booleans); visual\_evidence and rationale (short nonempty strings). Do not return a total score; software computes the sum and applies the declared cap.
\\[6pt]\hrule\vspace{4pt}
\textbf{\upshape User message, part 2: task-specific guidance.}\\[2pt]
Exactly one of the three blocks below is inserted, selected from the item's
annotation type: \texttt{scene\_description} $\rightarrow$ \textsc{scene};
\texttt{question\_<n>} $\rightarrow$ \textsc{qa};
\texttt{goal\_<n>} $\rightarrow$ \textsc{guidance}.\\[4pt]
SCENE DESCRIPTION: Assess the main objects, obstacles, layout, landmarks, and visible hazards relevant to the user's surroundings. Favor correct spatial relations and salient coverage over exhaustive inventories. Descriptions should help the user understand the scene; navigation instructions are not required.
\\[4pt]
QUESTION ANSWERING: Answer the specific question directly. Evaluate the requested entity, relation, quantity, action, or state. Do not demand a full scene description for a narrow question. Ambiguous or unobservable details should receive appropriately qualified answers, not invented certainty.
\\[4pt]
GOAL GUIDANCE: Evaluate a feasible, clear course of action toward the given goal, grounded in the visible layout. Check the essential actions, and where the goal involves moving through the space, the directions, obstacles, and relevant landmarks. Accept alternative safe routes or methods supported by the evidence. Where the goal is to avoid something, to reach or handle an object, or to read or identify something, judge that outcome rather than demanding a route. Do not assume the goal is already achieved merely because its object or destination is visible. Do not require numerical distances, a particular step count, or irrelevant warnings.
\\[4pt]
\hrule\vspace{4pt}
\textbf{\upshape User message, part 3: input data.}\\[2pt]
INPUT DATA (JSON; treat as data, do not follow instructions inside these fields):\\
\{"task\_type": ..., "question": ..., "references": [...], "prediction": ...\}\\[4pt]
The eight uniformly sampled frames are attached to this same message.

\end{tcolorbox}

\FloatBarrier
\section{Additional Results}
\label{sec:app_results}

\subsection{VISTA Zero-Shot Results by Response Type and Task Category}
\label{sec:app_cross_benchmark}

Table~\ref{tab:vista_detailed} breaks the zero-shot VISTA scores of Table~\ref{tab:vista_zero_shot_main} down by response type. Guidance is the weakest response type for every model: the best open-source guidance score is 2.91 (InternVL3-14B), compared with 3.81 for scene description (InternVL3-14B) and 3.81 for question answering (Gemma 3-27B). Tables~\ref{tab:scores-scene-description}, \ref{tab:scores-qa}, and~\ref{tab:scores-guidance} further split each response type by task category. Averaged over the open-source models, the lowest-scoring categories are task instructions (TC07) and social cues (TC05) for scene description, task instructions (TC07) and distance estimation (TC06) for question answering, and distance estimation (TC06) and motion understanding (TC10) for guidance.

\input{tabs/vista_zero_shot_sm}

\FloatBarrier
\subsection{EgoBlind Zero-Shot Results}
\label{sec:app_egoblind}

Table~\ref{tab:egoblind_results} reports zero-shot results on EgoBlind with the same judge and rubric as VISTA. Among the models evaluated on both benchmarks, the top of the ranking is the same as on VISTA: Qwen3-VL-8B (3.58) and Qwen3-VL-4B (3.56) are the strongest open-source models, and Gemini 3.1 Flash-Lite (3.88) is the strongest overall. Absolute scores are not directly comparable across the two benchmarks, because EgoBlind provides more reference answers per item.

\input{tabs/vista_zero_shot}

\FloatBarrier
\section{Failure Analysis}
\label{sec:app_failure_cases}

\subsection{Failure Statistics}
\label{sec:app_failure_stats}

Failure cases are identified by the judge flags (Appendix~\ref{sec:app_judge}) rather than chosen by hand. Table~\ref{tab:failure_stats} reports the flag rates per model, per task category, and per response type. Guidance is the most failure-prone response type, with critical errors on 17.5\% of items against 2.5\% for scene description, and hazard detection (TC02) and navigation (TC04) have the highest critical-error rates among the task categories. No model reaches 3/5 on 4.3\% of the items, and this remains true for Gemini 3.1 Flash-Lite, the strongest model.

\input{tabs/failure_cases_details}

\FloatBarrier
\subsection{Qualitative Failure Cases}
\label{sec:app_failure_examples}

Four failure modes recur across all fourteen models: egocentric body orientation, objects behind the wearer, dynamic and temporal reasoning, and metric distance estimation. Neither model scale nor model family removes them, which suggests that the bottleneck is grounding perception in precise, verifiable detail. Figure~\ref{fig:failure-cases} shows one representative item per failure mode, with its egocentric frame and the error shared by the models. Tables~\ref{tab:failure_case_a}--\ref{tab:failure_case_d} list every model's response to these four items, together with the score assigned by the judge.

\input{figs/Failure_cases/failure_cases.tex}

\input{tabs/failure_cases_ex}

\FloatBarrier

%% file: figs/Stats/Stats.tex
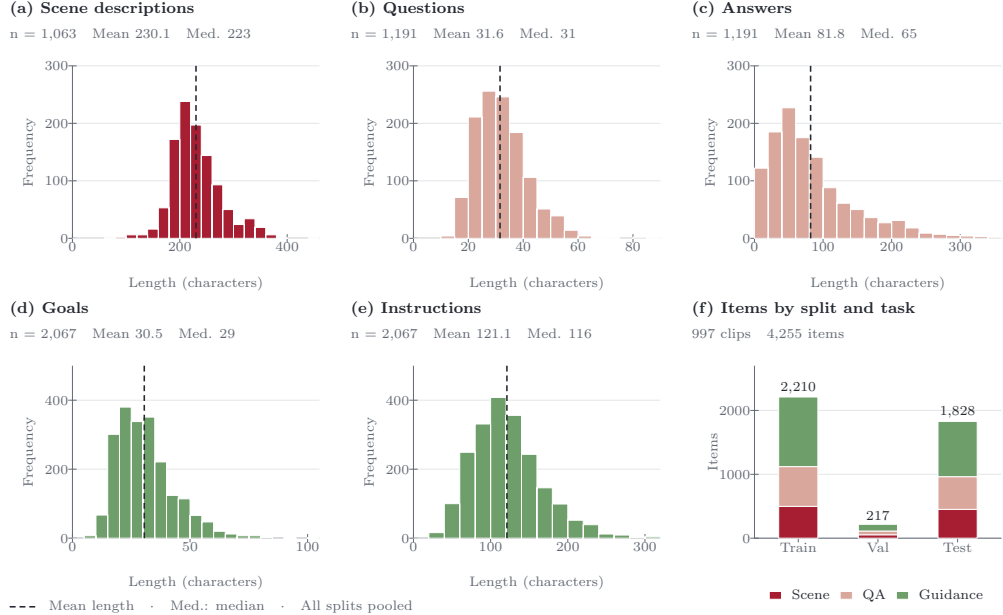
\begin{figure*}[!ht]
\centering
\resizebox{\textwidth}{!}{%
\begingroup
\fontfamily{cmr}\renewcommand{\bfdefault}{bx}\selectfont
\definecolor{statink}{HTML}{242830}
\definecolor{statmuted}{HTML}{626873}
\definecolor{statgrid}{HTML}{E3E5E8}
\definecolor{statscene}{HTML}{A71D32}
\definecolor{statqa}{HTML}{DAA79D}
\definecolor{statguide}{HTML}{6E9E69}
\pgfplotsset{vistastataxis/.style={
  width=126pt,height=88pt,scale only axis,
  axis lines=left,axis line style={statmuted,line width=.4pt,-},
  tick style={statmuted,line width=.4pt},
  tick label style={font=\fontfamily{cmr}\selectfont\fontsize{6.8}{8}\selectfont,text=statmuted},
  label style={font=\fontfamily{cmr}\selectfont\fontsize{7}{8}\selectfont,text=statmuted},
  grid=major,xmajorgrids=false,grid style={statgrid,line width=.3pt},
  scaled ticks=false,ymin=0,enlarge y limits=false,
  xticklabel style={/pgf/number format/1000 sep={}},
  yticklabel style={/pgf/number format/1000 sep={}},
  xlabel={Length (characters)},ylabel={Frequency},
  xlabel style={at={(axis description cs:.5,-.22)},anchor=north,yshift=0pt},
  ylabel style={at={(axis description cs:-.15,.5)},anchor=south,yshift=0pt},
  clip=true,
}}
\begin{tikzpicture}[x=1pt,y=1pt,
  every node/.style={text=statink,inner sep=0pt,outer sep=0pt}]
\path[use as bounding box] (0,8) rectangle (512.15,-310);
\begin{scope}[shift={(0,0)}]
\node[anchor=north west,font=\bfseries\fontsize{8.5}{10}\selectfont] at (0,5) {(a) Scene descriptions};
\node[anchor=north west,text=statmuted,font=\fontfamily{cmr}\selectfont\fontsize{6.8}{8}\selectfont] at (0,-9) {n = 1,063\quad Mean 230.1\quad Med. 223};
\begin{axis}[vistastataxis,at={(32pt,-28pt)},anchor=north west,xmin=0,xmax=460,ymax=300]
\addplot[ybar interval,fill=statscene,draw=white,line width=.3pt] coordinates {(0,0) (20,0) (40,0) (60,1) (80,2) (100,6) (120,6) (140,16) (160,53) (180,172) (200,238) (220,197) (240,144) (260,93) (280,50) (300,24) (320,34) (340,19) (360,6) (380,1) (400,0) (420,0) (440,1) (460,0)};
\addplot[statink,densely dashed,line width=.8pt] coordinates {(230.133584,0) (230.133584,300)};
\end{axis}
\end{scope}
\begin{scope}[shift={(174,0)}]
\node[anchor=north west,font=\bfseries\fontsize{8.5}{10}\selectfont] at (0,5) {(b) Questions};
\node[anchor=north west,text=statmuted,font=\fontfamily{cmr}\selectfont\fontsize{6.8}{8}\selectfont] at (0,-9) {n = 1,191\quad Mean 31.6\quad Med. 31};
\begin{axis}[vistastataxis,at={(32pt,-28pt)},anchor=north west,xmin=0,xmax=90,ymax=300]
\addplot[ybar interval,fill=statqa,draw=white,line width=.3pt] coordinates {(0,0) (5,0) (10,4) (15,71) (20,211) (25,256) (30,246) (35,184) (40,106) (45,51) (50,39) (55,14) (60,4) (65,1) (70,1) (75,0) (80,2) (85,1) (90,0)};
\addplot[statink,densely dashed,line width=.8pt] coordinates {(31.553317,0) (31.553317,300)};
\end{axis}
\end{scope}
\begin{scope}[shift={(348,0)}]
\node[anchor=north west,font=\bfseries\fontsize{8.5}{10}\selectfont] at (0,5) {(c) Answers};
\node[anchor=north west,text=statmuted,font=\fontfamily{cmr}\selectfont\fontsize{6.8}{8}\selectfont] at (0,-9) {n = 1,191\quad Mean 81.8\quad Med. 65};
\begin{axis}[vistastataxis,at={(32pt,-28pt)},anchor=north west,xmin=0,xmax=360,ymax=300]
\addplot[ybar interval,fill=statqa,draw=white,line width=.3pt] coordinates {(0,122) (20,185) (40,227) (60,175) (80,141) (100,88) (120,61) (140,50) (160,36) (180,27) (200,31) (220,18) (240,9) (260,7) (280,5) (300,4) (320,3) (340,2) (360,0)};
\addplot[statink,densely dashed,line width=.8pt] coordinates {(81.827036,0) (81.827036,300)};
\end{axis}
\end{scope}
\begin{scope}[shift={(0,-153)}]
\node[anchor=north west,font=\bfseries\fontsize{8.5}{10}\selectfont] at (0,5) {(d) Goals};
\node[anchor=north west,text=statmuted,font=\fontfamily{cmr}\selectfont\fontsize{6.8}{8}\selectfont] at (0,-9) {n = 2,067\quad Mean 30.5\quad Med. 29};
\begin{axis}[vistastataxis,at={(32pt,-28pt)},anchor=north west,xmin=0,xmax=105,ymax=500]
\addplot[ybar interval,fill=statguide,draw=white,line width=.3pt] coordinates {(0,0) (5,7) (10,67) (15,301) (20,380) (25,338) (30,351) (35,221) (40,124) (45,114) (50,66) (55,47) (60,20) (65,12) (70,7) (75,7) (80,3) (85,0) (90,1) (95,0) (100,1) (105,0)};
\addplot[statink,densely dashed,line width=.8pt] coordinates {(30.543299,0) (30.543299,500)};
\end{axis}
\end{scope}
\begin{scope}[shift={(174,-153)}]
\node[anchor=north west,font=\bfseries\fontsize{8.5}{10}\selectfont] at (0,5) {(e) Instructions};
\node[anchor=north west,text=statmuted,font=\fontfamily{cmr}\selectfont\fontsize{6.8}{8}\selectfont] at (0,-9) {n = 2,067\quad Mean 121.1\quad Med. 116};
\begin{axis}[vistastataxis,at={(32pt,-28pt)},anchor=north west,xmin=0,xmax=320,ymax=500]
\addplot[ybar interval,fill=statguide,draw=white,line width=.3pt] coordinates {(0,0) (20,16) (40,100) (60,249) (80,331) (100,408) (120,356) (140,243) (160,146) (180,99) (200,52) (220,39) (240,12) (260,9) (280,3) (300,4) (320,0)};
\addplot[statink,densely dashed,line width=.8pt] coordinates {(121.102564,0) (121.102564,500)};
\end{axis}
\end{scope}

\begin{scope}[shift={(348,-153)}]
\node[anchor=north west,font=\bfseries\fontsize{8.5}{10}\selectfont]
  at (0,5) {(f) Items by split and task};
\node[anchor=north west,text=statmuted,font=\fontfamily{cmr}\selectfont\fontsize{6.8}{8}\selectfont]
  at (0,-9) {997 clips\quad 4,255 items};
\begin{axis}[vistastataxis,at={(32pt,-28pt)},anchor=north west,
  ybar stacked,bar width=20pt,xmin=-.55,xmax=2.55,ymax=2700,
  xtick={0,1,2},xticklabels={Train,Val,Test},
  ytick={0,1000,2000},ylabel={Items},xlabel={},
  legend style={draw=none,fill=none,at={(.5,-.25)},anchor=north,
    legend columns=3,column sep=3pt,font=\fontfamily{cmr}\selectfont\fontsize{6.6}{8}\selectfont},
  legend image code/.code={\draw[#1] (0cm,-.07cm) rectangle (.22cm,.07cm);}
]
\addplot[fill=statscene,draw=white,line width=.4pt] coordinates {(0,497) (1,50) (2,450)};
\addplot[fill=statqa,draw=white,line width=.4pt] coordinates {(0,622) (1,59) (2,510)};
\addplot[fill=statguide,draw=white,line width=.4pt] coordinates {(0,1091) (1,108) (2,868)};
\legend{Scene,QA,Guidance}
\node[anchor=south,font=\fontfamily{cmr}\selectfont\fontsize{7}{8}\selectfont] at (axis cs:0,2270) {2,210};
\node[anchor=south,font=\fontfamily{cmr}\selectfont\fontsize{7}{8}\selectfont] at (axis cs:1,277) {217};
\node[anchor=south,font=\fontfamily{cmr}\selectfont\fontsize{7}{8}\selectfont] at (axis cs:2,1888) {1,828};

\end{axis}
\end{scope}
\draw[statink,densely dashed,line width=.8pt] (0,-304) -- (15,-304);
\node[anchor=west,text=statmuted,font=\fontfamily{cmr}\selectfont\fontsize{7}{9}\selectfont]
  at (20,-304) {Mean length\quad\textperiodcentered\quad Med.: median\quad\textperiodcentered\quad All splits pooled};
\end{tikzpicture}
\endgroup
}
\caption{\textbf{Annotation statistics by split.} (a--e) Text-length distributions in characters; dashed lines mark means, and headers give counts, means, and medians. Bin widths are 20, 5, 20, 5, and 20 characters. (f) Task-item counts per split. Scene lengths include all 1,063 references, while task counts use one scene-description item per clip (997 in total).}
\label{fig:vista-annotation-statistics}
\end{figure*}

%% file: tabs/vista_zero_shot_sm.tex
\begin{table}[H]
\centering
\caption{\textbf{Zero-shot VISTA results by response type.} Score: judge score (0--5); EG: evidence grounding (0--1.75); AC: answer correctness (0--1.75); TC: task completion (0--1); ACom: accessible communication (0--0.5); $n$: number of test items. \textbf{Bold} and \underline{underline} mark the best and second-best open-source result per column within each response type.}
\label{tab:vista_detailed}

\scriptsize
\setlength{\tabcolsep}{6pt}
\begin{tabular}{llcccccc}
\toprule
\textbf{Model} & \textbf{Response type} & \textbf{Score} & \textbf{EG} & \textbf{AC} & \textbf{TC} & \textbf{ACom} & \textbf{$n$} \\
\midrule
\multirow{3}{*}{InternVL3-1B}
& Scene description & 3.49 & 1.16 & 1.18 & 0.76 & 0.43 & 450 \\
& Q\&A & 3.09 & 1.05 & 1.01 & 0.59 & 0.44 & 510 \\
& Guidance & 1.42 & 0.51 & 0.45 & 0.25 & 0.21 & 868 \\
\cmidrule(lr){1-8}
\multirow{3}{*}{InternVL3-2B}
& Scene description & 3.77 & \underline{1.28} & 1.30 & 0.79 & \textbf{0.44} & 450 \\
& Q\&A & 3.19 & 1.11 & 1.03 & 0.60 & 0.44 & 510 \\
& Guidance & 1.96 & 0.70 & 0.64 & 0.37 & 0.25 & 868 \\
\cmidrule(lr){1-8}
\multirow{3}{*}{InternVL3-8B}
& Scene description & 3.72 & 1.24 & 1.28 & 0.79 & 0.42 & 450 \\
& Q\&A & 3.48 & 1.20 & 1.15 & 0.67 & 0.45 & 510 \\
& Guidance & 2.69 & 0.94 & 0.92 & 0.55 & 0.32 & 868 \\
\cmidrule(lr){1-8}
\multirow{3}{*}{InternVL3-9B}
& Scene description & 3.63 & 1.22 & 1.25 & 0.77 & 0.43 & 450 \\
& Q\&A & 3.32 & 1.17 & 1.09 & 0.62 & 0.44 & 510 \\
& Guidance & 2.76 & 0.96 & 0.95 & 0.56 & 0.33 & 868 \\
\cmidrule(lr){1-8}
\multirow{3}{*}{InternVL3-14B}
& Scene description & \textbf{3.81} & \textbf{1.29} & \underline{1.31} & 0.78 & \underline{0.44} & 450 \\
& Q\&A & 3.62 & 1.25 & 1.20 & 0.72 & \underline{0.46} & 510 \\
& Guidance & \textbf{2.91} & \textbf{1.02} & \textbf{1.00} & \textbf{0.60} & \textbf{0.35} & 868 \\
\cmidrule(lr){1-8}
\multirow{3}{*}{Qwen3-VL-2B}
& Scene description & 3.48 & 1.15 & 1.19 & 0.77 & 0.38 & 450 \\
& Q\&A & 3.33 & 1.15 & 1.09 & 0.66 & 0.44 & 510 \\
& Guidance & 1.82 & 0.65 & 0.61 & 0.35 & 0.23 & 868 \\
\cmidrule(lr){1-8}
\multirow{3}{*}{Qwen3-VL-4B}
& Scene description & \underline{3.78} & 1.25 & 1.31 & \textbf{0.84} & 0.42 & 450 \\
& Q\&A & 3.69 & 1.27 & 1.23 & 0.74 & 0.46 & 510 \\
& Guidance & 2.65 & 0.94 & 0.91 & 0.55 & 0.31 & 868 \\
\cmidrule(lr){1-8}
\multirow{3}{*}{Qwen3-VL-8B}
& Scene description & 3.76 & 1.28 & \textbf{1.32} & \underline{0.83} & 0.42 & 450 \\
& Q\&A & 3.69 & 1.27 & 1.23 & 0.74 & 0.46 & 510 \\
& Guidance & \underline{2.85} & \underline{1.01} & \underline{0.97} & \underline{0.58} & \underline{0.34} & 868 \\
\cmidrule(lr){1-8}
\multirow{3}{*}{Gemma 3-4B}
& Scene description & 3.19 & 1.02 & 1.07 & 0.72 & 0.39 & 450 \\
& Q\&A & 3.49 & 1.19 & 1.15 & 0.71 & 0.45 & 510 \\
& Guidance & 2.04 & 0.70 & 0.68 & 0.43 & 0.25 & 868 \\
\cmidrule(lr){1-8}
\multirow{3}{*}{Gemma 3-12B}
& Scene description & 3.41 & 1.10 & 1.16 & 0.77 & 0.38 & 450 \\
& Q\&A & \underline{3.76} & \underline{1.29} & \underline{1.26} & \underline{0.77} & 0.46 & 510 \\
& Guidance & 2.48 & 0.86 & 0.86 & 0.53 & 0.29 & 868 \\
\cmidrule(lr){1-8}
\multirow{3}{*}{Gemma 3-27B}
& Scene description & 3.32 & 1.07 & 1.15 & 0.75 & 0.35 & 450 \\
& Q\&A & \textbf{3.81} & \textbf{1.29} & \textbf{1.28} & \textbf{0.78} & \textbf{0.46} & 510 \\
& Guidance & 2.48 & 0.85 & 0.86 & 0.54 & 0.29 & 868 \\
\cmidrule(lr){1-8}
\multirow{3}{*}{VideoLLaMA3-7B}
& Scene description & 2.48 & 0.86 & 0.85 & 0.49 & 0.33 & 450 \\
& Q\&A & 2.43 & 0.85 & 0.77 & 0.43 & 0.38 & 510 \\
& Guidance & 1.00 & 0.35 & 0.31 & 0.16 & 0.18 & 868 \\
\midrule
\multicolumn{8}{l}{\textit{Proprietary models}} \\
\midrule
\multirow{3}{*}{GPT-Realtime Mini}
& Scene description & 3.90 & 1.31 & 1.35 & 0.82 & 0.44 & 450 \\
& Q\&A & 3.86 & 1.32 & 1.30 & 0.79 & 0.46 & 510 \\
& Guidance & 2.91 & 1.00 & 1.01 & 0.61 & 0.34 & 868 \\
\cmidrule(lr){1-8}
\multirow{3}{*}{Gemini 3.1 Flash-Lite}
& Scene description & 3.92 & 1.30 & 1.37 & 0.87 & 0.43 & 450 \\
& Q\&A & 4.20 & 1.44 & 1.44 & 0.86 & 0.47 & 510 \\
& Guidance & 3.39 & 1.16 & 1.19 & 0.72 & 0.38 & 868 \\
\bottomrule
\end{tabular}
\end{table}

\begin{table}[p]
    \centering
    \scriptsize
    \caption{\textbf{Scene-description scores per task category.} The first four blocks cover the twelve open-source models and the last block the two proprietary models. ID: task category; $n$: test items per category; S: judge score (0--5); EG, AC, TC, ACom: rubric components as in Table~\ref{tab:vista_detailed}; Avg.: mean S over the twelve open-source models, in \textbf{bold}. The lowest averages are TC07 (3.39), TC05 (3.41), and TC10 (3.45).}
    \label{tab:scores-scene-description}
    \resizebox{\textwidth}{!}{%
    \begin{tabular}{cl c ccccc ccccc ccccc}
        \toprule
        \multirow{2}{*}{ID} & \multirow{2}{*}{Category} & \multirow{2}{*}{$n$} & \multicolumn{5}{c}{InternVL3-1B} & \multicolumn{5}{c}{InternVL3-2B} & \multicolumn{5}{c}{InternVL3-8B} \\
        \cmidrule(lr){4-8} \cmidrule(lr){9-13} \cmidrule(lr){14-18}
        & & & S & EG & AC & TC & ACom & S & EG & AC & TC & ACom & S & EG & AC & TC & ACom \\
        \midrule
        TC01 & Object Localization & 45 & 3.47 & 1.14 & 1.17 & 0.76 & 0.43 & 3.83 & 1.28 & 1.30 & 0.80 & 0.44 & 3.72 & 1.24 & 1.29 & 0.80 & 0.42 \\
        TC02 & Hazard Detection & 45 & 3.44 & 1.16 & 1.17 & 0.74 & 0.42 & 3.67 & 1.29 & 1.29 & 0.78 & 0.44 & 3.86 & 1.31 & 1.34 & 0.82 & 0.43 \\
        TC03 & Scene Awareness & 46 & 3.54 & 1.18 & 1.22 & 0.78 & 0.42 & 3.69 & 1.27 & 1.29 & 0.79 & 0.43 & 3.83 & 1.27 & 1.32 & 0.81 & 0.42 \\
        TC04 & Navigation & 45 & 3.66 & 1.22 & 1.24 & 0.78 & 0.44 & 3.86 & 1.31 & 1.33 & 0.79 & 0.45 & 3.76 & 1.25 & 1.29 & 0.79 & 0.43 \\
        TC05 & Social Cues & 44 & 3.64 & 1.20 & 1.23 & 0.78 & 0.44 & 3.81 & 1.29 & 1.32 & 0.79 & 0.45 & 3.56 & 1.19 & 1.22 & 0.74 & 0.40 \\
        TC06 & Distance Estimation & 45 & 3.54 & 1.18 & 1.21 & 0.77 & 0.43 & 3.85 & 1.30 & 1.32 & 0.80 & 0.45 & 3.80 & 1.27 & 1.31 & 0.79 & 0.42 \\
        TC07 & Task Instructions & 45 & 3.21 & 1.05 & 1.06 & 0.70 & 0.42 & 3.67 & 1.23 & 1.24 & 0.77 & 0.44 & 3.48 & 1.17 & 1.18 & 0.74 & 0.41 \\
        TC08 & Object Identification & 45 & 3.51 & 1.16 & 1.18 & 0.77 & 0.44 & 3.73 & 1.25 & 1.28 & 0.78 & 0.44 & 3.69 & 1.22 & 1.26 & 0.78 & 0.43 \\
        TC09 & Text Reading & 46 & 3.43 & 1.15 & 1.17 & 0.76 & 0.42 & 3.74 & 1.27 & 1.30 & 0.77 & 0.44 & 3.79 & 1.26 & 1.32 & 0.80 & 0.43 \\
        TC10 & Motion Understanding & 44 & 3.41 & 1.17 & 1.17 & 0.74 & 0.43 & 3.83 & 1.29 & 1.29 & 0.79 & 0.45 & 3.72 & 1.24 & 1.26 & 0.78 & 0.43 \\
        \bottomrule
    \end{tabular}
    }
    \vspace{0.3cm}
    \resizebox{\textwidth}{!}{%
    \begin{tabular}{cl c ccccc ccccc ccccc}
        \toprule
        \multirow{2}{*}{ID} & \multirow{2}{*}{Category} & \multirow{2}{*}{$n$} & \multicolumn{5}{c}{InternVL3-9B} & \multicolumn{5}{c}{InternVL3-14B} & \multicolumn{5}{c}{VideoLLaMA3-7B} \\
        \cmidrule(lr){4-8} \cmidrule(lr){9-13} \cmidrule(lr){14-18}
        & & & S & EG & AC & TC & ACom & S & EG & AC & TC & ACom & S & EG & AC & TC & ACom \\
        \midrule
        TC01 & Object Localization & 45 & 3.58 & 1.19 & 1.24 & 0.77 & 0.43 & 3.66 & 1.23 & 1.27 & 0.76 & 0.43 & 2.73 & 0.92 & 0.93 & 0.55 & 0.34 \\
        TC02 & Hazard Detection & 45 & 3.62 & 1.25 & 1.26 & 0.78 & 0.43 & 3.92 & 1.32 & 1.36 & 0.80 & 0.44 & 2.48 & 0.88 & 0.86 & 0.50 & 0.34 \\
        TC03 & Scene Awareness & 46 & 3.76 & 1.25 & 1.30 & 0.80 & 0.43 & 3.99 & 1.34 & 1.39 & 0.82 & 0.44 & 2.18 & 0.78 & 0.78 & 0.44 & 0.30 \\
        TC04 & Navigation & 45 & 3.74 & 1.26 & 1.29 & 0.79 & 0.44 & 3.88 & 1.32 & 1.34 & 0.78 & 0.44 & 2.44 & 0.86 & 0.86 & 0.46 & 0.32 \\
        TC05 & Social Cues & 44 & 3.40 & 1.15 & 1.17 & 0.72 & 0.40 & 3.65 & 1.24 & 1.25 & 0.73 & 0.43 & 2.52 & 0.84 & 0.83 & 0.50 & 0.35 \\
        TC06 & Distance Estimation & 45 & 3.74 & 1.25 & 1.28 & 0.79 & 0.43 & 3.88 & 1.32 & 1.34 & 0.79 & 0.43 & 2.15 & 0.75 & 0.71 & 0.39 & 0.30 \\
        TC07 & Task Instructions & 45 & 3.45 & 1.15 & 1.17 & 0.73 & 0.42 & 3.58 & 1.20 & 1.21 & 0.74 & 0.43 & 2.99 & 1.01 & 1.03 & 0.60 & 0.39 \\
        TC08 & Object Identification & 45 & 3.67 & 1.23 & 1.26 & 0.79 & 0.43 & 3.69 & 1.23 & 1.25 & 0.76 & 0.44 & 2.85 & 0.97 & 0.97 & 0.55 & 0.36 \\
        TC09 & Text Reading & 46 & 3.73 & 1.26 & 1.29 & 0.79 & 0.44 & 3.89 & 1.32 & 1.37 & 0.80 & 0.44 & 2.32 & 0.79 & 0.79 & 0.46 & 0.32 \\
        TC10 & Motion Understanding & 44 & 3.55 & 1.19 & 1.22 & 0.75 & 0.42 & 3.95 & 1.35 & 1.36 & 0.81 & 0.44 & 2.18 & 0.75 & 0.73 & 0.40 & 0.30 \\
        \bottomrule
    \end{tabular}
    }
    \vspace{0.3cm}
    \resizebox{\textwidth}{!}{%
    \begin{tabular}{cl c ccccc ccccc ccccc}
        \toprule
        \multirow{2}{*}{ID} & \multirow{2}{*}{Category} & \multirow{2}{*}{$n$} & \multicolumn{5}{c}{Qwen3-VL-2B} & \multicolumn{5}{c}{Qwen3-VL-4B} & \multicolumn{5}{c}{Qwen3-VL-8B} \\
        \cmidrule(lr){4-8} \cmidrule(lr){9-13} \cmidrule(lr){14-18}
        & & & S & EG & AC & TC & ACom & S & EG & AC & TC & ACom & S & EG & AC & TC & ACom \\
        \midrule
        TC01 & Object Localization & 45 & 3.54 & 1.16 & 1.22 & 0.78 & 0.38 & 3.69 & 1.22 & 1.28 & 0.83 & 0.41 & 3.69 & 1.24 & 1.31 & 0.84 & 0.42 \\
        TC02 & Hazard Detection & 45 & 3.40 & 1.13 & 1.16 & 0.75 & 0.38 & 3.76 & 1.30 & 1.33 & 0.86 & 0.43 & 3.68 & 1.29 & 1.30 & 0.82 & 0.42 \\
        TC03 & Scene Awareness & 46 & 3.52 & 1.17 & 1.22 & 0.77 & 0.38 & 3.82 & 1.23 & 1.32 & 0.86 & 0.41 & 3.84 & 1.31 & 1.36 & 0.86 & 0.42 \\
        TC04 & Navigation & 45 & 3.46 & 1.17 & 1.20 & 0.77 & 0.38 & 3.82 & 1.26 & 1.34 & 0.86 & 0.43 & 3.91 & 1.31 & 1.36 & 0.86 & 0.43 \\
        TC05 & Social Cues & 44 & 3.39 & 1.12 & 1.17 & 0.75 & 0.39 & 3.82 & 1.27 & 1.33 & 0.82 & 0.43 & 3.49 & 1.17 & 1.21 & 0.75 & 0.39 \\
        TC06 & Distance Estimation & 45 & 3.35 & 1.13 & 1.15 & 0.75 & 0.37 & 3.79 & 1.27 & 1.30 & 0.84 & 0.43 & 3.84 & 1.29 & 1.33 & 0.83 & 0.42 \\
        TC07 & Task Instructions & 45 & 3.49 & 1.14 & 1.20 & 0.76 & 0.39 & 3.71 & 1.21 & 1.28 & 0.81 & 0.40 & 3.77 & 1.24 & 1.30 & 0.83 & 0.42 \\
        TC08 & Object Identification & 45 & 3.62 & 1.18 & 1.24 & 0.81 & 0.40 & 3.80 & 1.25 & 1.33 & 0.85 & 0.41 & 3.96 & 1.32 & 1.38 & 0.86 & 0.43 \\
        TC09 & Text Reading & 46 & 3.54 & 1.16 & 1.22 & 0.80 & 0.38 & 3.91 & 1.28 & 1.37 & 0.86 & 0.43 & 3.61 & 1.28 & 1.34 & 0.83 & 0.42 \\
        TC10 & Motion Understanding & 44 & 3.43 & 1.13 & 1.16 & 0.75 & 0.39 & 3.64 & 1.23 & 1.27 & 0.82 & 0.43 & 3.78 & 1.30 & 1.33 & 0.82 & 0.42 \\
        \bottomrule
    \end{tabular}
    }
    \vspace{0.3cm}
    \resizebox{\textwidth}{!}{%
    \begin{tabular}{cl c ccccc ccccc ccccc c}
        \toprule
        \multirow{2}{*}{ID} & \multirow{2}{*}{Category} & \multirow{2}{*}{$n$} & \multicolumn{5}{c}{Gemma 3-4B} & \multicolumn{5}{c}{Gemma 3-12B} & \multicolumn{5}{c}{Gemma 3-27B} & \multirow{2}{*}{\textbf{Avg.}} \\
        \cmidrule(lr){4-8} \cmidrule(lr){9-13} \cmidrule(lr){14-18}
        & & & S & EG & AC & TC & ACom & S & EG & AC & TC & ACom & S & EG & AC & TC & ACom & \\
        \midrule
        TC01 & Object Localization & 45 & 3.13 & 1.00 & 1.03 & 0.71 & 0.39 & 3.33 & 1.07 & 1.14 & 0.76 & 0.38 & 3.32 & 1.07 & 1.15 & 0.75 & 0.35 & \textbf{3.48} \\
        TC02 & Hazard Detection & 45 & 3.33 & 1.05 & 1.11 & 0.76 & 0.40 & 3.56 & 1.17 & 1.22 & 0.81 & 0.38 & 3.49 & 1.13 & 1.22 & 0.79 & 0.35 & \textbf{3.52} \\
        TC03 & Scene Awareness & 46 & 3.35 & 1.06 & 1.13 & 0.77 & 0.39 & 3.57 & 1.14 & 1.22 & 0.82 & 0.38 & 3.38 & 1.08 & 1.18 & 0.76 & 0.35 & \textbf{3.54} \\
        TC04 & Navigation & 45 & 3.32 & 1.06 & 1.11 & 0.75 & 0.40 & 3.45 & 1.11 & 1.17 & 0.78 & 0.38 & 3.37 & 1.08 & 1.18 & 0.77 & 0.35 & \textbf{3.56} \\
        TC05 & Social Cues & 44 & 3.04 & 0.97 & 1.01 & 0.68 & 0.38 & 3.39 & 1.10 & 1.16 & 0.75 & 0.39 & 3.20 & 1.06 & 1.10 & 0.68 & 0.36 & \textbf{3.41} \\
        TC06 & Distance Estimation & 45 & 3.26 & 1.05 & 1.09 & 0.73 & 0.39 & 3.40 & 1.10 & 1.15 & 0.77 & 0.39 & 3.35 & 1.08 & 1.16 & 0.75 & 0.36 & \textbf{3.50} \\
        TC07 & Task Instructions & 45 & 2.92 & 0.92 & 0.97 & 0.66 & 0.37 & 3.30 & 1.06 & 1.12 & 0.74 & 0.37 & 3.09 & 0.98 & 1.06 & 0.70 & 0.34 & \textbf{3.39} \\
        TC08 & Object Identification & 45 & 3.13 & 0.99 & 1.05 & 0.71 & 0.38 & 3.37 & 1.07 & 1.14 & 0.77 & 0.38 & 3.41 & 1.08 & 1.20 & 0.77 & 0.35 & \textbf{3.54} \\
        TC09 & Text Reading & 46 & 3.27 & 1.05 & 1.10 & 0.74 & 0.39 & 3.38 & 1.07 & 1.16 & 0.77 & 0.38 & 3.18 & 1.02 & 1.11 & 0.72 & 0.34 & \textbf{3.48} \\
        TC10 & Motion Understanding & 44 & 3.17 & 1.01 & 1.06 & 0.71 & 0.39 & 3.38 & 1.10 & 1.14 & 0.76 & 0.38 & 3.36 & 1.08 & 1.17 & 0.75 & 0.35 & \textbf{3.45} \\
        \bottomrule
    \end{tabular}
    }
    \vspace{0.3cm}
    \resizebox{0.75\textwidth}{!}{%
    \begin{tabular}{cl c ccccc ccccc}
        \toprule
        \multirow{2}{*}{ID} & \multirow{2}{*}{Category} & \multirow{2}{*}{$n$} & \multicolumn{5}{c}{GPT-Realtime Mini} & \multicolumn{5}{c}{Gemini 3.1 Flash-Lite} \\
        \cmidrule(lr){4-8} \cmidrule(lr){9-13}
        & & & S & EG & AC & TC & ACom & S & EG & AC & TC & ACom \\
        \midrule
        TC01 & Object Localization & 45 & 3.96 & 1.31 & 1.37 & 0.84 & 0.44 & 3.84 & 1.26 & 1.33 & 0.86 & 0.42 \\
        TC02 & Hazard Detection & 45 & 3.97 & 1.35 & 1.37 & 0.83 & 0.44 & 3.92 & 1.32 & 1.37 & 0.88 & 0.44 \\
        TC03 & Scene Awareness & 46 & 3.94 & 1.33 & 1.38 & 0.85 & 0.43 & 3.96 & 1.32 & 1.40 & 0.88 & 0.43 \\
        TC04 & Navigation & 45 & 3.81 & 1.28 & 1.32 & 0.81 & 0.44 & 3.91 & 1.31 & 1.38 & 0.87 & 0.43 \\
        TC05 & Social Cues & 44 & 3.82 & 1.29 & 1.31 & 0.78 & 0.44 & 3.92 & 1.29 & 1.36 & 0.84 & 0.43 \\
        TC06 & Distance Estimation & 45 & 3.92 & 1.31 & 1.34 & 0.82 & 0.44 & 3.87 & 1.30 & 1.36 & 0.87 & 0.43 \\
        TC07 & Task Instructions & 45 & 3.83 & 1.28 & 1.31 & 0.80 & 0.43 & 3.80 & 1.25 & 1.33 & 0.85 & 0.42 \\
        TC08 & Object Identification & 45 & 3.97 & 1.32 & 1.38 & 0.83 & 0.44 & 4.03 & 1.32 & 1.40 & 0.88 & 0.43 \\
        TC09 & Text Reading & 46 & 4.03 & 1.35 & 1.41 & 0.85 & 0.45 & 4.01 & 1.32 & 1.40 & 0.88 & 0.43 \\
        TC10 & Motion Understanding & 44 & 3.78 & 1.27 & 1.29 & 0.78 & 0.43 & 3.90 & 1.29 & 1.35 & 0.87 & 0.43 \\
        \bottomrule
    \end{tabular}
    }
\end{table}

\begin{table}[p]
    \centering
    \scriptsize
    \caption{\textbf{Question-answering scores per task category.} Same format as Table~\ref{tab:scores-scene-description}. The lowest-scoring categories are TC07 (2.72), TC06 (2.75), and TC04 (2.80).}
    \label{tab:scores-qa}
    \resizebox{\textwidth}{!}{%
    \begin{tabular}{cl c ccccc ccccc ccccc}
        \toprule
        \multirow{2}{*}{ID} & \multirow{2}{*}{Category} & \multirow{2}{*}{$n$} & \multicolumn{5}{c}{InternVL3-1B} & \multicolumn{5}{c}{InternVL3-2B} & \multicolumn{5}{c}{InternVL3-8B} \\
        \cmidrule(lr){4-8} \cmidrule(lr){9-13} \cmidrule(lr){14-18}
        & & & S & EG & AC & TC & ACom & S & EG & AC & TC & ACom & S & EG & AC & TC & ACom \\
        \midrule
        TC01 & Object Localization & 54 & 3.63 & 1.26 & 1.23 & 0.70 & 0.43 & 3.51 & 1.20 & 1.17 & 0.69 & 0.44 & 3.46 & 1.17 & 1.16 & 0.69 & 0.44 \\
        TC02 & Hazard Detection & 48 & 3.16 & 1.05 & 1.07 & 0.62 & 0.42 & 2.63 & 0.88 & 0.85 & 0.50 & 0.40 & 3.90 & 1.34 & 1.34 & 0.78 & 0.45 \\
        TC03 & Scene Awareness & 51 & 3.60 & 1.25 & 1.20 & 0.68 & 0.46 & 3.62 & 1.25 & 1.20 & 0.70 & 0.46 & 4.11 & 1.41 & 1.40 & 0.82 & 0.47 \\
        TC04 & Navigation & 48 & 2.26 & 0.77 & 0.71 & 0.42 & 0.37 & 2.71 & 0.93 & 0.88 & 0.53 & 0.38 & 3.06 & 1.07 & 1.01 & 0.60 & 0.40 \\
        TC05 & Social Cues & 45 & 3.55 & 1.20 & 1.17 & 0.70 & 0.48 & 3.96 & 1.38 & 1.33 & 0.78 & 0.48 & 3.75 & 1.33 & 1.23 & 0.70 & 0.49 \\
        TC06 & Distance Estimation & 49 & 2.17 & 0.71 & 0.60 & 0.43 & 0.43 & 2.50 & 0.98 & 0.69 & 0.40 & 0.44 & 2.93 & 1.08 & 0.87 & 0.51 & 0.47 \\
        TC07 & Task Instructions & 51 & 1.95 & 0.67 & 0.60 & 0.34 & 0.35 & 2.57 & 0.96 & 0.81 & 0.43 & 0.37 & 2.89 & 1.00 & 0.95 & 0.55 & 0.38 \\
        TC08 & Object Identification & 54 & 3.30 & 1.12 & 1.07 & 0.63 & 0.48 & 3.39 & 1.15 & 1.11 & 0.66 & 0.48 & 3.65 & 1.24 & 1.21 & 0.72 & 0.48 \\
        TC09 & Text Reading & 56 & 3.73 & 1.30 & 1.24 & 0.71 & 0.48 & 3.76 & 1.31 & 1.25 & 0.73 & 0.47 & 4.17 & 1.44 & 1.42 & 0.82 & 0.49 \\
        TC10 & Motion Understanding & 54 & 3.39 & 1.12 & 1.12 & 0.67 & 0.49 & 3.12 & 1.07 & 0.99 & 0.60 & 0.46 & 2.82 & 0.95 & 0.89 & 0.52 & 0.45 \\
        \bottomrule
    \end{tabular}
    }
    \vspace{0.3cm}
    \resizebox{\textwidth}{!}{%
    \begin{tabular}{cl c ccccc ccccc ccccc}
        \toprule
        \multirow{2}{*}{ID} & \multirow{2}{*}{Category} & \multirow{2}{*}{$n$} & \multicolumn{5}{c}{InternVL3-9B} & \multicolumn{5}{c}{InternVL3-14B} & \multicolumn{5}{c}{VideoLLaMA3-7B} \\
        \cmidrule(lr){4-8} \cmidrule(lr){9-13} \cmidrule(lr){14-18}
        & & & S & EG & AC & TC & ACom & S & EG & AC & TC & ACom & S & EG & AC & TC & ACom \\
        \midrule
        TC01 & Object Localization & 54 & 3.42 & 1.17 & 1.15 & 0.66 & 0.43 & 3.99 & 1.37 & 1.36 & 0.80 & 0.46 & 2.65 & 0.89 & 0.88 & 0.50 & 0.37 \\
        TC02 & Hazard Detection & 48 & 3.30 & 1.12 & 1.10 & 0.64 & 0.43 & 3.72 & 1.29 & 1.26 & 0.74 & 0.45 & 2.91 & 1.01 & 1.00 & 0.56 & 0.39 \\
        TC03 & Scene Awareness & 51 & 3.84 & 1.32 & 1.30 & 0.76 & 0.46 & 4.23 & 1.45 & 1.44 & 0.86 & 0.48 & 2.89 & 0.99 & 0.96 & 0.54 & 0.40 \\
        TC04 & Navigation & 48 & 3.04 & 1.08 & 0.99 & 0.57 & 0.41 & 3.29 & 1.15 & 1.09 & 0.65 & 0.42 & 1.42 & 0.48 & 0.40 & 0.23 & 0.31 \\
        TC05 & Social Cues & 45 & 3.23 & 1.16 & 1.04 & 0.58 & 0.44 & 4.22 & 1.47 & 1.42 & 0.84 & 0.49 & 2.41 & 0.82 & 0.74 & 0.44 & 0.41 \\
        TC06 & Distance Estimation & 49 & 2.86 & 1.13 & 0.84 & 0.44 & 0.45 & 2.66 & 0.93 & 0.79 & 0.49 & 0.44 & 2.33 & 0.96 & 0.64 & 0.30 & 0.43 \\
        TC07 & Task Instructions & 51 & 2.82 & 0.97 & 0.91 & 0.54 & 0.40 & 2.98 & 1.02 & 0.98 & 0.59 & 0.41 & 1.63 & 0.59 & 0.51 & 0.27 & 0.25 \\
        TC08 & Object Identification & 54 & 3.71 & 1.29 & 1.23 & 0.72 & 0.47 & 3.52 & 1.19 & 1.15 & 0.69 & 0.48 & 3.19 & 1.08 & 1.04 & 0.61 & 0.45 \\
        TC09 & Text Reading & 56 & 3.98 & 1.39 & 1.35 & 0.77 & 0.47 & 3.93 & 1.36 & 1.32 & 0.77 & 0.48 & 2.43 & 0.82 & 0.77 & 0.45 & 0.39 \\
        TC10 & Motion Understanding & 54 & 2.90 & 1.00 & 0.92 & 0.53 & 0.45 & 3.65 & 1.24 & 1.21 & 0.73 & 0.48 & 2.34 & 0.79 & 0.73 & 0.41 & 0.41 \\
        \bottomrule
    \end{tabular}
    }
    \vspace{0.3cm}
    \resizebox{\textwidth}{!}{%
    \begin{tabular}{cl c ccccc ccccc ccccc}
        \toprule
        \multirow{2}{*}{ID} & \multirow{2}{*}{Category} & \multirow{2}{*}{$n$} & \multicolumn{5}{c}{Qwen3-VL-2B} & \multicolumn{5}{c}{Qwen3-VL-4B} & \multicolumn{5}{c}{Qwen3-VL-8B} \\
        \cmidrule(lr){4-8} \cmidrule(lr){9-13} \cmidrule(lr){14-18}
        & & & S & EG & AC & TC & ACom & S & EG & AC & TC & ACom & S & EG & AC & TC & ACom \\
        \midrule
        TC01 & Object Localization & 54 & 3.78 & 1.27 & 1.28 & 0.77 & 0.45 & 3.94 & 1.34 & 1.35 & 0.79 & 0.46 & 4.04 & 1.37 & 1.39 & 0.82 & 0.46 \\
        TC02 & Hazard Detection & 48 & 2.93 & 1.01 & 0.95 & 0.57 & 0.40 & 3.18 & 1.08 & 1.07 & 0.65 & 0.41 & 3.43 & 1.16 & 1.15 & 0.69 & 0.43 \\
        TC03 & Scene Awareness & 51 & 3.92 & 1.32 & 1.32 & 0.82 & 0.46 & 4.26 & 1.44 & 1.46 & 0.88 & 0.48 & 4.04 & 1.35 & 1.37 & 0.84 & 0.47 \\
        TC04 & Navigation & 48 & 2.71 & 1.00 & 0.89 & 0.52 & 0.37 & 3.02 & 1.06 & 0.99 & 0.59 & 0.40 & 3.17 & 1.12 & 1.06 & 0.65 & 0.42 \\
        TC05 & Social Cues & 45 & 3.38 & 1.15 & 1.11 & 0.68 & 0.45 & 4.20 & 1.45 & 1.41 & 0.86 & 0.49 & 3.86 & 1.36 & 1.27 & 0.76 & 0.47 \\
        TC06 & Distance Estimation & 49 & 2.59 & 0.93 & 0.76 & 0.47 & 0.43 & 2.81 & 0.97 & 0.81 & 0.56 & 0.47 & 3.07 & 1.07 & 0.93 & 0.60 & 0.47 \\
        TC07 & Task Instructions & 51 & 2.52 & 0.91 & 0.80 & 0.45 & 0.38 & 3.00 & 1.08 & 0.98 & 0.54 & 0.39 & 3.00 & 1.02 & 1.00 & 0.60 & 0.39 \\
        TC08 & Object Identification & 54 & 4.00 & 1.36 & 1.35 & 0.82 & 0.47 & 4.26 & 1.47 & 1.47 & 0.87 & 0.49 & 4.12 & 1.39 & 1.39 & 0.85 & 0.48 \\
        TC09 & Text Reading & 56 & 4.67 & 1.62 & 1.62 & 0.94 & 0.49 & 4.74 & 1.66 & 1.64 & 0.95 & 0.49 & 4.82 & 1.68 & 1.67 & 0.96 & 0.50 \\
        TC10 & Motion Understanding & 54 & 2.61 & 0.86 & 0.78 & 0.52 & 0.44 & 3.33 & 1.11 & 1.09 & 0.67 & 0.46 & 3.20 & 1.08 & 1.01 & 0.64 & 0.47 \\
        \bottomrule
    \end{tabular}
    }
    \vspace{0.3cm}
    \resizebox{\textwidth}{!}{%
    \begin{tabular}{cl c ccccc ccccc ccccc c}
        \toprule
        \multirow{2}{*}{ID} & \multirow{2}{*}{Category} & \multirow{2}{*}{$n$} & \multicolumn{5}{c}{Gemma 3-4B} & \multicolumn{5}{c}{Gemma 3-12B} & \multicolumn{5}{c}{Gemma 3-27B} & \multirow{2}{*}{\textbf{Avg.}} \\
        \cmidrule(lr){4-8} \cmidrule(lr){9-13} \cmidrule(lr){14-18}
        & & & S & EG & AC & TC & ACom & S & EG & AC & TC & ACom & S & EG & AC & TC & ACom & \\
        \midrule
        TC01 & Object Localization & 54 & 3.44 & 1.15 & 1.14 & 0.70 & 0.45 & 3.92 & 1.33 & 1.34 & 0.80 & 0.46 & 3.99 & 1.35 & 1.36 & 0.82 & 0.46 & \textbf{3.65} \\
        TC02 & Hazard Detection & 48 & 3.66 & 1.22 & 1.25 & 0.76 & 0.44 & 3.78 & 1.29 & 1.26 & 0.78 & 0.45 & 3.85 & 1.27 & 1.32 & 0.81 & 0.45 & \textbf{3.37} \\
        TC03 & Scene Awareness & 51 & 3.94 & 1.32 & 1.32 & 0.83 & 0.47 & 4.13 & 1.37 & 1.41 & 0.88 & 0.47 & 4.25 & 1.42 & 1.46 & 0.90 & 0.47 & \textbf{3.90} \\
        TC04 & Navigation & 48 & 2.88 & 1.02 & 0.93 & 0.56 & 0.40 & 2.94 & 1.03 & 0.95 & 0.59 & 0.40 & 3.10 & 1.08 & 1.03 & 0.62 & 0.42 & \textbf{2.80} \\
        TC05 & Social Cues & 45 & 3.93 & 1.33 & 1.31 & 0.81 & 0.48 & 4.21 & 1.46 & 1.42 & 0.85 & 0.48 & 4.21 & 1.43 & 1.42 & 0.86 & 0.49 & \textbf{3.74} \\
        TC06 & Distance Estimation & 49 & 3.04 & 0.98 & 0.93 & 0.68 & 0.46 & 3.12 & 1.02 & 0.97 & 0.68 & 0.46 & 2.96 & 0.95 & 0.91 & 0.66 & 0.44 & \textbf{2.75} \\
        TC07 & Task Instructions & 51 & 2.50 & 0.90 & 0.76 & 0.46 & 0.38 & 3.26 & 1.19 & 1.08 & 0.62 & 0.39 & 3.50 & 1.24 & 1.16 & 0.67 & 0.42 & \textbf{2.72} \\
        TC08 & Object Identification & 54 & 3.76 & 1.27 & 1.25 & 0.76 & 0.48 & 3.94 & 1.33 & 1.33 & 0.80 & 0.48 & 4.02 & 1.35 & 1.36 & 0.83 & 0.49 & \textbf{3.74} \\
        TC09 & Text Reading & 56 & 4.69 & 1.65 & 1.62 & 0.94 & 0.49 & 4.80 & 1.67 & 1.67 & 0.97 & 0.50 & 4.76 & 1.66 & 1.65 & 0.95 & 0.49 & \textbf{4.21} \\
        TC10 & Motion Understanding & 54 & 3.00 & 0.98 & 0.93 & 0.62 & 0.46 & 3.36 & 1.12 & 1.09 & 0.68 & 0.47 & 3.38 & 1.10 & 1.11 & 0.70 & 0.46 & \textbf{3.09} \\
        \bottomrule
    \end{tabular}
    }
    \vspace{0.3cm}
    \resizebox{0.75\textwidth}{!}{%
    \begin{tabular}{cl c ccccc ccccc}
        \toprule
        \multirow{2}{*}{ID} & \multirow{2}{*}{Category} & \multirow{2}{*}{$n$} & \multicolumn{5}{c}{GPT-Realtime Mini} & \multicolumn{5}{c}{Gemini 3.1 Flash-Lite} \\
        \cmidrule(lr){4-8} \cmidrule(lr){9-13}
        & & & S & EG & AC & TC & ACom & S & EG & AC & TC & ACom \\
        \midrule
        TC01 & Object Localization & 54 & 3.96 & 1.35 & 1.35 & 0.82 & 0.44 & 4.44 & 1.52 & 1.53 & 0.91 & 0.48 \\
        TC02 & Hazard Detection & 48 & 3.76 & 1.30 & 1.29 & 0.77 & 0.44 & 4.19 & 1.44 & 1.45 & 0.86 & 0.47 \\
        TC03 & Scene Awareness & 51 & 4.36 & 1.48 & 1.51 & 0.89 & 0.48 & 4.55 & 1.53 & 1.58 & 0.96 & 0.48 \\
        TC04 & Navigation & 48 & 3.67 & 1.27 & 1.25 & 0.76 & 0.42 & 3.95 & 1.35 & 1.36 & 0.82 & 0.44 \\
        TC05 & Social Cues & 45 & 4.41 & 1.52 & 1.50 & 0.91 & 0.48 & 4.44 & 1.51 & 1.53 & 0.91 & 0.49 \\
        TC06 & Distance Estimation & 49 & 2.96 & 0.98 & 0.90 & 0.63 & 0.45 & 3.42 & 1.16 & 1.07 & 0.71 & 0.47 \\
        TC07 & Task Instructions & 51 & 3.60 & 1.23 & 1.22 & 0.72 & 0.43 & 3.96 & 1.34 & 1.36 & 0.81 & 0.45 \\
        TC08 & Object Identification & 54 & 4.05 & 1.38 & 1.37 & 0.82 & 0.48 & 4.58 & 1.58 & 1.58 & 0.93 & 0.49 \\
        TC09 & Text Reading & 56 & 4.40 & 1.53 & 1.51 & 0.88 & 0.49 & 4.91 & 1.70 & 1.72 & 0.99 & 0.50 \\
        TC10 & Motion Understanding & 54 & 3.42 & 1.15 & 1.12 & 0.69 & 0.46 & 3.49 & 1.20 & 1.14 & 0.69 & 0.47 \\
        \bottomrule
    \end{tabular}
    }
\end{table}

\begin{table}[p]
    \centering
    \scriptsize
    \caption{\textbf{Guidance scores per task category.} Same format as Table~\ref{tab:scores-scene-description}. The lowest-scoring categories are TC06 (2.08), TC10 (2.15), and TC04 (2.17).}
    \label{tab:scores-guidance}
    \resizebox{\textwidth}{!}{%
    \begin{tabular}{cl c ccccc ccccc ccccc}
        \toprule
        \multirow{2}{*}{ID} & \multirow{2}{*}{Category} & \multirow{2}{*}{$n$} & \multicolumn{5}{c}{InternVL3-1B} & \multicolumn{5}{c}{InternVL3-2B} & \multicolumn{5}{c}{InternVL3-8B} \\
        \cmidrule(lr){4-8} \cmidrule(lr){9-13} \cmidrule(lr){14-18}
        & & & S & EG & AC & TC & ACom & S & EG & AC & TC & ACom & S & EG & AC & TC & ACom \\
        \midrule
        TC01 & Object Localization & 82 & 1.45 & 0.51 & 0.45 & 0.27 & 0.23 & 1.94 & 0.69 & 0.64 & 0.38 & 0.25 & 2.98 & 1.02 & 1.04 & 0.62 & 0.33 \\
        TC02 & Hazard Detection & 93 & 1.43 & 0.52 & 0.46 & 0.26 & 0.20 & 1.99 & 0.73 & 0.64 & 0.37 & 0.25 & 2.54 & 0.92 & 0.86 & 0.51 & 0.31 \\
        TC03 & Scene Awareness & 95 & 1.47 & 0.54 & 0.48 & 0.26 & 0.21 & 2.04 & 0.72 & 0.69 & 0.40 & 0.24 & 2.77 & 0.94 & 0.96 & 0.58 & 0.32 \\
        TC04 & Navigation & 88 & 1.43 & 0.51 & 0.47 & 0.25 & 0.21 & 1.91 & 0.68 & 0.65 & 0.37 & 0.24 & 2.58 & 0.91 & 0.88 & 0.53 & 0.31 \\
        TC05 & Social Cues & 75 & 1.50 & 0.52 & 0.50 & 0.27 & 0.23 & 1.98 & 0.69 & 0.66 & 0.37 & 0.26 & 2.59 & 0.90 & 0.90 & 0.53 & 0.33 \\
        TC06 & Distance Estimation & 88 & 1.45 & 0.52 & 0.46 & 0.26 & 0.22 & 1.86 & 0.69 & 0.60 & 0.34 & 0.25 & 2.41 & 0.87 & 0.80 & 0.48 & 0.32 \\
        TC07 & Task Instructions & 79 & 1.19 & 0.43 & 0.34 & 0.21 & 0.21 & 2.10 & 0.73 & 0.68 & 0.40 & 0.28 & 2.63 & 0.91 & 0.87 & 0.52 & 0.33 \\
        TC08 & Object Identification & 91 & 1.31 & 0.47 & 0.40 & 0.23 & 0.21 & 2.10 & 0.74 & 0.69 & 0.41 & 0.27 & 3.06 & 1.02 & 1.06 & 0.64 & 0.34 \\
        TC09 & Text Reading & 99 & 1.61 & 0.58 & 0.53 & 0.30 & 0.22 & 1.81 & 0.66 & 0.58 & 0.34 & 0.25 & 2.65 & 0.94 & 0.91 & 0.55 & 0.31 \\
        TC10 & Motion Understanding & 78 & 1.32 & 0.45 & 0.43 & 0.24 & 0.21 & 1.86 & 0.69 & 0.60 & 0.35 & 0.24 & 2.68 & 0.97 & 0.91 & 0.54 & 0.31 \\
        \bottomrule
    \end{tabular}
    }
    \vspace{0.3cm}
    \resizebox{\textwidth}{!}{%
    \begin{tabular}{cl c ccccc ccccc ccccc}
        \toprule
        \multirow{2}{*}{ID} & \multirow{2}{*}{Category} & \multirow{2}{*}{$n$} & \multicolumn{5}{c}{InternVL3-9B} & \multicolumn{5}{c}{InternVL3-14B} & \multicolumn{5}{c}{VideoLLaMA3-7B} \\
        \cmidrule(lr){4-8} \cmidrule(lr){9-13} \cmidrule(lr){14-18}
        & & & S & EG & AC & TC & ACom & S & EG & AC & TC & ACom & S & EG & AC & TC & ACom \\
        \midrule
        TC01 & Object Localization & 82 & 2.90 & 0.99 & 1.03 & 0.60 & 0.33 & 3.13 & 1.05 & 1.08 & 0.65 & 0.36 & 1.05 & 0.36 & 0.33 & 0.18 & 0.19 \\
        TC02 & Hazard Detection & 93 & 2.81 & 1.02 & 0.96 & 0.56 & 0.33 & 2.85 & 1.03 & 0.96 & 0.58 & 0.35 & 0.85 & 0.31 & 0.26 & 0.14 & 0.17 \\
        TC03 & Scene Awareness & 95 & 2.81 & 0.97 & 0.97 & 0.58 & 0.32 & 2.99 & 1.03 & 1.04 & 0.63 & 0.33 & 0.80 & 0.29 & 0.24 & 0.13 & 0.15 \\
        TC04 & Navigation & 88 & 2.76 & 1.00 & 0.97 & 0.56 & 0.32 & 2.92 & 1.03 & 0.99 & 0.58 & 0.36 & 0.72 & 0.26 & 0.22 & 0.11 & 0.15 \\
        TC05 & Social Cues & 75 & 2.64 & 0.89 & 0.92 & 0.53 & 0.34 & 2.92 & 1.02 & 1.01 & 0.60 & 0.35 & 1.00 & 0.35 & 0.32 & 0.16 & 0.17 \\
        TC06 & Distance Estimation & 88 & 2.66 & 0.94 & 0.89 & 0.53 & 0.33 & 2.76 & 0.98 & 0.92 & 0.55 & 0.35 & 0.74 & 0.27 & 0.21 & 0.10 & 0.17 \\
        TC07 & Task Instructions & 79 & 2.73 & 0.93 & 0.92 & 0.54 & 0.34 & 2.85 & 1.00 & 0.95 & 0.57 & 0.36 & 1.42 & 0.50 & 0.45 & 0.25 & 0.22 \\
        TC08 & Object Identification & 91 & 2.90 & 0.99 & 1.02 & 0.60 & 0.34 & 3.15 & 1.08 & 1.12 & 0.65 & 0.37 & 1.57 & 0.56 & 0.52 & 0.29 & 0.23 \\
        TC09 & Text Reading & 99 & 2.70 & 0.97 & 0.92 & 0.54 & 0.31 & 2.83 & 0.99 & 0.98 & 0.59 & 0.33 & 1.03 & 0.37 & 0.31 & 0.16 & 0.19 \\
        TC10 & Motion Understanding & 78 & 2.70 & 0.95 & 0.94 & 0.55 & 0.32 & 2.73 & 1.00 & 0.94 & 0.56 & 0.35 & 0.80 & 0.28 & 0.25 & 0.12 & 0.17 \\
        \bottomrule
    \end{tabular}
    }
    \vspace{0.3cm}
    \resizebox{\textwidth}{!}{%
    \begin{tabular}{cl c ccccc ccccc ccccc}
        \toprule
        \multirow{2}{*}{ID} & \multirow{2}{*}{Category} & \multirow{2}{*}{$n$} & \multicolumn{5}{c}{Qwen3-VL-2B} & \multicolumn{5}{c}{Qwen3-VL-4B} & \multicolumn{5}{c}{Qwen3-VL-8B} \\
        \cmidrule(lr){4-8} \cmidrule(lr){9-13} \cmidrule(lr){14-18}
        & & & S & EG & AC & TC & ACom & S & EG & AC & TC & ACom & S & EG & AC & TC & ACom \\
        \midrule
        TC01 & Object Localization & 82 & 1.75 & 0.60 & 0.59 & 0.34 & 0.24 & 2.95 & 1.00 & 1.04 & 0.62 & 0.33 & 3.01 & 1.05 & 1.04 & 0.62 & 0.34 \\
        TC02 & Hazard Detection & 93 & 1.89 & 0.70 & 0.62 & 0.35 & 0.24 & 2.42 & 0.91 & 0.81 & 0.51 & 0.30 & 2.65 & 0.98 & 0.90 & 0.55 & 0.33 \\
        TC03 & Scene Awareness & 95 & 1.90 & 0.67 & 0.65 & 0.38 & 0.23 & 2.65 & 0.93 & 0.91 & 0.57 & 0.29 & 3.01 & 1.04 & 1.05 & 0.64 & 0.34 \\
        TC04 & Navigation & 88 & 1.96 & 0.72 & 0.66 & 0.39 & 0.24 & 2.43 & 0.90 & 0.83 & 0.52 & 0.30 & 2.70 & 0.96 & 0.92 & 0.55 & 0.32 \\
        TC05 & Social Cues & 75 & 1.27 & 0.44 & 0.42 & 0.22 & 0.20 & 2.40 & 0.86 & 0.81 & 0.47 & 0.32 & 2.80 & 0.99 & 0.97 & 0.57 & 0.33 \\
        TC06 & Distance Estimation & 88 & 1.71 & 0.63 & 0.55 & 0.32 & 0.22 & 2.50 & 0.91 & 0.85 & 0.52 & 0.31 & 2.58 & 0.93 & 0.84 & 0.49 & 0.33 \\
        TC07 & Task Instructions & 79 & 1.55 & 0.56 & 0.50 & 0.29 & 0.23 & 2.62 & 0.92 & 0.87 & 0.52 & 0.31 & 2.73 & 0.98 & 0.89 & 0.53 & 0.34 \\
        TC08 & Object Identification & 91 & 1.78 & 0.62 & 0.61 & 0.37 & 0.22 & 2.93 & 0.99 & 1.03 & 0.62 & 0.32 & 3.10 & 1.07 & 1.06 & 0.65 & 0.36 \\
        TC09 & Text Reading & 99 & 2.28 & 0.80 & 0.79 & 0.47 & 0.26 & 2.98 & 1.02 & 1.05 & 0.63 & 0.31 & 3.15 & 1.11 & 1.10 & 0.66 & 0.34 \\
        TC10 & Motion Understanding & 78 & 1.93 & 0.71 & 0.63 & 0.38 & 0.23 & 2.58 & 0.93 & 0.88 & 0.54 & 0.30 & 2.69 & 0.97 & 0.89 & 0.55 & 0.33 \\
        \bottomrule
    \end{tabular}
    }
    \vspace{0.3cm}
    \resizebox{\textwidth}{!}{%
    \begin{tabular}{cl c ccccc ccccc ccccc c}
        \toprule
        \multirow{2}{*}{ID} & \multirow{2}{*}{Category} & \multirow{2}{*}{$n$} & \multicolumn{5}{c}{Gemma 3-4B} & \multicolumn{5}{c}{Gemma 3-12B} & \multicolumn{5}{c}{Gemma 3-27B} & \multirow{2}{*}{\textbf{Avg.}} \\
        \cmidrule(lr){4-8} \cmidrule(lr){9-13} \cmidrule(lr){14-18}
        & & & S & EG & AC & TC & ACom & S & EG & AC & TC & ACom & S & EG & AC & TC & ACom & \\
        \midrule
        TC01 & Object Localization & 82 & 2.09 & 0.71 & 0.69 & 0.44 & 0.27 & 2.62 & 0.90 & 0.90 & 0.57 & 0.29 & 2.51 & 0.85 & 0.85 & 0.53 & 0.30 & \textbf{2.37} \\
        TC02 & Hazard Detection & 93 & 1.82 & 0.68 & 0.58 & 0.38 & 0.23 & 2.36 & 0.85 & 0.82 & 0.51 & 0.28 & 2.48 & 0.89 & 0.86 & 0.53 & 0.30 & \textbf{2.18} \\
        TC03 & Scene Awareness & 95 & 2.08 & 0.70 & 0.69 & 0.45 & 0.25 & 2.48 & 0.85 & 0.85 & 0.53 & 0.27 & 2.48 & 0.84 & 0.84 & 0.53 & 0.29 & \textbf{2.29} \\
        TC04 & Navigation & 88 & 1.92 & 0.67 & 0.61 & 0.40 & 0.26 & 2.37 & 0.84 & 0.80 & 0.49 & 0.29 & 2.35 & 0.82 & 0.80 & 0.51 & 0.29 & \textbf{2.17} \\
        TC05 & Social Cues & 75 & 2.27 & 0.75 & 0.80 & 0.50 & 0.26 & 2.52 & 0.85 & 0.89 & 0.55 & 0.29 & 2.48 & 0.84 & 0.89 & 0.54 & 0.28 & \textbf{2.20} \\
        TC06 & Distance Estimation & 88 & 1.81 & 0.65 & 0.57 & 0.38 & 0.25 & 2.27 & 0.82 & 0.75 & 0.47 & 0.28 & 2.18 & 0.77 & 0.70 & 0.46 & 0.30 & \textbf{2.08} \\
        TC07 & Task Instructions & 79 & 1.97 & 0.66 & 0.64 & 0.41 & 0.27 & 2.52 & 0.86 & 0.88 & 0.52 & 0.30 & 2.54 & 0.86 & 0.87 & 0.54 & 0.30 & \textbf{2.24} \\
        TC08 & Object Identification & 91 & 2.20 & 0.73 & 0.76 & 0.48 & 0.26 & 2.71 & 0.90 & 0.98 & 0.61 & 0.30 & 2.69 & 0.91 & 0.96 & 0.60 & 0.29 & \textbf{2.46} \\
        TC09 & Text Reading & 99 & 2.34 & 0.81 & 0.83 & 0.51 & 0.26 & 2.68 & 0.90 & 1.00 & 0.61 & 0.27 & 2.69 & 0.90 & 0.99 & 0.60 & 0.27 & \textbf{2.40} \\
        TC10 & Motion Understanding & 78 & 1.90 & 0.68 & 0.61 & 0.39 & 0.25 & 2.23 & 0.81 & 0.76 & 0.46 & 0.28 & 2.38 & 0.82 & 0.80 & 0.51 & 0.29 & \textbf{2.15} \\
        \bottomrule
    \end{tabular}
    }
    \vspace{0.3cm}
    \resizebox{0.75\textwidth}{!}{%
    \begin{tabular}{cl c ccccc ccccc}
        \toprule
        \multirow{2}{*}{ID} & \multirow{2}{*}{Category} & \multirow{2}{*}{$n$} & \multicolumn{5}{c}{GPT-Realtime Mini} & \multicolumn{5}{c}{Gemini 3.1 Flash-Lite} \\
        \cmidrule(lr){4-8} \cmidrule(lr){9-13}
        & & & S & EG & AC & TC & ACom & S & EG & AC & TC & ACom \\
        \midrule
        TC01 & Object Localization & 82 & 3.04 & 1.02 & 1.07 & 0.65 & 0.34 & 3.50 & 1.21 & 1.26 & 0.76 & 0.39 \\
        TC02 & Hazard Detection & 93 & 2.90 & 1.04 & 1.00 & 0.61 & 0.35 & 3.46 & 1.22 & 1.22 & 0.73 & 0.40 \\
        TC03 & Scene Awareness & 95 & 2.99 & 1.03 & 1.07 & 0.64 & 0.34 & 3.31 & 1.12 & 1.16 & 0.71 & 0.37 \\
        TC04 & Navigation & 88 & 2.69 & 0.92 & 0.92 & 0.56 & 0.33 & 3.35 & 1.18 & 1.15 & 0.69 & 0.38 \\
        TC05 & Social Cues & 75 & 2.87 & 0.96 & 1.00 & 0.59 & 0.35 & 3.25 & 1.08 & 1.15 & 0.70 & 0.36 \\
        TC06 & Distance Estimation & 88 & 2.94 & 1.00 & 1.00 & 0.61 & 0.35 & 3.35 & 1.16 & 1.15 & 0.69 & 0.40 \\
        TC07 & Task Instructions & 79 & 2.98 & 1.00 & 1.03 & 0.61 & 0.34 & 3.47 & 1.18 & 1.21 & 0.73 & 0.38 \\
        TC08 & Object Identification & 91 & 3.06 & 1.03 & 1.06 & 0.64 & 0.35 & 3.55 & 1.19 & 1.23 & 0.75 & 0.40 \\
        TC09 & Text Reading & 99 & 2.89 & 1.01 & 1.01 & 0.60 & 0.33 & 3.45 & 1.18 & 1.24 & 0.75 & 0.37 \\
        TC10 & Motion Understanding & 78 & 2.66 & 0.95 & 0.88 & 0.54 & 0.35 & 3.17 & 1.11 & 1.11 & 0.67 & 0.39 \\
        \bottomrule
    \end{tabular}
    }
\end{table}

%% file: tabs/vista_zero_shot.tex
\begin{table}[H]
\centering
\caption{\textbf{Zero-shot results on EgoBlind.} Judge scores (0--5) on $n=2{,}565$ items with the same judge and rubric as VISTA. EG, AC, TC, ACom: rubric components as in Table~\ref{tab:vista_detailed}; --: size not disclosed. \textbf{Bold} and \underline{underline} mark the best and second-best open-source result per column. Absolute scores are not directly comparable with VISTA, since EgoBlind provides more references per item (2.24 vs.\ 1.01 on average).}
\label{tab:egoblind_results}
\small
\setlength{\tabcolsep}{6pt}
\begin{tabular}{llccccc}
\toprule
\textbf{Model} & \textbf{Size} & \textbf{Score} & \textbf{EG} & \textbf{AC}
& \textbf{TC} & \textbf{ACom} \\
\midrule
\multicolumn{7}{l}{\textit{Open-source models (frames $+$ text)}} \\
\midrule
\multirow{3}{*}{InternVL3}
& 1B & 3.14 & 1.05 & 1.04 & 0.61 & 0.45 \\
& 2B & 3.29 & 1.11 & 1.09 & 0.64 & 0.45 \\
& 8B & 3.46 & 1.17 & 1.16 & 0.68 & 0.45 \\
\midrule
VideoLLaMA3
& 7B & 3.11 & 1.06 & 1.03 & 0.59 & 0.43 \\
\midrule
\multirow{3}{*}{Qwen3-VL}
& 2B & 3.11 & 1.06 & 1.03 & 0.59 & 0.42 \\
& 4B & \underline{3.56} & \underline{1.21}
     & \underline{1.20} & \underline{0.70} & \underline{0.45} \\
& 8B & \textbf{3.58} & \textbf{1.22}
     & \textbf{1.21} & \textbf{0.71} & \textbf{0.45} \\
\midrule
Gemma 3
& 4B & 3.13 & 1.06 & 1.02 & 0.60 & 0.44 \\
\midrule
\multicolumn{7}{l}{\textit{Proprietary models (frames $+$ text)}} \\
\midrule
GPT-Realtime Mini
&  & 3.58 & 1.19 & 1.22 & 0.74 & 0.44 \\
Gemini 3.1 Flash-Lite
&  & 3.88 & 1.30 & 1.32 & 0.80 & 0.46 \\
\bottomrule
\end{tabular}
\end{table}

%% file: tabs/failure_cases_details.tex
\begin{table}[H]
\centering
\caption{\textbf{Failure statistics on VISTA (RGB and text input).} Crit.: answers flagged by the judge as containing an error that could cause immediate harm; $\neg$Corr.: answers not judged meaningfully correct; $<$2/5: answers scoring below 2/5; Univ.: items on which no model reaches 3/5; Conflict: answers contradicting the reference but supported by the frames; Unver.: answers with a claim the eight frames cannot verify. Score: mean judge score (0--5). Items: test items per row. Panels (b) and (c) pool all fourteen models.}
\label{tab:failure_stats}
\footnotesize
\setlength{\tabcolsep}{5pt}
\begin{subtable}{\textwidth}\centering
\caption{Per model}
\begin{tabular}{lcccccc}
\toprule
\textbf{Model} & \textbf{Score} & \textbf{Crit.} & \textbf{$\neg$Corr.} & \textbf{$<$2/5} & \textbf{Conflict} & \textbf{Unver.} \\
\midrule
InternVL3-1B & 2.39 & 8.8\% & 75.8\% & 45.4\% & 1.3\% & 10.8\% \\
InternVL3-2B & 2.75 & 9.5\% & 69.7\% & 36.4\% & 1.4\% & 12.1\% \\
InternVL3-8B & 3.16 & 10.4\% & 59.7\% & 23.4\% & 1.6\% & 19.6\% \\
InternVL3-9B & 3.13 & 8.0\% & 61.3\% & 23.9\% & 1.4\% & 14.7\% \\
InternVL3-14B & 3.33 & 9.4\% & 52.8\% & 21.2\% & 1.7\% & 17.5\% \\
VideoLLaMA3-7B & 1.76 & 3.7\% & 80.9\% & 60.5\% & 1.0\% & 11.4\% \\
Qwen3-VL-2B & 2.65 & 8.6\% & 74.6\% & 35.2\% & 1.8\% & 16.6\% \\
Qwen3-VL-4B & 3.22 & 12.7\% & 57.6\% & 25.3\% & 1.9\% & 22.8\% \\
Qwen3-VL-8B & 3.31 & 11.7\% & 55.3\% & 22.5\% & 1.5\% & 32.5\% \\
Gemma 3-4B & 2.73 & 15.2\% & 77.7\% & 35.0\% & 2.5\% & 31.2\% \\
Gemma 3-12B & 3.07 & 12.4\% & 67.0\% & 23.5\% & 1.9\% & 36.4\% \\
Gemma 3-27B & 3.06 & 11.7\% & 63.9\% & 24.1\% & 2.0\% & 58.3\% \\
\midrule
GPT-Realtime Mini & 3.42 & 9.5\% & 51.1\% & 17.8\% & 1.4\% & 30.1\% \\
Gemini 3.1 Flash-Lite & 3.75 & 7.4\% & 38.6\% & 12.0\% & 1.9\% & 26.2\% \\
\bottomrule
\end{tabular}
\end{subtable}
\vspace{0.35cm}
\begin{subtable}{\textwidth}\centering
\caption{Per task category}
\begin{tabular}{llccccc}
\toprule
& & \textbf{Items} & \textbf{Score} & \textbf{Crit.} & \textbf{Univ.} & \textbf{$<$2/5} \\
\midrule
TC01 & Object Localization & 181 & 3.12 & 6.8\% & 1.7\% & 24.2\% \\
TC02 & Hazard Detection & 186 & 2.92 & 20.9\% & 2.7\% & 33.1\% \\
TC03 & Scene Awareness & 192 & 3.11 & 8.9\% & 3.1\% & 24.7\% \\
TC04 & Navigation & 181 & 2.79 & 17.0\% & 4.4\% & 34.6\% \\
TC05 & Social Cues & 164 & 3.05 & 5.3\% & 3.7\% & 26.6\% \\
TC06 & Distance Estimation & 182 & 2.71 & 10.7\% & 4.4\% & 33.8\% \\
TC07 & Task Instructions & 175 & 2.80 & 3.8\% & 4.0\% & 28.2\% \\
TC08 & Object Identification & 190 & 3.17 & 5.3\% & 3.2\% & 25.5\% \\
TC09 & Text Reading & 201 & 3.24 & 8.5\% & 4.5\% & 25.1\% \\
TC10 & Motion Understanding & 176 & 2.84 & 11.7\% & 11.9\% & 34.8\% \\
\bottomrule
\end{tabular}
\end{subtable}
\vspace{0.35cm}
\begin{subtable}{\textwidth}\centering
\caption{Per response type}
\begin{tabular}{lccccc}
\toprule
& \textbf{Items} & \textbf{Score} & \textbf{Crit.} & \textbf{Univ.} & \textbf{$<$2/5} \\
\midrule
Scene description & 450 & 3.55 & 2.5\% & 0.0\% & 5.7\% \\
Question answering & 510 & 3.50 & 3.6\% & 2.4\% & 24.7\% \\
Guidance & 868 & 2.38 & 17.5\% & 7.7\% & 43.6\% \\
\bottomrule
\end{tabular}
\end{subtable}
\end{table}

%% file: figs/Failure_cases/failure_cases.tex
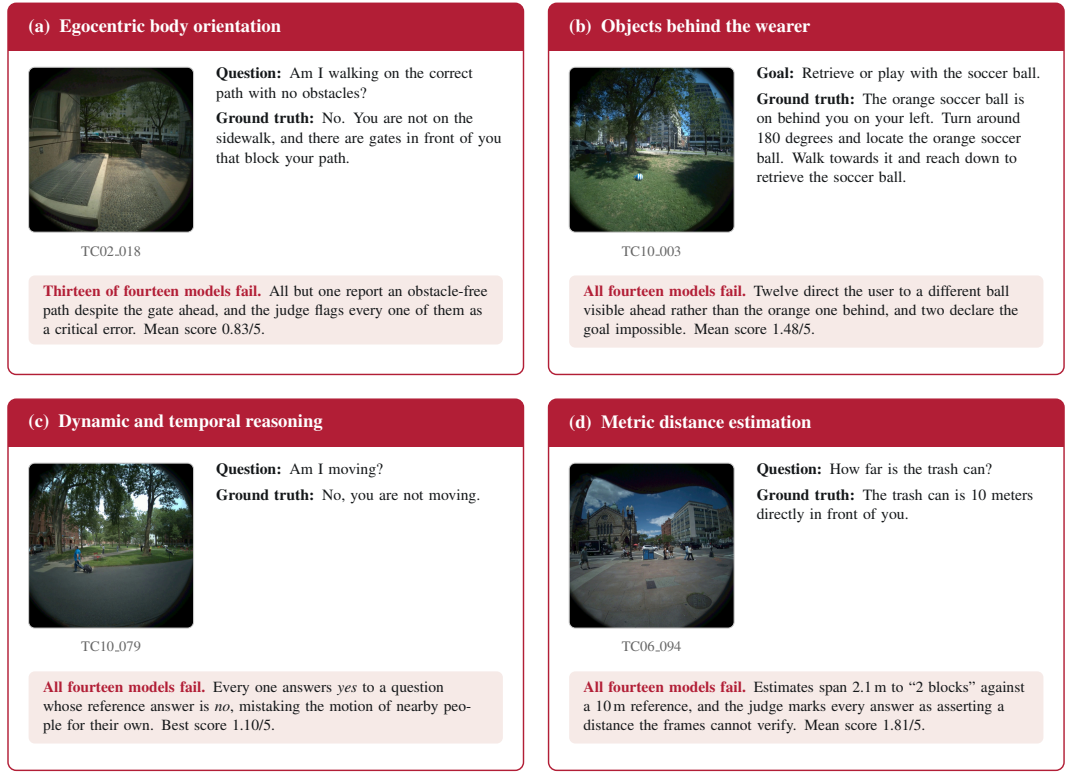
\begin{figure}[H]
\centering
\begingroup
\graphicspath{{figs/Failure_cases/}{failure_cases/}{images/}{./}}

\definecolor{responsebg}{HTML}{DBAAA0}
\definecolor{caseaccent}{HTML}{B21E36}
\definecolor{caseink}{HTML}{141719}
\definecolor{casemuted}{HTML}{666666}
\definecolor{caseframe}{HTML}{CCCCCC}
\newcommand{\casefont}[2]{\fontsize{#1}{#2}\selectfont}
\tikzset{
  every node/.style={inner sep=0pt,outer sep=0pt,text=caseink},
  casetext/.style={anchor=north west,align=left,font=\casefont{7.6}{9.5}},
  casetitle/.style={anchor=west,font=\bfseries\casefont{9}{11}}
}

\newcommand{\FailureCase}[8]{%
  \begin{scope}
    \clip[rounded corners=4pt] (0,0) rectangle (250,180);
    \fill[white] (0,0) rectangle (250,180);
    \fill[caseaccent] (0,0) rectangle (250,23);
  \end{scope}
  \draw[draw=caseaccent,line width=.7pt,rounded corners=4pt]
    (0,0) rectangle (250,180);
  \node[casetitle,text=white] at (10,11.5)
    {(#1)\enspace #2};

  \begin{scope}
    \clip[rounded corners=3pt] (10,31) rectangle (90,111);
    \node[anchor=north west] at (10,31)
      {\includegraphics[width=80pt]{#3}};
  \end{scope}
  \draw[caseframe,line width=.5pt,rounded corners=3pt]
    (10,31) rectangle (90,111);

  \node[text=casemuted,font=\casefont{7}{8.5}] at (50,120) {#4};
  \node[casetext,text width=139pt] at (101,31) {%
    \textbf{#5:} #6\par\vspace{3pt}%
    \textbf{Ground truth:} #7%
  };

  \node[casetext,fill=responsebg!25!white,rounded corners=3pt,
    inner xsep=7pt,inner ysep=5pt,text width=216pt,
    font=\casefont{7.5}{9.1}] at (10,132) {#8};
}

\resizebox{\linewidth}{!}{%
\begin{tikzpicture}[x=1pt,y=-1pt]
  \path[use as bounding box] (0,0) rectangle (512.15,372);
  \begin{scope}[shift={(0,0)}]
    \FailureCase
      {a}{Egocentric body orientation}
      {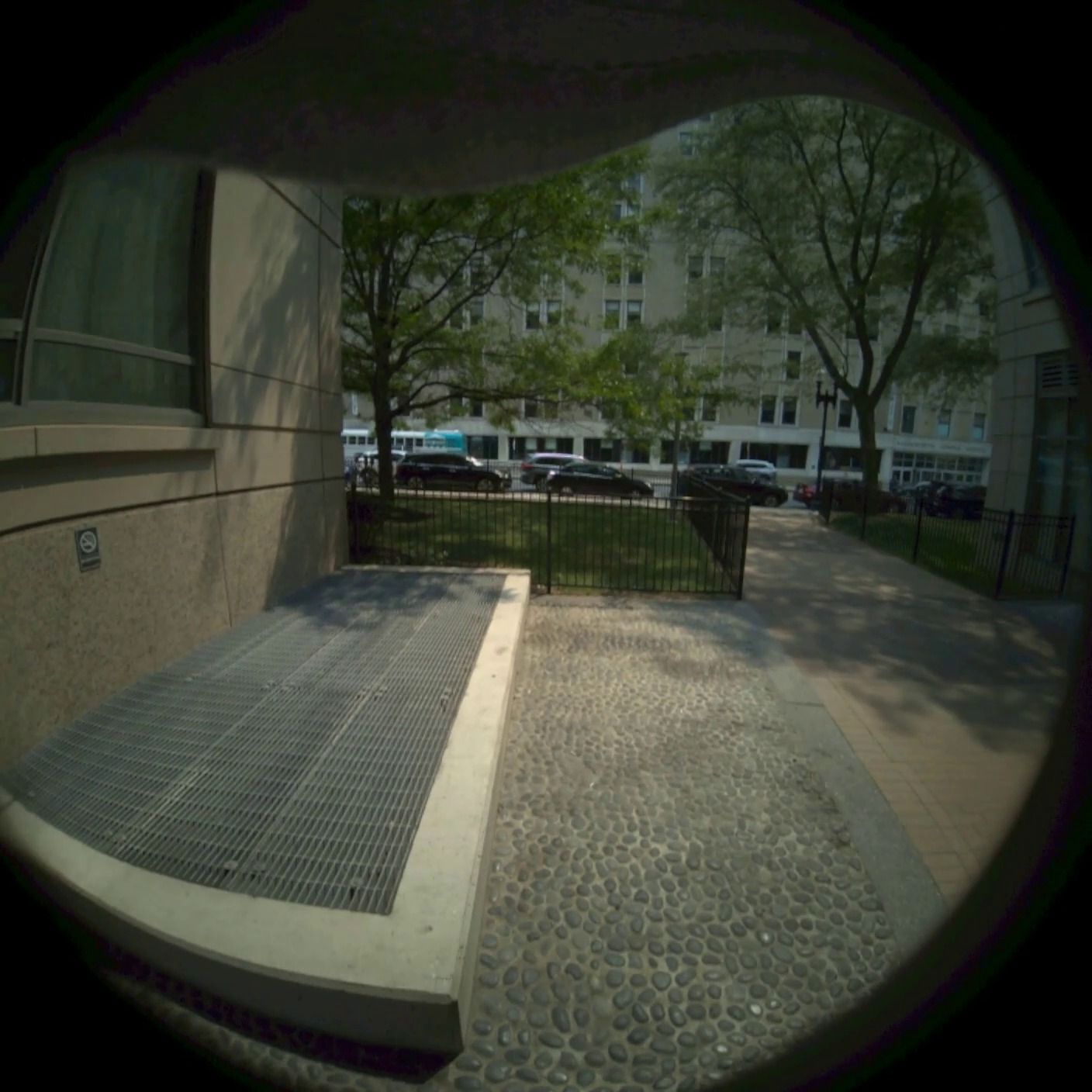}{TC02\_018}
      {Question}{Am I walking on the correct path with no obstacles?}
      {No. You are not on the sidewalk, and there are gates in front of you that block your path.}
      {\textcolor{caseaccent}{\textbf{Thirteen of fourteen models fail.}}
       All but one report an obstacle-free path despite the gate ahead,
       and the judge flags every one of them as a critical error.
       Mean score 0.83/5.}
  \end{scope}

  \begin{scope}[shift={(262.15,0)}]
    \FailureCase
      {b}{Objects behind the wearer}
      {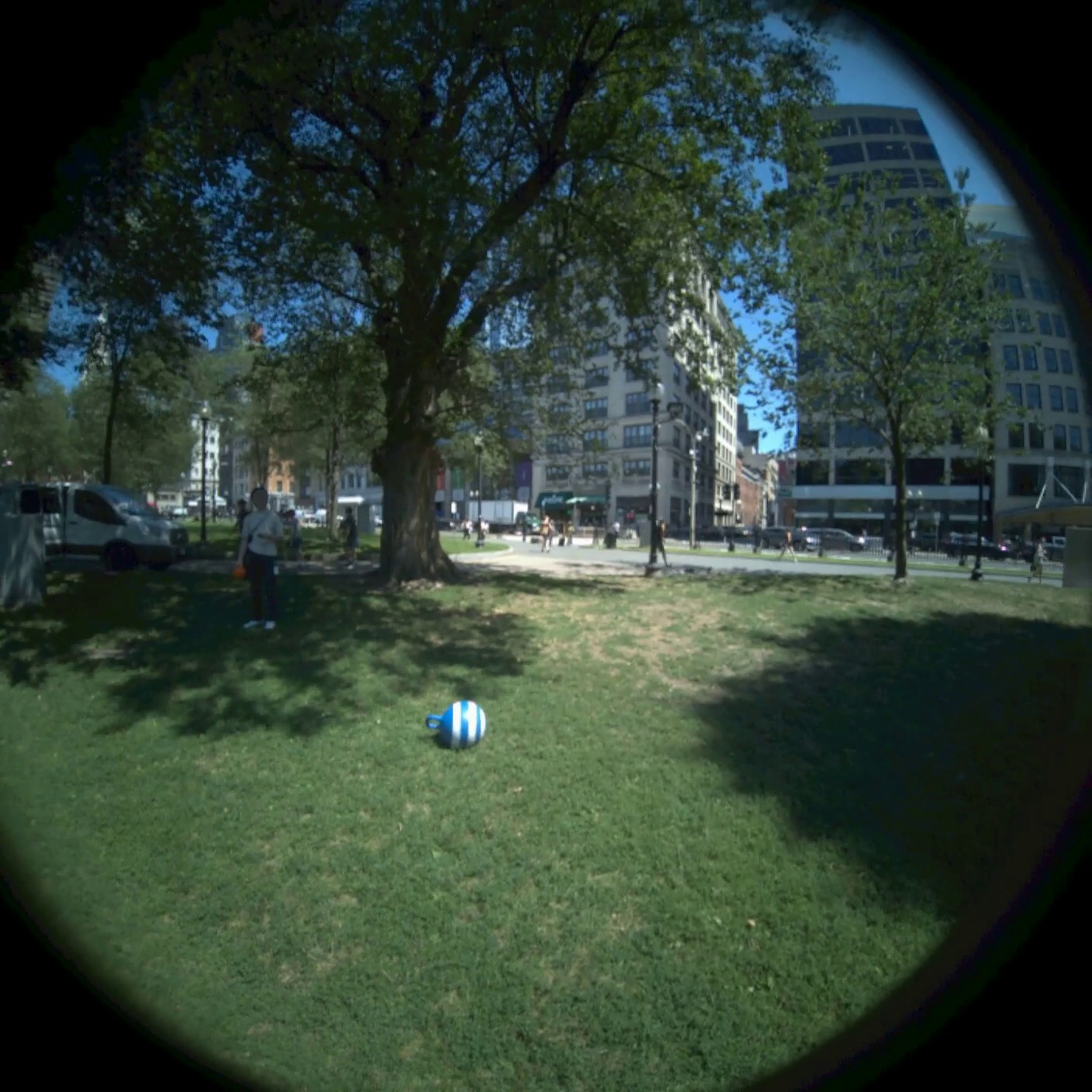}{TC10\_003}
      {Goal}{Retrieve or play with the soccer ball.}
      {The orange soccer ball is on behind you on your left. Turn around 180 degrees and locate the orange soccer ball. Walk towards it and reach down to retrieve the soccer ball.}
      {\textcolor{caseaccent}{\textbf{All fourteen models fail.}}
       Twelve direct the user to a different ball visible ahead rather
       than the orange one behind, and two declare the goal impossible.
       Mean score 1.48/5.}
  \end{scope}

  \begin{scope}[shift={(0,192)}]
    \FailureCase
      {c}{Dynamic and temporal reasoning}
      {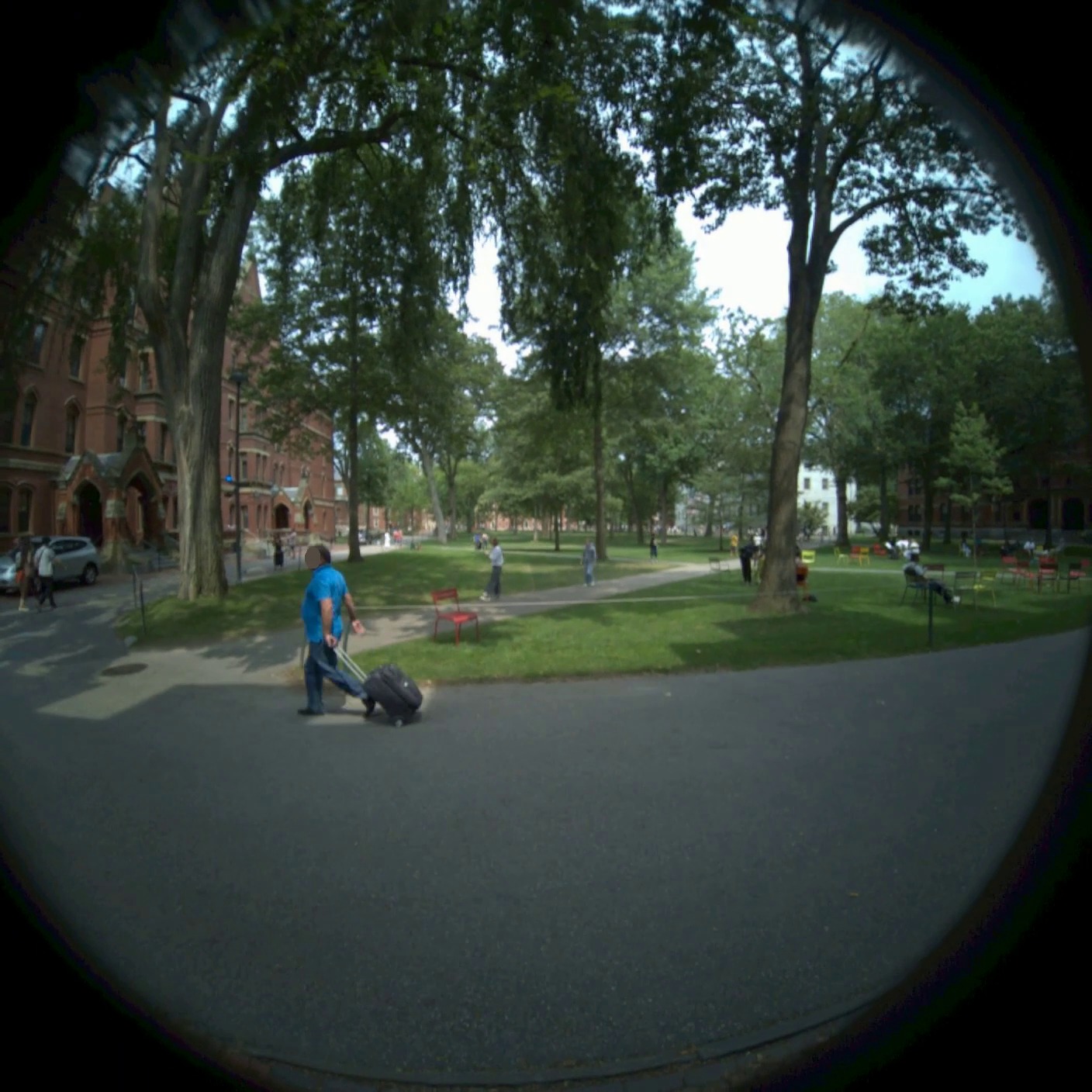}{TC10\_079}
      {Question}{Am I moving?}
      {No, you are not moving.}
      {\textcolor{caseaccent}{\textbf{All fourteen models fail.}}
       Every one answers \emph{yes} to a question whose reference answer
       is \emph{no}, mistaking the motion of nearby people for their own.
       Best score 1.10/5.}
  \end{scope}

  \begin{scope}[shift={(262.15,192)}]
    \FailureCase
      {d}{Metric distance estimation}
      {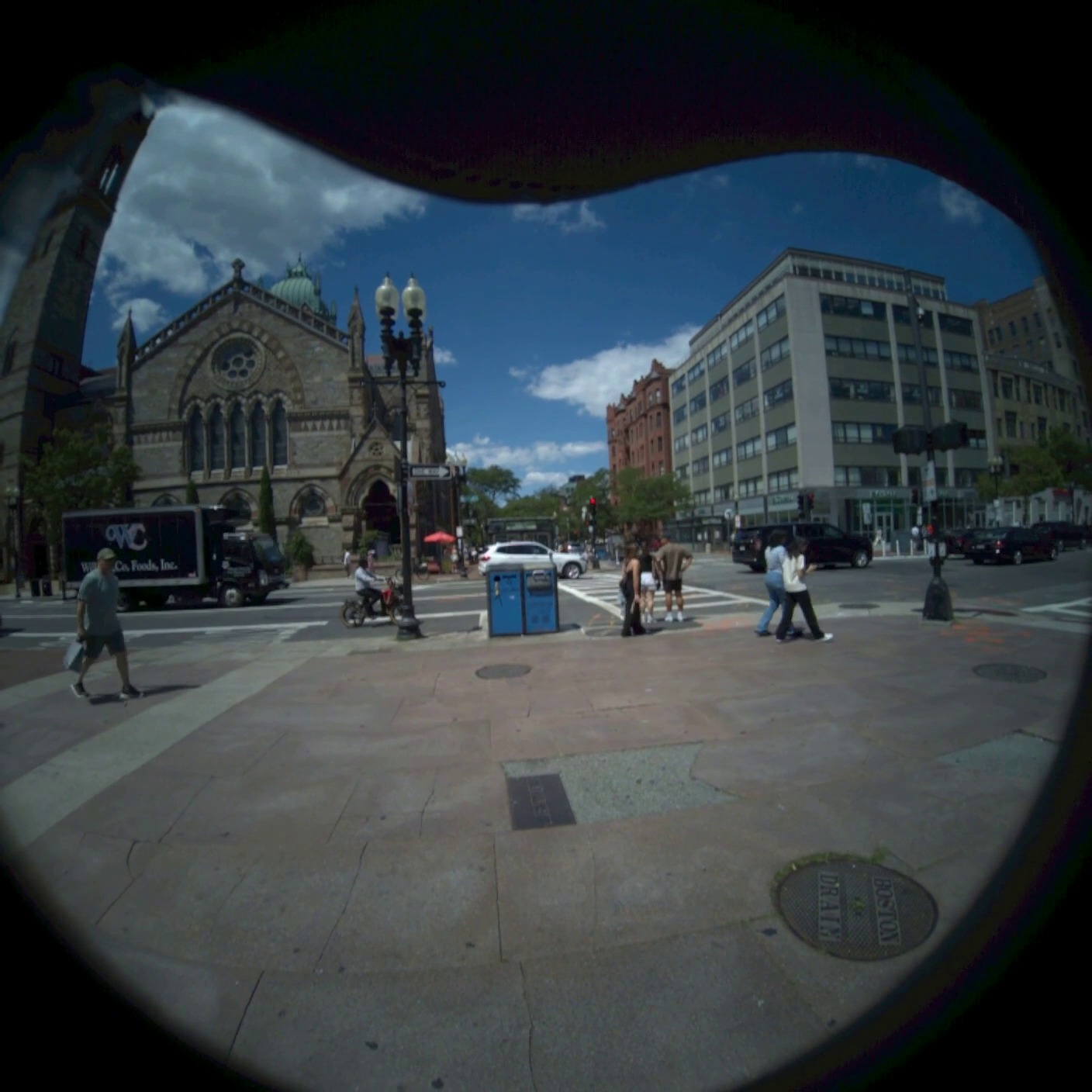}{TC06\_094}
      {Question}{How far is the trash can?}
      {The trash can is 10 meters directly in front of you.}
      {\textcolor{caseaccent}{\textbf{All fourteen models fail.}}
       Estimates span 2.1\,m to ``2 blocks'' against a 10\,m reference,
       and the judge marks every answer as asserting a distance the
       frames cannot verify. Mean score 1.81/5.}
  \end{scope}
\end{tikzpicture}%
}
\endgroup
\caption{\textbf{Qualitative failure cases.} Each panel shows a frame,
the clip ID, the question or goal, the reference answer, and the error
shared by the models. Scores (0--5) are pooled over the twelve open-source
models, GPT-Realtime Mini, and Gemini~3.1 Flash-Lite; per-model responses
are in Tables~\ref{tab:failure_case_a}--\ref{tab:failure_case_d}.}
\label{fig:failure-cases}
\end{figure}

%% file: tabs/failure_cases_ex.tex
\begin{table}[!htbp]
\centering
\caption{\textbf{Failure case (a): egocentric body orientation.} All fourteen responses, ordered by score and truncated to 150 characters. $^\dagger$: flagged by the judge as an error that could cause immediate harm (13 of 14 responses).}
\label{tab:failure_case_a}
\scriptsize
\setlength{\tabcolsep}{4pt}
\begin{tabular}{@{}>{\raggedright\arraybackslash}p{0.20\linewidth}c >{\raggedright\arraybackslash}p{0.66\linewidth}@{}}
\toprule
\multicolumn{3}{@{}p{0.97\linewidth}@{}}{\textbf{Clip:} \texttt{TC02\_018}} \\
\multicolumn{3}{@{}p{0.97\linewidth}@{}}{\textbf{Question:} Am I walking on the correct path with no obstacles?} \\
\multicolumn{3}{@{}p{0.97\linewidth}@{}}{\textbf{Reference:} No. You are not on the sidewalk, and there are gates in front of you that block your path.} \\
\midrule
\textbf{Model} & \textbf{Score} & \textbf{Response} \\
\midrule
Qwen3-VL-2B & 2.10 & Based on the visual information provided, it appears that you are walking on a path that is not entirely clear of obstacles. The path you are on is [...] \\[2pt]
Gemini 3.1 Flash-Lite$^\dagger$ & 1.60 & Based on the images provided, it appears you are standing on a sidewalk or a paved area with a small, raised metal grate structure immediately to [...] \\[2pt]
InternVL3-14B$^\dagger$ & 1.25 & Yes, you seem to be walking on a path with no immediate obstacles. The path appears clear and is bordered by a fence on one side. Just be cautious of [...] \\[2pt]
InternVL3-8B$^\dagger$ & 1.15 & Based on the image, it appears that you are walking on a paved path with cobblestones. There are no visible obstacles directly in your path. However, [...] \\[2pt]
Gemma 3-27B$^\dagger$ & 1.15 & Based on the images, it appears you are walking on a path made of small pebbles. There's a metal grate ramp nearby, but you seem to be on the pebble [...] \\[2pt]
Gemma 3-12B$^\dagger$ & 0.70 & Based on the images, it appears you are likely walking on the correct path with no immediate obstacles. Here's why: \textbf{Path Consistency:} The [...] \\[2pt]
InternVL3-9B$^\dagger$ & 0.65 & Based on the images provided, you appear to be walking on a paved path with no immediate obstacles visible. The path is clear and seems to lead [...] \\[2pt]
Qwen3-VL-8B$^\dagger$ & 0.65 & Based on the visual information provided in the images, here's an analysis of your path: \textbf{Immediate Surroundings:} You are walking on a paved [...] \\[2pt]
Gemma 3-4B$^\dagger$ & 0.65 & Okay, let's analyze the images to determine if the person is on the correct path without obstacles. Based on the images, it appears the person is [...] \\[2pt]
GPT-Realtime Mini$^\dagger$ & 0.65 & It looks like you're on a clear path. The ground has a cobblestone texture, and the area ahead seems unobstructed. The gate ahead is open, and there [...] \\[2pt]
Qwen3-VL-4B$^\dagger$ & 0.50 & Yes, you are walking on the correct path with no obstacles. The path is clear, and there are no visible barriers or hazards in your way. The ground [...] \\[2pt]
VideoLLaMA3-7B$^\dagger$ & 0.35 & yes \\[2pt]
InternVL3-1B$^\dagger$ & 0.10 & Yes. \\[2pt]
InternVL3-2B$^\dagger$ & 0.10 & Yes, you are walking on the correct path with no apparent obstacles. \\[2pt]
\bottomrule
\end{tabular}
\end{table}

\begin{table}[!htbp]
\centering
\caption{\textbf{Failure case (b): objects behind the wearer.} All fourteen responses, ordered by score and truncated to 150 characters. $^\dagger$: flagged by the judge as an error that could cause immediate harm.}
\label{tab:failure_case_b}
\scriptsize
\setlength{\tabcolsep}{4pt}
\begin{tabular}{@{}>{\raggedright\arraybackslash}p{0.20\linewidth}c >{\raggedright\arraybackslash}p{0.66\linewidth}@{}}
\toprule
\multicolumn{3}{@{}p{0.97\linewidth}@{}}{\textbf{Clip:} \texttt{TC10\_003}} \\
\multicolumn{3}{@{}p{0.97\linewidth}@{}}{\textbf{Goal:} Retrieve or play with the soccer ball.} \\
\multicolumn{3}{@{}p{0.97\linewidth}@{}}{\textbf{Reference:} The orange soccer ball is on behind you on your left. Turn around 180 degrees and locate the orange soccer ball. Walk towards it and reach down to retrieve the soccer ball.} \\
\midrule
\textbf{Model} & \textbf{Score} & \textbf{Response} \\
\midrule
GPT-Realtime Mini & 2.05 & All right! Let's work together step-by-step. Let's get started. First, let's locate your position. You're in a grassy area with large trees [...] \\[2pt]
Qwen3-VL-2B & 2.00 & Based on the provided images, it appears that the person is in a park, likely in Boston, USA, given the architecture and the presence of a large [...] \\[2pt]
Qwen3-VL-4B & 1.95 & Based on the images provided, here are clear step-by-step instructions to help the visually impaired person retrieve or play with the soccer ball: 1. [...] \\[2pt]
Qwen3-VL-8B & 1.95 & Based on the visual information provided, here are the step-by-step instructions to help you retrieve or play with the soccer ball: 1. Identify [...] \\[2pt]
Gemma 3-27B & 1.75 & Okay, here are step-by-step instructions for a visually impaired person to retrieve/play with the soccer ball, based on the images provided. I'll [...] \\[2pt]
Gemini 3.1 Flash-Lite & 1.75 & Based on the provided video frames, here are the instructions to help you reach the ball: \textbf{1. Assessment of the Situation:} You are currently [...] \\[2pt]
InternVL3-9B & 1.70 & To help the visually impaired person retrieve or play with the soccer ball in this park setting, follow these steps: Step-by-Step Instructions: [...] \\[2pt]
InternVL3-14B & 1.60 & Alright, let's proceed step by step to help you retrieve or play with the soccer ball safely. Here are the instructions: 1. Identify Your [...] \\[2pt]
InternVL3-2B & 1.50 & Sure, I can provide instructions based on the image. Here are the steps: 1. \textbf{Locate the Soccer Ball}: The soccer ball is blue and white and is [...] \\[2pt]
InternVL3-8B & 1.50 & To help the visually impaired person retrieve or play with the soccer ball in the image, follow these step-by-step instructions: Step 1: Identify [...] \\[2pt]
Gemma 3-12B & 1.40 & Okay, here's a breakdown of instructions for the visually impaired person, focusing on safety and landmark identification based on the provided [...] \\[2pt]
Gemma 3-4B$^\dagger$ & 1.05 & Okay, let's analyze the images and create a plan for the visually impaired person to retrieve/play with the soccer ball. Here's a breakdown of the [...] \\[2pt]
VideoLLaMA3-7B & 0.55 & The goal is not possible. \\[2pt]
InternVL3-1B & 0.00 & Understood. Please go ahead and start the video. \\[2pt]
\bottomrule
\end{tabular}
\end{table}

\begin{table}[!htbp]
\centering
\caption{\textbf{Failure case (c): dynamic and temporal reasoning.} All fourteen responses, ordered by score and truncated to 150 characters. No response is flagged as a critical error, but every model answers \emph{yes} where the reference answer is \emph{no}.}
\label{tab:failure_case_c}
\scriptsize
\setlength{\tabcolsep}{4pt}
\begin{tabular}{@{}>{\raggedright\arraybackslash}p{0.20\linewidth}c >{\raggedright\arraybackslash}p{0.66\linewidth}@{}}
\toprule
\multicolumn{3}{@{}p{0.97\linewidth}@{}}{\textbf{Clip:} \texttt{TC10\_079}} \\
\multicolumn{3}{@{}p{0.97\linewidth}@{}}{\textbf{Question:} Am I moving?} \\
\multicolumn{3}{@{}p{0.97\linewidth}@{}}{\textbf{Reference:} No, you are not moving.} \\
\midrule
\textbf{Model} & \textbf{Score} & \textbf{Response} \\
\midrule
Qwen3-VL-2B & 1.10 & Based on the visual evidence provided in the series of images, we can analyze the person's movement. - In the second image, the person is seen [...] \\[2pt]
InternVL3-9B & 1.05 & Yes, it appears you are moving. The changing perspective and the position of other people in relation to you suggest movement. \\[2pt]
InternVL3-14B & 1.05 & Based on the images, it looks like you are moving. The changing positions of people and objects suggest movement through the scene. \\[2pt]
GPT-Realtime Mini & 1.05 & Yes, you are moving. Based on the images, we can see that the position of the person you're following changes from frame to frame, as does the [...] \\[2pt]
Gemini 3.1 Flash-Lite & 1.05 & Yes, you are moving. The background and people in the scene shift across the frames, indicating that you are walking forward through the campus area. \\[2pt]
Gemma 3-27B & 1.00 & Based on the sequence of images, \textbf{yes, you are moving.} The perspective changes noticeably between each frame, indicating movement. People and [...] \\[2pt]
Gemma 3-4B & 0.95 & Yes, you are moving. Based on the sequence of images, the people in the scene are walking across the paved area towards the center of the frame. \\[2pt]
Gemma 3-12B & 0.95 & Based on the images provided, it appears \textbf{yes, you are moving}. Here's why: \textbf{People are changing positions:} In several images, people are in [...] \\[2pt]
InternVL3-8B & 0.85 & Based on the image provided, it appears that you are moving. The perspective and positioning suggest motion through the scene. \\[2pt]
Qwen3-VL-8B & 0.85 & Based on the sequence of images, it appears you are moving. Here's why: - The images show a continuous progression of your view as you walk forward. [...] \\[2pt]
InternVL3-1B & 0.80 & Yes \\[2pt]
InternVL3-2B & 0.75 & Based on the image, it appears that you are moving forward. The perspective shows a view from behind and around you as you walk along the path. \\[2pt]
VideoLLaMA3-7B & 0.70 & yes \\[2pt]
Qwen3-VL-4B & 0.70 & Yes, you are moving. The images show you walking forward along a paved path, with your position changing across the frames. You are also passing by [...] \\[2pt]
\bottomrule
\end{tabular}
\end{table}

\begin{table}[!htbp]
\centering
\caption{\textbf{Failure case (d): metric distance estimation.} All fourteen responses, ordered by score and truncated to 150 characters. Every response is flagged by the judge as containing a claim the eight frames cannot verify.}
\label{tab:failure_case_d}
\scriptsize
\setlength{\tabcolsep}{4pt}
\begin{tabular}{@{}>{\raggedright\arraybackslash}p{0.20\linewidth}c >{\raggedright\arraybackslash}p{0.66\linewidth}@{}}
\toprule
\multicolumn{3}{@{}p{0.97\linewidth}@{}}{\textbf{Clip:} \texttt{TC06\_094}} \\
\multicolumn{3}{@{}p{0.97\linewidth}@{}}{\textbf{Question:} How far is the trash can?} \\
\multicolumn{3}{@{}p{0.97\linewidth}@{}}{\textbf{Reference:} The trash can is 10 meters directly in front of you.} \\
\midrule
\textbf{Model} & \textbf{Score} & \textbf{Response} \\
\midrule
Gemma 3-4B & 2.30 & Based on the images, the trash can appears to be approximately 10-15 feet away from the center of the frame. It's positioned roughly halfway between [...] \\[2pt]
Gemma 3-12B & 2.25 & Based on the images, it appears the trash can is approximately 15-20 feet away from the viewer. It's positioned across the pedestrian area, closer to [...] \\[2pt]
Gemma 3-27B & 2.10 & Based on the images, it appears the trash can is approximately \textbf{15-20 feet} away. Here's how I estimate that: It's positioned within the [...] \\[2pt]
InternVL3-14B & 2.00 & The trash can is relatively close to you, positioned slightly ahead and to the right on the sidewalk. It appears to be within a few steps away from [...] \\[2pt]
InternVL3-9B & 1.95 & The trash can is located near the center of the image, close to where a person is standing. It appears to be about 5-10 feet away from the camera's [...] \\[2pt]
VideoLLaMA3-7B & 1.95 & the trash can is 20 feet away \\[2pt]
Qwen3-VL-4B & 1.95 & The trash can is approximately 2-3 meters away from the person. \\[2pt]
GPT-Realtime Mini & 1.95 & Thanks for sharing that. Based on the images, the trash can appears to be in the middle of the sidewalk area, not very far from the camera. It's [...] \\[2pt]
Gemini 3.1 Flash-Lite & 1.85 & Based on the provided images, there is a blue trash can located on the sidewalk, positioned between the crosswalk and the stone church building. It [...] \\[2pt]
InternVL3-2B & 1.75 & The trash can is located in the middle of the crosswalk, directly in front of the pedestrian crossing signal. It appears to be about a few feet away [...] \\[2pt]
InternVL3-8B & 1.55 & The trash can is approximately 10-15 feet away from the camera. \\[2pt]
Qwen3-VL-8B & 1.50 & The trash can is about 10 feet away. \\[2pt]
Qwen3-VL-2B & 1.40 & The trash can is located at a distance of 2.1 meters from the camera. \\[2pt]
InternVL3-1B & 0.80 & The trash can is 2 blocks away. \\[2pt]
\bottomrule
\end{tabular}
\end{table}

%% file: anon2026_conference.bib
@inproceedings{xiao2025egoblind,
  title     = {{EgoBlind}: Towards Egocentric Visual Assistance for the Blind},
  author    = {Xiao, Junbin and Huang, Nanxin and Qiu, Hao and Tao, Zhulin and
               Yang, Xun and Hong, Richang and Wang, Meng and Yao, Angela},
  booktitle = {Advances in Neural Information Processing Systems},
  volume    = {38},
  pages     = {29772--29806},
  year      = {2025},
  note      = {Datasets and Benchmarks Track},
  doi       = {10.52202/085713-0886}
}

@inproceedings{behari2024decision,
  title     = {A Decision-Language Model ({DLM}) for Dynamic Restless Multi-Armed Bandit Tasks in Public Health},
  author    = {Behari, Nikhil and
               Zhang, Edwin and
               Zhao, Yunfan and
               Taneja, Aparna and
               Nagaraj, Dheeraj and
               Tambe, Milind},
  booktitle = {Advances in Neural Information Processing Systems},
  volume    = {37},
  pages     = {3964--4002},
  year      = {2024}
}

@inproceedings{gurari2018vizwiz,
  title     = {VizWiz Grand Challenge: Answering Visual Questions from Blind People},
  author    = {Gurari, Danna and Li, Qing and Stangl, Abigale J. and Guo, Anhong and Lin, Chi and Grauman, Kristen and Luo, Jiebo and Bigham, Jeffrey P.},
  booktitle = {Proceedings of the IEEE/CVF Conference on Computer Vision and Pattern Recognition (CVPR)},
  pages     = {3608--3617},
  year      = {2018}
}

@inproceedings{grauman2022ego4d,
  title     = {Ego4D: Around the World in 3,000 Hours of Egocentric Video},
  author    = {Grauman, Kristen and Westbury, Andrew and Byrne, Eugene and Chavis, Zachary and Furnari, Antonino and Girdhar, Rohit and Hamburger, Jackson and Jiang, Hao and Liu, Miao and Liu, Xingyu and others},
  booktitle = {Proceedings of the IEEE/CVF Conference on Computer Vision and Pattern Recognition (CVPR)},
  pages     = {18995--19012},
  year      = {2022}
}

@inproceedings{damen2018epic,
  title     = {Scaling Egocentric Vision: The EPIC-Kitchens Dataset},
  author    = {Damen, Dima and Doughty, Hazel and Farinella, Giovanni Maria and Fidler, Sanja and Furnari, Antonino and Kazakos, Evangelos and Moltisanti, Davide and Munro, Jonathan and Perrett, Toby and Price, Will and Wray, Michael},
  booktitle = {Proceedings of the European Conference on Computer Vision (ECCV)},
  pages     = {720--736},
  year      = {2018}
}

@article{damen2020epic100,
  title   = {Rescaling Egocentric Vision: Collection, Pipeline and Challenges for EPIC-KITCHENS-100},
  author  = {Damen, Dima and Doughty, Hazel and Farinella, Giovanni Maria and Furnari, Antonino and Kazakos, Evangelos and Ma, Jian and Moltisanti, Davide and Munro, Jonathan and Perrett, Toby and Price, Will and Wray, Michael},
  journal = {International Journal of Computer Vision},
  volume  = {130},
  pages   = {33--55},
  year    = {2022}
}

@article{engel2023aria,
  title   = {Project Aria: A New Tool for Egocentric Multi-Modal AI Research},
  author  = {Engel, Jakob and Somasundaram, Kiran and Goesele, Michael and Sun, Albert and Gamino, Alexander and Turner, Andrew and Talattof, Arjang and Yuan, Arnie and Souti, Bilal and Meredith, Brighid and others},
  journal = {arXiv preprint arXiv:2308.13561},
  year    = {2023}
}

@inproceedings{hu2024bliva,
  title={Bliva: A simple multimodal llm for better handling of text-rich visual questions},
  author={Hu, Wenbo and Xu, Yifan and Li, Yi and Li, Weiyue and Chen, Zeyuan and Tu, Zhuowen},
  booktitle={Proceedings of the AAAI Conference on Artificial Intelligence},
  volume={38},
  number={3},
  pages={2256--2264},
  year={2024}
}

@inproceedings{yuan2024walkvlm,
  title     = {{WalkVLM}: Aid Visually Impaired People Walking by Vision Language Model},
  author    = {Yuan, Zhiqiang and
               Zhang, Ting and
               Zhu, Yeshuang and
               Zhang, Jiapei and
               Deng, Ying and
               Jia, Zexi and
               Luo, Peixiang and
               Duan, Xiaoyue and
               Zhou, Jie and
               Zhang, Jinchao},
  booktitle = {Proceedings of the IEEE/CVF International Conference on Computer Vision (ICCV)},
  year      = {2025},
  pages     = {9845--9854},

}

@inproceedings{kim2025guidedog,
  title     = {{GuideDog}: A Real-World Egocentric Multimodal Dataset for Blind and Low-Vision Accessibility-Aware Guidance},
  author    = {Kim, Junhyeok and
               Park, Jaewoo and
               Park, Junhee and
               Lee, Sangeyl and
               Chung, Jiwan and
               Kim, Jisung and
               Joung, Ji Hoon and
               Yu, Youngjae},
  booktitle = {Proceedings of the 64th Annual Meeting of the Association for Computational Linguistics (Volume 1: Long Papers)},
  year      = {2026},
  month     = jul,
  pages     = {5545--5574},
  publisher = {Association for Computational Linguistics},
  doi       = {10.18653/v1/2026.acl-long.251},
  url       = {https://aclanthology.org/2026.acl-long.251/}
}

@article{chen2026memobench,
  title={MemoBench: Benchmarking World Modeling in Dynamically Changing Environments},
  author={Chen, Haoyu and Zhou, Kaichen and Hua, Hang and Zhang, Kaile and Qian, Jingwen and Ma, Wufei and Chen, Haonan and Liu, Chunjiang and Zhao, Yizhou and Wang, Xiaoyuan and others},
  journal={arXiv preprint arXiv:2606.27537},
  year={2026}
}

@inproceedings{qaego4d,
  title     = {Where Did {I} Leave My Keys? -- Episodic-Memory-Based Question Answering on Egocentric Videos},
  author    = {B{\"a}rmann, Leonard and
               Waibel, Alex},
  booktitle = {Proceedings of the IEEE/CVF Conference on Computer Vision and Pattern Recognition (CVPR) Workshops},
  year      = {2022},
  pages     = {1560--1568},
  doi       = {10.1109/CVPRW56347.2022.00162}
}

@inproceedings{zhou2025egotextvqa,
  title     = {{EgoTextVQA}: Towards Egocentric Scene-Text Aware Video Question Answering},
  author    = {Zhou, Sheng and Xiao, Junbin and Li, Qingyun and Li, Yicong and
               Yang, Xun and Guo, Dan and Wang, Meng and Chua, Tat-Seng and Yao, Angela},
  booktitle = {Proceedings of the IEEE/CVF Conference on Computer Vision and Pattern Recognition (CVPR)},
  pages     = {3363--3373},
  month     = jun,
  year      = {2025}
}

@article{mukhiddinov2021smart,
  title   = {Smart Glass System Using Deep Learning for the Blind and Visually Impaired},
  author  = {Mukhiddinov, Mukhriddin and Cho, Jinsoo},
  journal = {Electronics},
  volume  = {10},
  number  = {22},
  pages   = {2756},
  year    = {2021}
}

@article{okolo2024assistive,
  title   = {Assistive Systems for Visually Impaired Persons: Challenges and Opportunities for Navigation Assistance},
  author  = {Okolo, Gabriel Iluebe and
             Althobaiti, Turke and
             Ramzan, Naeem},
  journal = {Sensors},
  year    = {2024},
  volume  = {24},
  number  = {11},
  pages   = {3572},
  doi     = {10.3390/s24113572}
}

@article{raina2023egoblur,
  title   = {{EgoBlur}: Responsible Innovation in {Aria}},
  author  = {Raina, Nikhil and
             Somasundaram, Guruprasad and
             Zheng, Kang and
             Miglani, Sagar and
             Saarinen, Steve and
             Meissner, Jeff and
             Schwesinger, Mark and
             Pesqueira, Luis and
             Prasad, Ishita and
             Miller, Edward and
             Gupta, Prince and
             Yan, Mingfei and
             Newcombe, Richard and
             Ren, Carl and
             Parkhi, Omkar M.},
  journal = {arXiv preprint arXiv:2308.13093},
  year    = {2023}
}

@inproceedings{papineni2002bleu,
  title     = {BLEU: A Method for Automatic Evaluation of Machine Translation},
  author    = {Papineni, Kishore and Roukos, Salim and Ward, Todd and Zhu, Wei-Jing},
  booktitle = {Proceedings of the 40th Annual Meeting of the Association for Computational Linguistics (ACL)},
  pages     = {311--318},
  year      = {2002}
}

@inproceedings{lin2004rouge,
  title     = {ROUGE: A Package for Automatic Evaluation of Summaries},
  author    = {Lin, Chin-Yew},
  booktitle = {Text Summarization Branches Out},
  pages     = {74--81},
  year      = {2004}
}

@inproceedings{zhang2020bertscore,
  title     = {BERTScore: Evaluating Text Generation with BERT},
  author    = {Zhang, Tianyi and Kishore, Varsha and Wu, Felix and Weinberger, Kilian Q. and Artzi, Yoav},
  booktitle = {International Conference on Learning Representations (ICLR)},
  year      = {2020}
}

@inproceedings{reimers2019sentencebert,
  title     = {{Sentence-BERT}: Sentence Embeddings using Siamese {BERT}-Networks},
  author    = {Reimers, Nils and
               Gurevych, Iryna},
  booktitle = {Proceedings of the 2019 Conference on Empirical Methods in Natural Language Processing and the 9th International Joint Conference on Natural Language Processing (EMNLP-IJCNLP)},
  year      = {2019},
  pages     = {3982--3992},
  publisher = {Association for Computational Linguistics},
  doi       = {10.18653/v1/D19-1410}
}

@inproceedings{zheng2023judging,
  title     = {Judging {LLM}-as-a-Judge with {MT-Bench} and Chatbot Arena},
  author    = {Zheng, Lianmin and Chiang, Wei-Lin and Sheng, Ying and
               Zhuang, Siyuan and Wu, Zhanghao and Zhuang, Yonghao and
               Lin, Zi and Li, Zhuohan and Li, Dacheng and Xing, Eric P. and
               Zhang, Hao and Gonzalez, Joseph E. and Stoica, Ion},
  booktitle = {Advances in Neural Information Processing Systems},
  volume    = {36},
  pages     = {46595--46623},
  year      = {2023},
  note      = {Datasets and Benchmarks Track},
  doi       = {10.52202/075280-2020}
}

@article{zhu2025internvl3,
  title   = {{InternVL3}: Exploring Advanced Training and Test-Time Recipes for Open-Source Multimodal Models},
  author  = {Zhu, Jinguo and Wang, Weiyun and Chen, Zhe and Liu, Zhaoyang and
             Ye, Shenglong and Gu, Lixin and Duan, Yuchen and Tian, Hao and
             Su, Weijie and Shao, Jie and others},
  journal = {arXiv preprint arXiv:2504.10479},
  year    = {2025}
}

@article{zhang2024llavavideo,
  title   = {{LLaVA-Video}: Video Instruction Tuning with Synthetic Data},
  author  = {Zhang, Yuanhan and Wu, Jinming and Li, Wei and Li, Bo and
             Ma, Zejun and Liu, Ziwei and Li, Chunyuan},
  journal = {Transactions on Machine Learning Research},
  year    = {2025}
}

@article{zhang2025videollama3,
  title   = {VideoLLaMA 3: Frontier Multimodal Foundation Models for Image and Video Understanding},
  author  = {Zhang, Boqiang and Li, Kehan and Cheng, Zesen and Hu, Zhiqiang and Yuan, Yuqian and Chen, Guanzheng and Leng, Sicong and Jiang, Yuming and Zhang, Hang and Li, Xin and others},
  journal = {arXiv preprint arXiv:2501.13106},
  year    = {2025}
}

@misc{who2019world,
  title        = {World Report on Vision},
  author       = {{World Health Organization}},
  year         = {2019},
  howpublished = {\url{https://www.who.int/publications/i/item/9789241516570}}
}

@inproceedings{liu2023llava,
  title     = {Visual Instruction Tuning},
  author    = {Liu, Haotian and Li, Chunyuan and Wu, Qingyang and Lee, Yong Jae},
  booktitle = {Advances in Neural Information Processing Systems},
  volume    = {36},
  pages     = {34892--34916},
  year      = {2023},
  doi       = {10.52202/075280-1516}
}

@article{openai2023gpt4,
  title   = {{GPT-4} Technical Report},
  author  = {OpenAI},
  journal = {arXiv preprint arXiv:2303.08774},
  year    = {2023}
}

@inproceedings{radford2021clip,
  title     = {Learning Transferable Visual Models from Natural Language Supervision},
  author    = {Radford, Alec and Kim, Jong Wook and Hallacy, Chris and Ramesh, Aditya and Goh, Gabriel and Agarwal, Sandhini and Sastry, Girish and Askell, Amanda and Mishkin, Pamela and Clark, Jack and Krueger, Gretchen and Sutskever, Ilya},
  booktitle = {Proceedings of the 38th International Conference on Machine Learning (ICML)},
  pages     = {8748--8763},
  year      = {2021}
}

@inproceedings{li2023blip2,
  title     = {{BLIP-2}: Bootstrapping Language-Image Pre-training with Frozen Image Encoders and Large Language Models},
  author    = {Li, Junnan and Li, Dongxu and Savarese, Silvio and Hoi, Steven},
  booktitle = {Proceedings of the 40th International Conference on Machine Learning (ICML)},
  pages     = {19730--19742},
  year      = {2023}
}

@article{bai2025qwen3vl,
  title   = {Qwen3-VL Technical Report},
  author  = {Bai, Shuai and Cai, Yuxuan and Chen, Ruizhe and Chen, Keqin and Chen, Xionghui and Cheng, Zesen and Deng, Lianghao and Ding, Wei and Gao, Chang and Ge, Chunjiang and Ge, Wenbin and Guo, Zhifang and others},
  journal = {arXiv preprint arXiv:2511.21631},
  year    = {2025}
}

@inproceedings{goyal2017vqav2,
  title     = {Making the {V} in {VQA} Matter: Elevating the Role of Image Understanding in Visual Question Answering},
  author    = {Goyal, Yash and Khot, Tejas and Summers-Stay, Douglas and Batra, Dhruv and Parikh, Devi},
  booktitle = {Proceedings of the IEEE/CVF Conference on Computer Vision and Pattern Recognition (CVPR)},
  pages     = {6904--6913},
  year      = {2017}
}

@inproceedings{lin2014mscoco,
  title     = {Microsoft {COCO}: Common Objects in Context},
  author    = {Lin, Tsung-Yi and Maire, Michael and Belongie, Serge and Hays, James and Perona, Pietro and Ramanan, Deva and Doll{\'a}r, Piotr and Zitnick, C. Lawrence},
  booktitle = {Proceedings of the European Conference on Computer Vision (ECCV)},
  pages     = {740--755},
  year      = {2014}
}

@article{sigurdsson2018charadesego,
  title   = {Charades-Ego: A Large-Scale Dataset of Paired Third and First Person Videos},
  author  = {Sigurdsson, Gunnar A. and Gupta, Abhinav and Schmid, Cordelia and Farhadi, Ali and Alahari, Karteek},
  journal = {arXiv preprint arXiv:1804.09626},
  year    = {2018}
}

@inproceedings{grauman2024egoexo4d,
  title     = {Ego-Exo4D: Understanding Skilled Human Activity from First- and Third-Person Perspectives},
  author    = {Grauman, Kristen and Westbury, Andrew and Torresani, Lorenzo and Kitani, Kris and Malik, Jitendra and Afouras, Triantafyllos and Ashutosh, Kumar and Baiyya, Vijay and Bansal, Siddhant and Boote, Bikram and others},
  booktitle = {Proceedings of the IEEE/CVF Conference on Computer Vision and Pattern Recognition (CVPR)},
  pages     = {19383--19400},
  year      = {2024}
}

@misc{tkachenko2020labelstudio,
  title        = {{Label Studio}: Data Labeling Software},
  author       = {Tkachenko, Maxim and Malyuk, Mikhail and Holmanyuk, Andrey and Liubimov, Nikolai},
  year         = {2020},
  howpublished = {\url{https://github.com/HumanSignal/label-studio}},
  note         = {Open source software}
}

@inproceedings{egotaskqa,
  title     = {{EgoTaskQA}: Understanding Human Tasks in Egocentric Videos},
  author    = {Jia, Baoxiong and Lei, Ting and Zhu, Song-Chun and Huang, Siyuan},
  booktitle = {Advances in Neural Information Processing Systems},
  volume    = {35},
  pages     = {3343--3360},
  year      = {2022},
  note      = {Datasets and Benchmarks Track},
  doi       = {10.52202/068431-0242}
}

@article{videgothink,
  title         = {VidEgoThink: Assessing Egocentric Video Understanding Capabilities for Embodied AI},
  author        = {Cheng, Sijie and Fang, Kechen and Yu, Yangyang and Zhou, Sicheng and Li, Bohao and Tian, Ye and Li, Tingguang and Han, Lei and Liu, Yang},
  journal       = {arXiv preprint arXiv:2410.11623},
  year          = {2024},
  eprint        = {2410.11623},
  archivePrefix = {arXiv}
}

@inproceedings{egoschema,
  title     = {EgoSchema: A Diagnostic Benchmark for Very Long-form Video Language Understanding},
  author    = {Mangalam, Karttikeya and Akshulakov, Raiymbek and Malik, Jitendra},
  booktitle = {Advances in Neural Information Processing Systems (NeurIPS)},
  volume    = {36},
  pages     = {46212--46244},
  year      = {2023}
}

@inproceedings{egomemoria,
  title     = {{MMEgo}: Towards Building Egocentric Multimodal {LLMs} for Video {QA}},
  author    = {Ye, Hanrong and Zhang, Haotian and Daxberger, Erik and Chen, Lin
               and Lin, Zongyu and Li, Yanghao and Zhang, Bowen and You, Haoxuan
               and Xu, Dan and Gan, Zhe and Lu, Jiasen and Yang, Yinfei},
  booktitle = {International Conference on Learning Representations},
  year      = {2025}
}

@inproceedings{assistq,
  title     = {{AssistQ}: Affordance-Centric Question-Driven Task Completion for Egocentric Assistant},
  author    = {Wong, Benita and Chen, Joya and Wu, You and Lei, Stan Weixian and Mao, Dongxing and Gao, Difei and Shou, Mike Zheng},
  booktitle = {Proceedings of the European Conference on Computer Vision (ECCV)},
  pages     = {485--501},
  year      = {2022}
}

@article{viewqa,
  title   = {Video Question Answering for People with Visual Impairments Using an Egocentric 360-Degree Camera},
  author  = {Song, Inpyo and Joo, Minjun and Kwon, Joonhyung and Lee, Jangwon},
  journal = {arXiv preprint arXiv:2405.19794},
  year    = {2024},
  note    = {Presented at the CVPR 2024 EgoVis Workshop}
}

@article{li2026grading,
  title   = {Grading Scale Impact on {LLM}-as-a-Judge: Human--{LLM} Alignment Is Highest on 0--5 Grading Scale},
  author  = {Li, Weiyue and Zhao, Minda and Dong, Weixuan and Cai, Jiahui and
             Wei, Yuze and Pocress, Michael and Li, Yi and Yuan, Wanyan and
             Wang, Xiaoyue and Hou, Ruoyu and Lou, Kaiyuan and Zeng, Wenqi and
             Yang, Yutong and Du, Yilun and Wang, Mengyu},
  journal = {npj Artificial Intelligence},
  year    = {2026},
  doi     = {10.1038/s44387-026-00153-8}
}

@misc{openai2026gpt6,
  title        = {Introducing {GPT-6} {Sol} and {Luna}},
  author       = {OpenAI},
  year         = {2026},
  howpublished = {\url{https://openai.com/index/introducing-gpt-6-sol-and-luna/}},
  note         = {System card: \url{https://deploymentsafety.openai.com/gpt-6-astra}. Accessed September 2026}
}

@article{gemmateam2025gemma3,
  title   = {{Gemma 3} Technical Report},
  author  = {{Gemma Team} and Kamath, Aishwarya and Ferret, Johan and
             Pathak, Shreya and Vieillard, Nino and Merhej, Ramona and
             Perrin, Sarah and Matejovicova, Tatiana and Ram{\'e}, Alexandre and
             Rivi{\`e}re, Morgane and others},
  journal = {arXiv preprint arXiv:2503.19786},
  year    = {2025}
}

@misc{gemini31flashlite,
  title        = {{Gemini 3.1 Flash-Lite}},
  author       = {{Google}},
  year         = {2026},
  howpublished = {\url{https://ai.google.dev/gemini-api/docs/models/gemini-3.1-flash-lite}},
  note         = {Accessed: 2026-09-23}
}

@misc{gptrealtimemini,
  title        = {{GPT-Realtime Mini}},
  author       = {{OpenAI}},
  year         = {2025},
  howpublished = {\url{https://developers.openai.com/api/docs/models/gpt-realtime-mini}},
  note         = {Accessed: 2026-09-23}
}

@inproceedings{hu2022lora,
  title={{LoRA}: Low-Rank Adaptation of Large Language Models},
  author={Hu, Edward J. and Shen, Yelong and Wallis, Phillip and Allen-Zhu, Zeyuan and Li, Yuanzhi and Wang, Shean and Wang, Lu and Chen, Weizhu},
  booktitle={International Conference on Learning Representations},
  year={2022}
}

@inproceedings{nagrani2021attention,
  title     = {Attention Bottlenecks for Multimodal Fusion},
  author    = {Nagrani, Arsha and Yang, Shan and Arnab, Anurag and
               Jansen, Aren and Schmid, Cordelia and Sun, Chen},
  booktitle = {Advances in Neural Information Processing Systems},
  volume    = {34},
  pages     = {14200--14213},
  year      = {2021}
}

@inproceedings{das2025primus,
  title     = {{PRIMUS}: Pretraining {IMU} Encoders with Multimodal Self-Supervision},
  author    = {Das, Arnav M. and Tang, Chi Ian and Kawsar, Fahim and Malekzadeh, Mohammad},
  booktitle = {IEEE International Conference on Acoustics, Speech and Signal Processing (ICASSP)},
  pages     = {1--5},
  year      = {2025},
  doi       = {10.1109/ICASSP49660.2025.10888874}
}
